%% file: main.tex
\documentclass{article}

\usepackage[numbers, sort&compress]{natbib}
\usepackage[preprint]{neurips_2025}

\usepackage{mathtools}
\usepackage{amsthm}
\usepackage{tabularx}
\usepackage{subfiles}
\usepackage{titlesec}
\usepackage{multirow}
\usepackage{graphicx}
\usepackage{amsmath}
\usepackage[utf8]{inputenc} 
\usepackage[T1]{fontenc}    
\usepackage{url}            
\usepackage{booktabs}       
\usepackage{amsfonts}       
\usepackage{nicefrac}       
\usepackage{microtype}      
\usepackage{xcolor}         
\usepackage{amssymb}
\usepackage{tikz}
\usetikzlibrary{matrix}
\usepackage{wrapfig}
\usepackage{lipsum}
\usepackage{caption} 
\usepackage{adjustbox} 
\usepackage{multicol}
\usepackage{subcaption}
\newtheorem{theorem}{Theorem}
\newtheorem{lemma}{Lemma}
\newtheorem{definition}{Definition}
\newtheorem{assumption}{Assumption}
\newtheorem{problem}{Goal}
\theoremstyle{remark}
\newtheorem{remark}{Remark}
\usepackage[backref=page]{hyperref}
\usepackage[capitalize,noabbrev]{cleveref}
\renewcommand*{\backref}[1]{}
\renewcommand*{\backrefalt}[4]{%
    \ifcase #1%
        \textbf{(Not cited.)}%
    \or
        \textbf{(Cited at p.~#2.)}%
    \else
        \textbf{(Cited at pp.~#2.)}%
    \fi
}
\hypersetup{
    colorlinks,
    linkcolor={red},
    citecolor={teal},
    urlcolor={blue}
}
\usepackage{algorithm}
\usepackage{algorithmic}
\usepackage{xspace}
\newcommand{\sage}{\textsc{Sage}\xspace}
\newcommand{\gcn}{\textsc{Gcn}\xspace}
\newcommand{\gat}{\textsc{Gat}\xspace}
\newcommand{\gin}{\textsc{Gin}\xspace}
\usepackage{newfloat}
\usepackage{listings}
\usepackage{enumitem}
\setlist[itemize]{
    leftmargin=*,  
    itemsep=0pt,   
    topsep=0pt,
    parsep=0pt
}
\definecolor{darkgreen}{rgb}{0.56, 0.93, 0.56}
\definecolor{lightgreen}{rgb}{0.851, 0.980, 0.851}

\setlist[itemize]{
    leftmargin=*,  
    itemsep=0pt,   
    topsep=0pt,
    parsep=0pt
}
\setlist[enumerate]{
    leftmargin=*,  
    itemsep=0pt,   
    topsep=0pt,
    parsep=0pt
}

\hypersetup{
    colorlinks,
    linkcolor={red},
    citecolor={blue},
    urlcolor={blue}
}
\definecolor{LightGreen}{RGB}{204,255,204}
\definecolor{LightRed}{RGB}{255,204,204}
\usepackage{colortbl}
\usepackage{pifont}

\titlespacing{\section}{0pt}{\parskip}{-\parskip}
\titlespacing{\subsection}{0pt}{0pt}{-\parskip}
\titlespacing{\subsubsection}{0pt}{0pt}{-\parskip}

\setcitestyle{numbers}
\title{GraphFLEx: Structure Learning \underline{F}ramework for \underline{L}arge \underline{Ex}panding \underline{Graph}s}

\author{
  \textbf{Mohit Kataria\textsuperscript{1}} \\ Mohit.Kataria@scai.iitd.ac.in \and
  \textbf{Nikita Malik\textsuperscript{3}} \\ bsz218185@iitd.ac.in \and
  \textbf{Sandeep Kumar\textsuperscript{2,1,3}} \\ ksandeep@ee.iitd.ac.in \and
  \textbf{Jayadeva\textsuperscript{2,1}} \\ jayadeva@ee.iitd.ac.in \and
  \textsuperscript{1} Yardi School of Artificial Intelligence\\
  \textsuperscript{2}Department of Electrical Engineering\\
  \textsuperscript{3}Bharti School of Telecommunication Technology and Management\\
  Indian Institute of Technology Delhi \\
}

\begin{document}

\maketitle

\subfile{chapters/abstract.tex}
\subfile{chapters/introduction.tex}
\subfile{chapters/problem_formulation_and_background.tex}
\subfile{chapters/proposed_framework.tex}
\subfile{chapters/experiments.tex}
\subfile{chapters/conclusion.tex}

\bibliographystyle{ieeetr}


\input{main.bbl}
\newpage
\appendix
\subfile{chapters/appendix.tex}

\end{document}

%% file: chapters/abstract.tex
\begin{abstract}
Graph structure learning is a core problem in graph-based machine learning, essential for uncovering latent relationships and ensuring model interpretability. However, most existing approaches are ill-suited for large-scale and dynamically evolving graphs, as they often require complete re-learning of the structure upon the arrival of new nodes and incur substantial computational and memory costs. In this work, we propose GraphFLEx—a unified and scalable framework for Graph Structure Learning in Large and Expanding Graphs. GraphFLEx mitigates the scalability bottlenecks by restricting edge formation to structurally relevant subsets of nodes identified through a combination of clustering and coarsening techniques. This dramatically reduces the search space and enables efficient, incremental graph updates. The framework supports 48 flexible configurations by integrating diverse choices of learning paradigms, coarsening strategies, and clustering methods, making it adaptable to a wide range of graph settings and learning objectives. Extensive experiments across 26 diverse datasets and Graph Neural Network architectures demonstrate that GraphFLEx achieves state-of-the-art performance with significantly improved scalability. 
\end{abstract}

%% file: chapters/introduction.tex
\section{Introduction} \label{sec:intro}
Graph representations capture relationships between entities, vital across diverse fields like biology, finance, sociology, engineering, and operations research \cite{zhou2020graph, fout2017protein, wu2020graph, Malik_Gupta_Kumar_2025}. While some relationships, such as social connections or sensor networks, are directly observable, many, including gene regulatory networks, scene graph generation \cite{gu2019scene}, brain networks, \cite{zhu2021survey} and drug interactions, require inference \cite{allen2012comparing}. Even when available, graph data often contains noise, requiring denoising and recalibration. In such cases, inferring the correct graph structure becomes more crucial than the specific graph model or downstream algorithm.\\ \textit{Graph Structure Learning (GSL)} offers a solution, enabling the construction and refinement of graph topologies. GSL has been widely studied in both supervised and unsupervised contexts \cite{liu2022towards, chen2022graph}. In supervised GSL (s-SGL), the adjacency matrix and Graph Neural Networks (GNNs) are jointly optimized for a downstream task, such as node classification. Notable examples of s-GSL include $NodeFormer$ \cite{wu2022nodeformer}, $Pro-GNN$ \cite{jin2020graph}, $WSGNN$ \cite{lao2022variational}, and $SLAPS$ \cite{fatemi2021slaps}. Unsupervised GSL (u-SGL), on the other hand, focuses solely on learning the underlying graph structure, typically through adjacency or Laplacian matrices. Methods in this category include approximate nearest neighbours ($A-NN$) \cite{dong2011efficient, muja2014scalable}, k-nearest neighbours ($k-NN$) \cite{kmeans, wang}, covariance estimation ($emp. Cov.$) \cite{hsieh2011sparse}, graphical lasso ($GLasso$) \cite{friedman2008sparse}, $SUBLIME$ \cite{liu2022towards}, and signal processing techniques like $l2$-model,$log$-model and $large$-model \cite{dong2016learning, kalofolias2016learn}.\\
Supervised structure learning (s-SGL) methods have demonstrated effectiveness in specific tasks; however, their reliance on labeled data and optimization for downstream objectives—particularly node classification—significantly constrains their generalizability to settings where annotations are scarce or unavailable \cite{liu2022towards}. Unsupervised structure learning (u-SGL) methods, which constitute the focus of this work, offer broader applicability. Nevertheless, both s-SGL and u-SGL approaches exhibit critical limitations in their ability to scale to large graphs or adapt efficiently to expanding datasets.
\captionsetup{font=footnotesize}
\begin{wrapfigure}{r}{0.40\linewidth}
  \centering
  \subfloat[GraphFLEx]{\includegraphics[width=0.195\textwidth, trim={0cm 0cm 1.2cm 1.7cm}, clip]{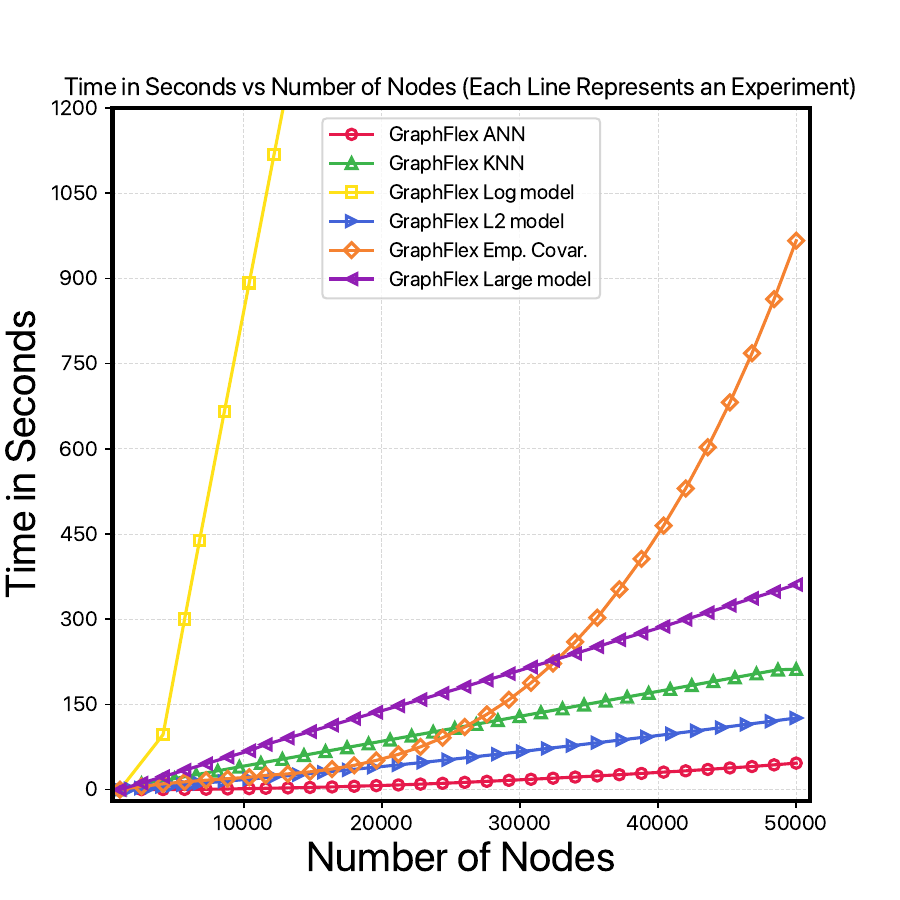}}
  \subfloat[Vanilla]{\includegraphics[width=0.195\textwidth, trim={0cm 0cm 1.2cm 1.7cm}, clip]{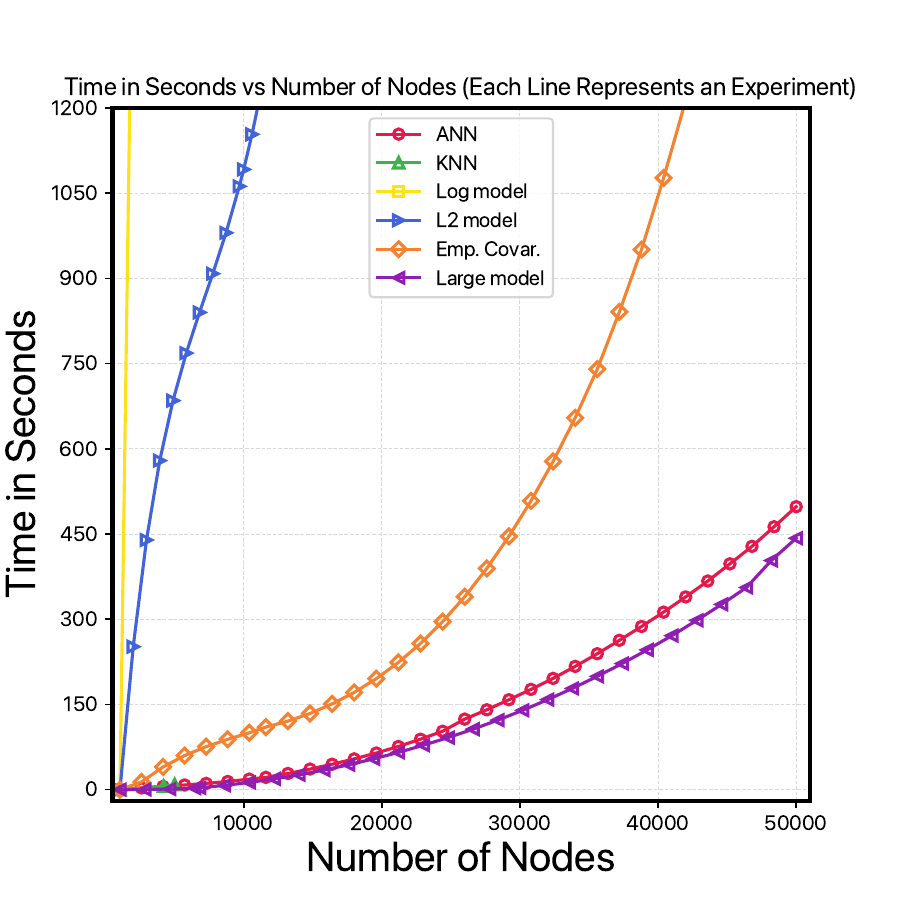}}
  \caption{High computational time required to learn graph structures using existing methods, whereas GraphFLEx effectively controls computational growth, achieving near-linear scalability. Notably, Vanilla KNN failed to construct graph structures for more than 10K nodes due to memory limitations.}
  \label{fig:expo_time}
  \vskip -0.15in
\end{wrapfigure}
\begin{figure*}[t]
\vskip -0.3in
\centering 
\includegraphics[width=1\textwidth, trim={4.7cm 4.3cm 4.3cm 4.9cm}, clip]{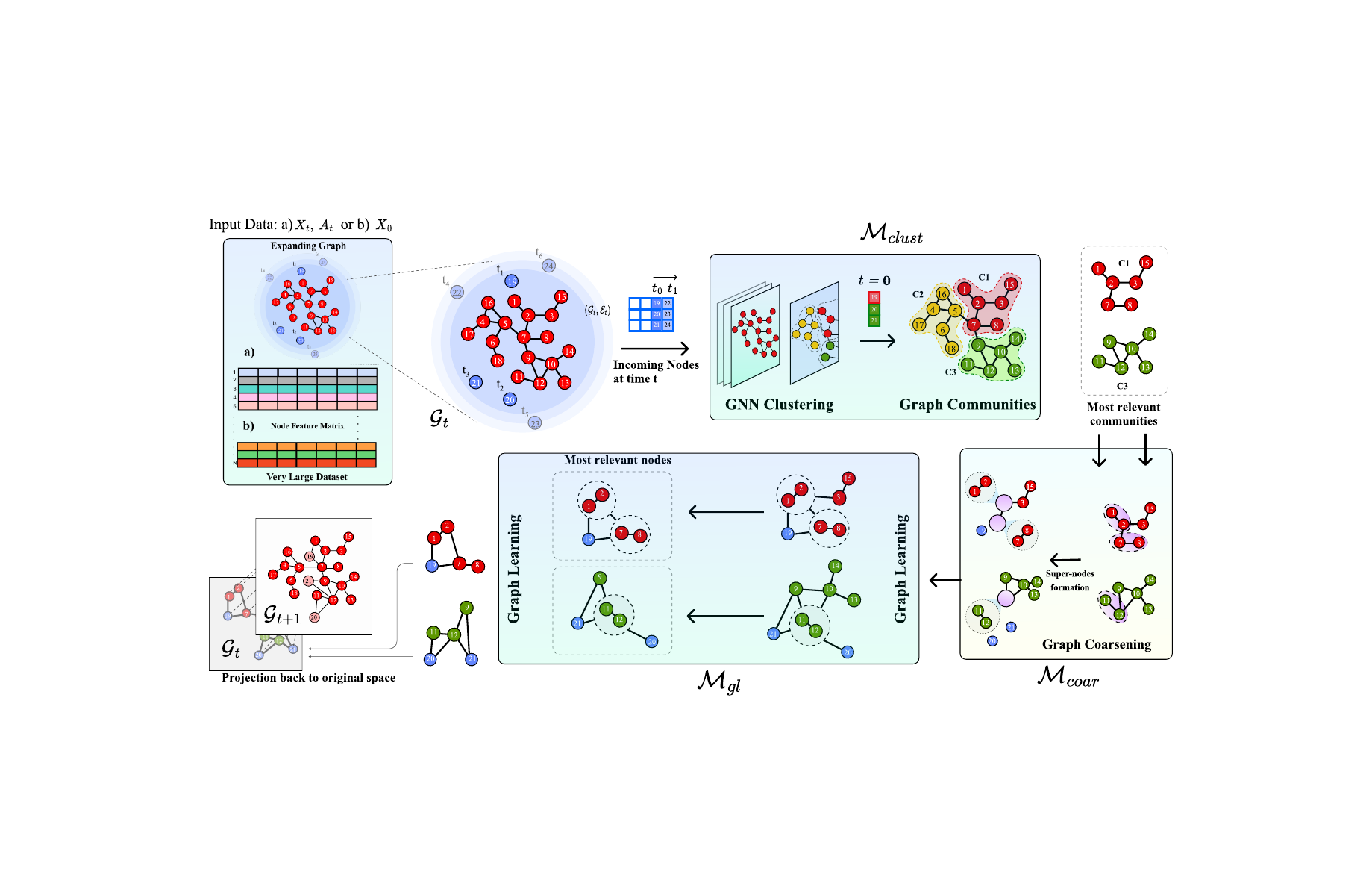}
\caption{This figure illustrates the general pipeline of GraphFLEx, designed to efficiently handle both a) large datasets with missing structure and b) expanding graphs. Both scenarios can be modeled as expanding graphs (details in Section \ref{sec:large_dynamic}). GraphFLEx processes a graph ($\mathcal{G}_{t}$) and incoming nodes ($\mathcal{E}_{t+1}$) at time $t$, newly arriving nodes are shown with different timestamps and shades of blue to indicate their arrival time. Our framework comprises of three main components: i) \textbf{Clustering}, which infers $\mathcal{E}_{t+1}$ nodes to existing communities using a pre-trained model $\mathcal{M}_{\text{clust}}(\mathcal{G}_0)$ into smaller, more manageable communities; ii) Since these communities may still be large, a \textbf{Coarsening}, module is applied to further reduce their size while preserving essential structural information; and iii) Finally, a \textbf{Learning} module, where the structure associated with $\mathcal{E}_{t+1}$ nodes are learned using the coarsened graph, followed by projecting this structure onto the $\mathcal{G}_{t}$ graph to create graph $\mathcal{G}_{t+1}$.}
\label{fig:main_fig}
\vskip -0.32in
\end{figure*}
\noindent
To address these challenges, we introduce \textbf{GraphFLEx}, a unified and scalable framework for \emph{Graph Structure Learning in Large and Expanding Graphs}. GraphFLEx is built upon the coordinated integration of three foundational paradigms in graph processing: \emph{graph clustering}, \emph{graph coarsening}, and \emph{structure learning}. While each of these methodologies has been studied extensively in isolation, their joint application within a single framework has remained largely unexplored. The novelty of GraphFLEx lies not merely in combining these components, but in the principled manner in which they are algorithmically aligned to reinforce one another—clustering serves to localize the search space, coarsening reduces structural redundancy while preserving global properties, and structure learning operates efficiently within this refined context. This integration enables GraphFLEx to scale effectively to large datasets and accommodate dynamic graphs through \emph{incremental updates}, eliminating the need for expensive re-training. Additionally, the framework supports \textbf{48 modular configurations}, enabling broad adaptability across datasets, learning objectives, and deployment constraints. Crucially, we establish \emph{theoretical guarantees} on edge recovery fidelity and computational complexity, offering rigorous foundations for the framework’s efficiency and reliability. As illustrated in Figure~\ref{fig:main_fig}, GraphFLEx significantly reduces the candidate edge space by operating on structurally relevant node subsets. Empirical evaluations, summarized in Figure~\ref{fig:expo_time}, demonstrate that GraphFLEx substantially outperforms existing baselines in both runtime and scalability.\\
\textbf{Key contributions of this work include:}\\
\vskip -0.25in
\begin{itemize}
\vskip -0.25in
    \item GraphFLEx unifies \textit{multiple structure learning strategies} within a single flexible framework.
    \item GraphFLEx demonstrates effectiveness in \textit{handling growing graphs}.
    \item GraphFLEx enhances the \textit{scalability} of graph structure learning on large-scale graphs.
    \item GraphFLEx serves as a \textit{comprehensive framework} applicable individually for clustering, coarsening, and learning tasks.
    \item We provide both \textit{empirical and theoretical results}, demonstrating the effectiveness of GraphFLEx across a range of datasets.
\end{itemize}

%% file: chapters/problem_formulation_and_background.tex
\section{Problem Formulation and Background}
\label{sec:problem_background}
A graph $\mathcal{G}$ is represented using $\mathcal{G}(V,A,X)$ where $V$ = $\{v_1,v_2...v_N\}$ is the set of $N$ nodes, each node $v_i$ has a $d-$dimensional feature vector $x_i$ in $X \in \mathbb{R}^{N \times d}$ and $A \in \mathbb{R}^{N \times N}$ is adjacency matrix representing connection between $i^{th}$ and $j^{th}$ nodes when entry $A_{ij} > 0$. An expanding graph $\mathcal{E}_{\mathcal{G}}$ can be considered a variant of graph $\mathcal{G}$ where nodes $v$ now have an associated timestamp $\tau_v$. We can represent a expanding graph as a sequence of graphs, i.e., $\mathcal{E}_{\mathcal{G}} = {\{\mathcal{G}_0,\mathcal{G}_1,...\mathcal{G}_T\}}$ where $\{\mathcal{G}_0 \subseteq \mathcal{G}_1.... \subseteq \mathcal{G}_T\}$ at $\tau \in \{0,...T\}$ timestamps. New nodes arriving at different timestamps are seamlessly integrating into initial graph $\mathcal{G}_0$.

\textit{\textbf{Problem statement.}} Given a partially known or missing graph structure, our goal is to incrementally learn the whole graph, i.e., learn adjacency or laplacian matrix. Specifically, we consider two unsupervised GSL tasks:
\begin{problem}
    \label{prob_1}
    \textit{\textbf{Large Datasets with Missing Graph Structure:}} In this setting, the graph structure is entirely unavailable, and existing methods are computationally infeasible for learning the whole graph in a single step. To address this issue, we first randomly partition the dataset into exclusive subsets. We then learn the initial graph $\mathcal{G}_0(V_0,X_0)$ over a small subset of nodes and incrementally expand it by integrating additional partitions, ultimately reconstructing the full graph $\mathcal{G}_T$.
\end{problem}
\begin{problem}
    \label{prob_2}
    \textit{\textbf{Partially Available Graph:}} In this case, we only have access to the graph $\mathcal{G}_t$ at timestamp $t$, with new nodes arriving over time. The goal is to update the graph incrementally to obtain $\mathcal{G}_T$, without re-learning it from scratch at each timestamp.
\end{problem}
GraphFlex addresses these challenges with a unified framework, outlined in Section~\ref{sec:our_method}. Before delving into the framework, we review some key concepts.
\vskip -0.2in
\subsection{Graph Reduction}
Graph reduction encompasses sparsification, clustering, coarsening, and condensation \cite{hashemi2024comprehensive}. GraphFlex employs clustering and coarsening to refine the set of relevant nodes for potential connections.\\
\textbf{Graph Clustering.}  Graphs often exhibit global heterogeneity with localized homogeneity, making them well-suited for clustering \cite{fortunato2010community}. Clusters capture higher-order structures, aiding graph learning. Methods like DMoN \cite{DMon} use GNNs for soft cluster assignments, while Spectral Clustering (SC) \cite{SC} and K-means \cite{KmeansClust, kmeans} efficiently detect communities. DiffPool \cite{bruna2014spectral,defferrard2016convolutional} applies SC for pooling in GNNs.\\
\textbf{Graph Coarsening.} Graph Coarsening (GC) reduces a graph $\mathcal{G}(V,E,X)$ with $N$ nodes and features $X\in \mathbb{R}^{N\times d}$ into a smaller graph $\mathcal{G}_c(\widetilde{V},\widetilde{E},\widetilde{X})$ with $n \ll N$ nodes and $\widetilde{X} \in \mathbb{R}^{n \times d}$. This is achieved via learning a coarsening matrix $\mathcal{P} \in \mathbb{R}^{n\times N}$, mapping similar nodes in $\mathcal{G}$ to super-nodes in $\mathcal{G}_c$, ensuring $\widetilde{X} = \mathcal{P} X$ while preserving key properties \cite{loukas2019graph,kataria2023linear,FGC,kataria2024ugc}.
\subsection{Unsupervised Graph Structure Learning}  
\label{sec:related_work}  
Unsupervised graph learning spans from simple k-NN weighting \cite{wang, Zhu} to advanced statistical and graph signal processing (GSP) techniques. Statistical methods, also known as probabilistic graphical models, assume an underlying graph $\mathcal{G}$ governs the joint distribution of data $X \in \mathbb{R}^{N \times d}$ \cite{koller2009probabilistic,banerjee2008model,friedman2008sparse}. Some approaches \cite{dempster1972covariance} prune elements in the inverse sample covariance matrix $ \widehat{\Sigma} = \frac{1}{d - 1} X X^T $ and sparse inverse covariance estimators, such as Graphical Lasso (GLasso) \cite{friedman2008sparse}:  
$\operatorname{maximize}_{\Theta} \log \operatorname{det} \Theta - \operatorname{tr}(\widehat{\boldsymbol{\Sigma}} \boldsymbol{\Theta}) - \rho\|\Theta\|_1,$ where $\Theta$ is the inverse covariance matrix. However, these methods struggle with small sample sizes. Graph Signal Processing (GSP) techniques analyze signals on known graphs, ensuring properties like smoothness and sparsity. Signal smoothness on a graph $\mathcal{G}$ is quantified by the Laplacian quadratic form: $Q(\mathbf{L}) = \mathbf{x}^T \mathbf{L x} = \frac{1}{2} \sum_{i, j} w_{i j} (\mathbf{x}(i)-\mathbf{x}(j))^2$. For a set of vectors $X$, smoothness is measured using the Dirichlet energy \cite{belkin2006manifold}: $\operatorname{tr}(X^T L X)$. State-of-the-art methods \cite{dong2016learning, kalofolias2016learn, hu2013graph} optimize Dirichlet energy while enforcing sparsity or specific structural constraints. Table~\ref{t:graph_learning_methods} in Appendix~\ref{app:related_work} compares various graph learning methods based on their formulations and time complexities.  More recently, SUBLIME~\cite{liu2022towards} learns graph structure in an unsupervised manner by leveraging self-supervised contrastive learning to align a learnable graph with a dynamically refined anchor graph derived from the data itself.
\begin{remark}
Graph Structure Learning (GSL) differs significantly from Continual Learning (CL) \cite{van2019three, zhang2022cglb, parisi2019continual} and Dynamic Graph Learning (DGL) \cite{kim2022dygrain, wu2023continual, you2022roland}, as discussed in Appendix~\ref{CL-DGL}. 
\end{remark}

%% file: chapters/proposed_framework.tex
\section{GraphFLEx}
\label{sec:our_method}
In this section, we introduce GraphFLEx, which has three main modules: 
\begin{itemize}
    \item \textbf{Graph Clustering.} Identifies communities and extracts higher-order structural information,
    \item \textbf{Graph Coarsening.} Is used to coarsen down the desired community, if the community itself is large,
    \item \textbf{Graph Learning.} Learns the graph's structure using a limited subset of nodes from the clustering and coarsening modules, \textit{enabling scalability}.
\end{itemize}
For pseudocode, see Algorithm~\ref{alg:algorithm} in Appendix ~\ref{app:algo}. 
\subsection{Incremental Graph Learning for Large Datasets}
\label{sec:large_dynamic}
Real-world graph data is continuously expanding. For instance, e-commerce networks accumulate new clicks and purchases daily \cite{xiang2010temporal}, while academic networks grow with new researchers and publications \cite{wang2020microsoft}. To manage such growth, we propose incrementally learning the graph structure over smaller segments. \\ Given a large dataset $\mathcal{L}(V_{\mathcal{L}},X_{\mathcal{L}})$, where $V_{\mathcal{L}}$ is the node set and $X_{\mathcal{L}}$ represents node features, we define an \textit{expanding dataset} setting $\mathcal{L}_{\mathcal{E}} = \{\mathcal{E}^T_{\tau=0}\}$. Initially, $\mathcal{L}$ is split into:  
(i) a \textit{static dataset} $\mathcal{E}_0(V_0,X_0)$ and  
(ii) an \textit{expanding dataset} $\mathcal{E} = \{\mathcal{E}_\tau(V_\tau,X_\tau)\}_{\tau=1}^{T}$.  
Both \textit{Goal \ref{prob_1}} (large datasets with missing graph structure) and \textit{Goal \ref{prob_2}} (partially available graphs with incremental updates), discussed in Section \ref{sec:problem_background}, share the common objective of incrementally learning and updating the graph structure as new data arrives. 
GraphFLEx handles these by decomposing the problem into two key components:  
\begin{itemize}
    \item \textbf{Initial Graph $\mathcal{G}_0(V_0, A_0, X_0)$:}  
    For \textit{Goal \ref{prob_1}}, where the graph structure is entirely missing, $\mathcal{E}_0(V_0, X_0)$ is used to construct $\mathcal{G}_0$ from scratch using structure learning methods (see Section \ref{sec:related_work}).  
    For \textit{Goal \ref{prob_2}}, the initial graph $\mathcal{G}_0(V_0, A_0, X_0)$ is already available and serves as the starting point for incremental updates.  
    \item \textbf{Expanding Dataset $\mathcal{E} = \{\mathcal{E}_\tau(V_\tau, X_\tau)\}_{\tau=1}^{T}$:}  
    In both cases, $\mathcal{E}$ consists of incoming nodes and features arriving over $T$ timestamps. These nodes are progressively integrated into the existing graph, enabling continuous adaptation and growth.  
\end{itemize}
The partition is controlled by a parameter $r$, which determines the proportion of static nodes: $r = \frac{\|V_0\|}{\|V_{\mathcal{L}}\|}$. For example, $r=0.2$ implies that 20\% of $V_{\mathcal{L}}$ is treated as static, while the remaining 80\% arrives incrementally over $T$ timestamps. In our experiments, we set $r = 0.5$ and $T = 25$.

\begin{remark}
    We can learn $\mathcal{G}_\tau(V_\tau,A_\tau,X_\tau)$ by aggregating $\mathcal{E}_{\tau}$ nodes in $\mathcal{G}_{\tau - 1}$ graph. Our goal is to learn $\mathcal{G}_T(V_T,A_T,X_T)$ after $T^{th}$-timestamp.
\end{remark}

\subsection{Detecting Communities}
From the static graph $\mathcal{G}_0$, our goal is to learn higher-order structural information, identifying potential communities to which incoming nodes ($V \in V\tau$) may belong. We train the community detection/clustering model $\mathcal{M}_{\text{clust}}$ once using $\mathcal{G}_0$, allowing subsequent inference of clusters for all incoming nodes. While our framework supports spectral and k-means clustering, our primary focus has been on Graph Neural Network (GNN)-based clustering methods. Specifically, we use DMoN \cite{DMon, bianchi2020spectral, bianchi2022simplifying}, which maximizes spectral modularity. Modularity \cite{newman2006modularity} measures the divergence between intra-cluster edges and the expected number. These methods use a GNN layer to compute the partition matrix $C = \text{softmax}(\text{MLP}(\widetilde{X}, \theta_{\text{MLP}})) \in \mathbb{R}^{N \times K}$, where $K$ is the number of clusters and $\widetilde{X}$ is the updated feature embedding generated by one or more message-passing layers. To optimize the $C$ matrix, we minimize the loss function $\Delta(C; A) = -\frac{1}{2m} \text{Tr}(C^T B C) + \frac{\sqrt{k}}{n} | \Sigma_i C_i^T |_F - 1$, which combines spectral modularity maximization with regularization to prevent trivial solutions, where $B$ is the modularity matrix \cite{DMon}. Our static graph $\mathcal{G}_0$ and incoming nodes $\mathcal{E}$ follow Assumption \ref{assum_1}. \\
\begin{assumption}
    \vspace{-0.45cm}
    \label{assum_1}
    Based on the well-established homophily principle, which forms the basis of most graph coarsening and learning methods. We assume that the generated graphs adhere to the Degree-Corrected Stochastic Block Model (DC-SBM) \cite{zhao2012consistency}, where intra-class (or intra-community) links are more likely than inter-class links. DC-SBM, an extension of SBM that accounts for degree heterogeneity, making it a more flexible and realistic choice for real-world networks.
\end{assumption}
For more details on DC-SBM, see Appendix \ref{app:dc_sbm}.
\begin{lemma}
\label{lemma_cons}
\textbf{$\mathcal{M}_{\text{clust}}$ Consistency.} We adopt the theoretical framework of \cite{zhao2012consistency} for a DC-SBM with \( N \) nodes and \( k \) classes.
The edge probability matrix is parameterized as \( P_N = \rho_N P \), where \( P \in \mathbb{R}^{k \times k} \) is a symmetric matrix containing the between/within community edge probabilities  and it is independent of \( N \), \( \rho_N = \lambda_N / N \), and \( \lambda_N \) is the average degree of the network. 
Let \( \hat{y}_N = [\hat{y}_1, \hat{y}_2, \ldots, \hat{y}_N] \) denote the predicted class labels, and let \( \hat{C}_N \) be the corresponding \( N \times k \) one-hot matrix. Let the true class label matrix is \( C_N \), and \( \mu \) is any \( k \times k \) permutation matrix. Under the adjacency matrix \( A^{(N)} \), the global maximum of the objective \( \Delta(\cdot; A^{(N)}) \) is denoted as \( \hat{C}_N^* \). The consistency of class predictions is defined as:
\begin{enumerate}
    \item \textbf{Strong Consistency.}  
    \small
    \abovedisplayskip=0pt \belowdisplayskip=0pt
    \[
    P_N \left[ \min_{\mu} \| \hat{C}_N^* \mu - C_N \|_F^2 = 0 \right] \to 1 \quad \text{as } N \to \infty,
    \]
    \item \textbf{Weak Consistency.}  
    \small
    \abovedisplayskip=0pt \belowdisplayskip=0pt
    \[
    \forall \varepsilon > 0, P_N \left[ \min_{\mu} \frac{1}{N} \| \hat{C}_N^* \mu - C_N \|_F^2 < \varepsilon \right] \to 1 \text{ as } N \to \infty.
    \]
\end{enumerate}
where \( \| \cdot \|_F \) is the Frobenius norm. Under the conditions of Theorem 3.1 from \cite{zhao2012consistency}:
\begin{itemize}
    \item The $\mathcal{M}_{\text{clust}}$ objective is strongly consistent if \( \lambda_N / \log(N) \to \infty \), and
    \item It is weakly consistent when \( \lambda_N \to \infty \).
\end{itemize}
\end{lemma}
\vskip 0.1in

\begin{remark}
\textbf{Structure Learning within Communities.} In $GraphFLEx$, we focus on learning the structure within each community rather than the structure of the entire dataset at once. Strong consistency ensures perfect community recovery, meaning no inter-community edges exist representing the ideal case. Weak consistency, however, allows for a small fraction (\(\epsilon\)) of inter-community edges, where \(\epsilon\) is controlled by \(\rho_n\) in \( P_n = \rho_n P \), influencing graph sparsity.  

By Lemma \ref{lemma_cons} and Assumption \ref{assum_1}, stronger consistency leads to more precise structure learning, whereas weaker consistency permits a limited number of inter-community edges.
\end{remark}

\subsection{Learning Graph Structure on a Coarse Graph}  
After training $\mathcal{M}_{\text{clust}}$, we identify communities for incoming nodes, starting with $\tau = 1$. Once assigned, we determine significant communities those with at least one incoming node and learn their connections to the respective community subgraphs. For large datasets, substantial community sizes may again introduce scalability issues. To mitigate this, we first coarsen the large community graph into a smaller graph and use it to identify potential connections for incoming nodes. This process constitutes the second module of GraphFLEx, denoted as $\mathcal{M}_{\text{coar}}$, which employs LSH-based hashing for graph coarsening. The supernode index for $i^{th}$ node is given as:  
\begin{equation}  
\mathcal{H}_i = maxOccurance \left\{\left\lfloor \frac{1}{r} \cdot (\mathcal{W} \cdot X_i + b) \right\rfloor \right\}
\end{equation}  
where $r$ (bin width) controls the coarsened graph size, $\mathcal{W}$ represents random projection matrix, $X$ is the feature matrix, and $b$ is the bias term. For further details, refer to UGC~\cite{kataria2024ugc}. After coarsening the $i^{th}$ community ($C_i$), $\mathcal{M}_{\text{coar}}(C_i) = \{\mathcal{P}_i, S_i\}$ yields a partition matrix $\mathcal{P}_i \in \mathbb{R}^{\|S_i\| \times \|C_i\|}$ and a set of coarsened supernodes ($S_i$), as discussed in Section~\ref{sec:problem_background}.
\\
To identify potential connections for incoming nodes, we define their neighborhood as follows:
\begin{definition} The neighborhood of a set of nodes \( \mathcal{E}_i \) is defined as the union of the top \( k \) most similar nodes in \( C_i \) for each node \( v \in \mathcal{E}_i \), where similarity is measured by the distance function \( d(v, u) \). A node \( u \in C_i \) is considered part of the neighborhood if its distance \( d(v, u) \) is among the \( k \) smallest distances for all \( u' \in C_i \).
\small
\[
\mathcal{N}_k(\mathcal{E}_i) = \bigcup_{v \in \mathcal{E}_i} \{ u \in C_i \mid d(v, u) \leq \text{top-}k[d(v, u') : u' \in C_i] \}
\]
\end{definition}
\begin{problem}
    The neighborhood of incoming nodes $\mathcal{N}_k(\mathcal{E}_i)$ represents the ideal set of nodes where the incoming nodes $\mathcal{E}_i$ are likely to establish connections when the entire community is provided to a structure learning framework.. A robust coarsening framework must reduce the number of nodes within each community $C_i$ while ensuring that the neighborhood of the incoming nodes is preserved.
\end{problem}
\subsection{Graph Learning only with Potential Nodes}
As we now have a smaller representation of the community, we can employ any graph learning algorithms discussed in Section~\ref{sec:related_work} to learn a graph between coarsened supernodes $S_i$ and incoming nodes ($V^i_\tau \in V_\tau$). This is the third module of GraphFLEx, i.e., graph learning; we denote it as $\mathcal{M}_{gl}$. The number of supernodes in $S_i$ is much smaller compared to the original size of the community, i.e., $\|S_i\| \ll \|C_i\|$; scalability is not an issue now. We learn a small graph first using $\mathcal{M}_{gl}(S_i,X_\tau^i) = \widetilde{\mathcal{G}}_\tau^i(V^c_\tau,A^c_\tau)$ where $X_\tau^i$ represents features of new nodes belonging to $i^{th}$ community at time $\tau$, $\widetilde{\mathcal{G}}_\tau^i(V^c_\tau,A^c_\tau)$ representing the graph between supernodes and incoming nodes. Utilizing the partition matrix $\mathcal{P}_i$ obtained from $\mathcal{M}_{\text{coar}}$, we can precisely determine the set of nodes associated with each supernode. For every new node $V \in V^i_\tau$, we identify the connected supernodes and subsequently select nodes within those supernodes. This subset of nodes is denoted by $\omega_{V_\tau^i}$, the sub-graph associated with $\omega_{V_\tau^i}$ represented by $\mathcal{G}_{\tau - 1}^i(\omega_{V_\tau^i})$ then undergoes an additional round of graph learning $\mathcal{M}_{gl}(\mathcal{G}_{\tau - 1}^i(\omega_{V_\tau^i}), X^i_\tau)$, ultimately providing a clear and accurate connection of new nodes $V^i_\tau$ with nodes of $\mathcal{G}_{\tau - 1}$, ultimately updating it to $\mathcal{G}_\tau$. This multi-step approach, characterized by coarsening, learning on coarsened graphs, and translation to the original graph, ensures scalability.
\\
\begin{theorem}
\label{theorem_neigh}
\textbf{Neighborhood Preservation.} Let \(\mathcal{N}_k(\mathcal{E}_i)\) denote the neighborhood of incoming nodes \(\mathcal{E}_i\) for the \(i^{\text{th}}\) community. With partition matrix $\mathcal{P}_i$ and $\mathcal{M}_{gl}(S_i,X_\tau^i) = \mathcal{G}_\tau^c(V^c_\tau,A^c_\tau)$ we identify the supernodes connected to incoming nodes $\mathcal{E}_i$ and subsequently select nodes within those supernodes; this subset of nodes is denoted by $\omega_{V_\tau^i}$. Formally, 
\small
\[
\omega_{V_\tau^i} = \bigcup_{v \in \mathcal{E}_i}  
\Bigl\{ \bigcup_{s\in S_i}\{\pi^{-1}(s)|A^c_\tau(v,s) \neq 0\}\Bigl\}
\]
Then, with probability \(\Pi_{\{c \in \phi\}} p(c)\), it holds that $\mathcal{N}_k(\mathcal{E}_i) \subseteq \omega_{V_\tau^i}$ where
\[
p(c) \leq 1 - \frac{2}{\sqrt{2 \pi}} \frac{c}{r}\left[1-e^{-r^2 /\left(2 c^2\right)}\right],
\]
and \(\phi\) is a set containing all pairwise distance values \((c = \|v - u\|)\) between every node \(v \in \mathcal{E}_i\) and the nodes \(u \in \omega_{V_\tau^i}\). Here, $\pi^{-1}(s)$ denotes the set of nodes mapped to supernode s, \(r\) is the bin-width hyperparameter of \(\mathcal{M}_{\text{coar}}\).
\end{theorem}
\begin{proof}
    The proof is deferred in Appendix \ref{app:neigh_preserv}.
\end{proof}

\begin{remark}
Theorem \ref{theorem_neigh} establishes that, with a constant probability of success, the neighborhood of incoming nodes $\mathcal{N}_k(\mathcal{E}_i)$ can be effectively recovered using the GraphFLEx multistep approach, which involves coarsening and learning on the coarsened graph, i.e., $\mathcal{N}_k(\mathcal{E}_i) \subseteq \omega_{V_\tau^i}$. The set $\omega_{V_\tau^i}$, estimated by GraphFLEx, identifies potential candidates where incoming nodes are likely to connect. The probability of failure can be reduced by regulating the average degree of connectivity in $\mathcal{M}_{gl}(S_i, X_\tau^i) = \mathcal{G}_\tau^c(V^c_\tau, A^c_\tau)$. While a fully connected network $\mathcal{G}_\tau^c$ ensures all nodes in the community are candidates, it significantly increases computational costs for large communities.
\end{remark}

\subsection{GraphFLEx: Multiple SGL Frameworks}
\captionsetup{font=footnotesize}
\begin{wrapfigure}{r}{0.47\linewidth}
\vspace{-1cm}
\centerline{\includegraphics[width=1\linewidth]{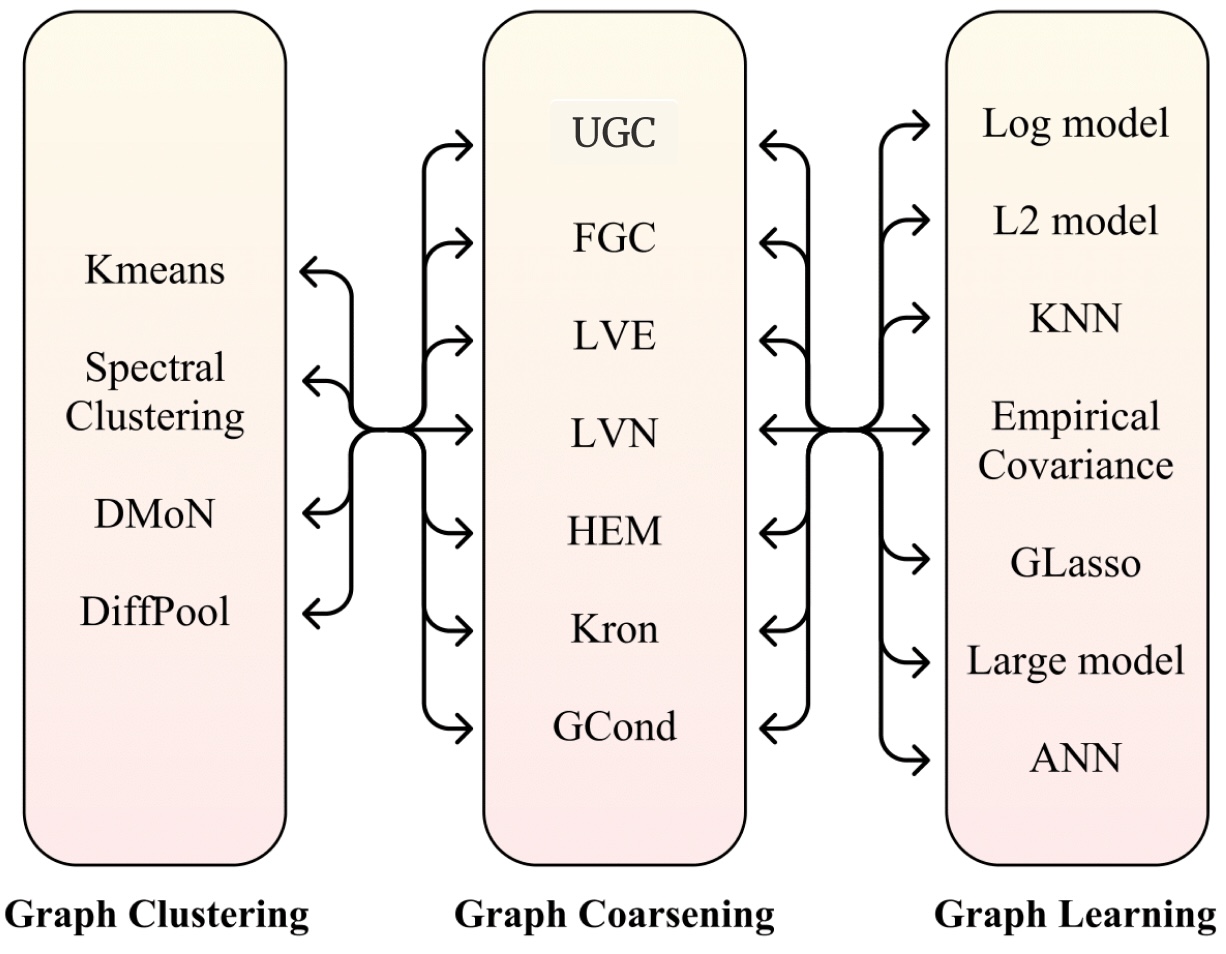}} 
\caption{The versatility of GraphFlex in supporting multiple GSL methods.}
\label{fig:multiple_frameworksinsec}
\end{wrapfigure}
Each module in Figure~\ref{fig:multiple_frameworksinsec} controls a distinct aspect of the graph learning process: clustering influences community detection, coarsening reduces graph complexity via supernodes, and the learning module governs structural inference. Altering any of these modules results in a new graph learning method. Currently, we support 48 different graph learning configurations, and this number scales exponentially with the addition of new methods to any module. The number of possible frameworks is given by $\alpha \times \beta \times \gamma$, where $\alpha$, $\beta$, and $\gamma$ represent the number of clustering, coarsening, and learning methods, respectively.  
\subsection{Run Time Analysis}
\label{sec:multi_frameworks}
\captionsetup{font=footnotesize}
\begin{table*}[t]
\centering
    \caption{\small Time complexity analysis of GraphFLEx. 
        Here, \(N\) is the number of nodes in the graph, \(k\) is the number of nodes in the static subgraph used for clustering (\(k \ll N\)), and \(c\) represents the number of detected communities. \(k_\tau\) denotes the number of nodes at timestamp \(\tau\). Finally, \(\alpha = \|S_\tau^i\| + \|\mathcal{E}^i_\tau\|\) is the sum of coarsened and incoming nodes in the relevant community at $\tau$ timestamp.}
    \setlength{\tabcolsep}{10pt} 
    \resizebox{\textwidth}{!}{
    \begin{tabular}{lcccc}
        \hline
        & $\mathcal{M}_{clust}$ & $\mathcal{M}_{coar}$ & $\mathcal{M}_{gl}$ & GraphFLEx \\ \hline
        \textbf{Best (kNN-UGC-ANN)} & $\mathcal{O}(k^2)$ & $\mathcal{O}\left(\frac{k_\tau}{c}\right)$ & $\mathcal{O}(\alpha \log \alpha)$ & $\mathcal{O}(k^2 + \frac{k_\tau}{c} + \alpha \log \alpha)$ \\
        \textbf{Worst (SC-FGC-GLasso)} & $\mathcal{O}(k^3)$ & $\mathcal{O}\left(\left(\frac{k_\tau}{c}\right)^2 \|S_\tau^i\|\right)$ & $\mathcal{O}(\alpha^3)$ & $\mathcal{O}(k^3 + \left(\frac{k_\tau}{c}\right)^2 \|S_\tau^i\| + \alpha^3)$ \\ \hline
    \end{tabular}
    }
    \label{tab:time_complexity}
    \vskip -0.25in
\end{table*}
GraphFLEx computational time is always bounded by existing approaches, as it operates on a significantly reduced set of nodes. We evaluate the run-time complexity of GraphFLEx in two scenarios: (a) the worst-case scenario, where computationally intensive clustering and coarsening modules are selected, providing an upper bound on time complexity, and (b) the best-case scenario, where the most efficient modules are chosen. Table \ref{tab:time_complexity} presents a summary of this analysis for both cases. Due to space limitations, a more comprehensive analysis is provided in Appendix ~\ref{app:run_time}.\\

%% file: chapters/experiments.tex
\vspace{-0.6cm}
\section{Experiments}
\label{sec:experiments}
\textbf{Tasks and Datasets.} To validate GraphFLEx’s utility, we evaluate it across four key dimensions: (i) computational efficiency, (ii) scalability to large graphs, (iii) quality of learned structures, and (iv) adaptability to dynamically growing graphs. To validate the characteristics of GraphFLEx, we conduct extensive experiments on 26 different datasets, including (a) datasets that already have a complete graph structure  (allowing comparison between the learned and the original structure), (b) datasets with missing graph structures, (c) synthetic datasets, (d) small datasets for visualizing the graph structure, and (e) large datasets, including datasets with even $2.4M$ nodes. More details about datasets and \textit{system specifications} are presented in Table \ref{t:datasets} in Appendix~\ref{app:datasets}.\\
\captionsetup{font=footnotesize}
\begin{table*}[!h]
\vspace{-0.5cm}
    \caption{Computational time(in seconds) for learning graph structures using GraphFLEx (GFlex) with existing methods (Vanilla referred to as Van.). The experimental setup involves treating 50\% of the data as static, while the remaining 50\% of nodes are treated as incoming nodes arriving in 25 different timestamps. The best times are highlighted by color \textcolor{green}{Green}. OOM and OOT denote out-of-memory and out-of-time, respectively.}
    \centering
    \footnotesize{
    \setlength{\tabcolsep}{2.5mm} 
    \renewcommand{\arraystretch}{1.2} 
    \resizebox{\textwidth}{!}{
    \begin{tabular}{l|cc|cc|cc|cc|cc|cc|cc}
        \toprule
        \bfseries Data & 
        \multicolumn{2}{c|}{\bfseries ANN} & 
        \multicolumn{2}{c|}{\bfseries KNN} & 
        \multicolumn{2}{c|}{\bfseries log-model} & 
        \multicolumn{2}{c|}{\bfseries l2-model} & 
        \multicolumn{2}{c|}{\bfseries emp-Covar.} & 
        \multicolumn{2}{c|}{\bfseries large-model} &
        \multicolumn{2}{c}{\bfseries Sublime} \\
        & \bfseries Van. & \bfseries GFlex & \bfseries Van. & \bfseries GFlex & \bfseries Van. & \bfseries GFlex & \bfseries Van. & \bfseries GFlex & \bfseries Van. & \bfseries GFlex & \bfseries Van. & \bfseries GFlex & \bfseries Van. & \bfseries GFlex \\
        \midrule
        Cora & 335 & \cellcolor{darkgreen}100 & \cellcolor{darkgreen}8.4 & 36.1 & 869 & \cellcolor{darkgreen}81.6 & 424 & \cellcolor{darkgreen}55 & \cellcolor{darkgreen}8.6 & 30 & 2115 & \cellcolor{darkgreen}18.4 & 7187 & \cellcolor{darkgreen}493\\
        Citeseer & 1535 & \cellcolor{darkgreen}454 & \cellcolor{darkgreen}21.9 & 75 & 1113 & \cellcolor{darkgreen}64.5 & 977 & \cellcolor{darkgreen}54.0 & \cellcolor{darkgreen}14.7 & 59.2 & 8319 & \cellcolor{darkgreen}43.9 & 8750 & \cellcolor{darkgreen}670\\
        DBLP & 2731 & \cellcolor{darkgreen}988 & OOM & \cellcolor{darkgreen}270 & 77000 & \cellcolor{darkgreen}919 & OOT & \cellcolor{darkgreen}1470 & 359 & \cellcolor{darkgreen}343 & OOT & \cellcolor{darkgreen}299 & OOM & \cellcolor{darkgreen}831\\
        CS & 22000 & \cellcolor{darkgreen}12000 & OOM & \cellcolor{darkgreen}789 & OOT & \cellcolor{darkgreen}838 & 32000 & \cellcolor{darkgreen}809 & 813 & \cellcolor{darkgreen}718 & OOT & \cellcolor{darkgreen}1469 & OOM & \cellcolor{darkgreen}1049\\
        PubMed & 770 & \cellcolor{darkgreen}227 & OOM & \cellcolor{darkgreen}164 & OOT & \cellcolor{darkgreen}176 & OOT & \cellcolor{darkgreen}165 & 488 & \cellcolor{darkgreen}299 & OOT & \cellcolor{darkgreen}262 & OOM & \cellcolor{darkgreen}914\\
        Phy. & 61000 & \cellcolor{darkgreen}21000 & OOM & \cellcolor{darkgreen}903 & OOT & \cellcolor{darkgreen}959 & OOT & \cellcolor{darkgreen}908 & 2152 & \cellcolor{darkgreen}1182 & OOT & \cellcolor{darkgreen}2414 & OOM & \cellcolor{darkgreen}2731\\
        Syn 3 & 95 & \cellcolor{darkgreen}37 & OOM & \cellcolor{darkgreen}30 & 58000 & \cellcolor{darkgreen}346 & 859 & \cellcolor{darkgreen}53 & 88 & \cellcolor{darkgreen}59 & 5416 & \cellcolor{darkgreen}42 & 6893 & \cellcolor{darkgreen}780\\
        Syn 4 & 482 & \cellcolor{darkgreen}71 & OOM & \cellcolor{darkgreen}73 & OOT & \cellcolor{darkgreen}555 & OOT & \cellcolor{darkgreen}145 & 2072 & \cellcolor{darkgreen}1043 & OOT & \cellcolor{darkgreen}392 & OOM & \cellcolor{darkgreen}1896\\
        \bottomrule
    \end{tabular}
    }
    }
    \label{t:time}
    \vskip -0.20in
\end{table*}
\captionsetup{font=footnotesize}
\begin{table*}[!htbp]
        \caption{Node classification accuracies on different GNN models using GraphFLEx (GFlex) with existing Vanilla (Van.) methods. The experimental setup involves treating 70\% of the data as static, while the remaining 30\% of nodes are treated as new nodes coming in 25 different timestamps. The best and the second-best accuracies in each row are highlighted by dark and lighter shades of \textcolor{green}{Green}, respectively. GraphFLEx's structure beats all of the vanilla structures for every dataset. OOM and OOT denotes out-of-memory and out-of-time respectively.}
    \centering
    \footnotesize{
    \setlength{\tabcolsep}{4.5pt}
    \resizebox{\textwidth}{!}{
        \begin{tabular}{lc|c*{9}{c|c}c}
            \toprule
            \textbf{Data} & \bfseries Model & 
            \multicolumn{2}{c|}{\bfseries ANN} & 
            \multicolumn{2}{c|}{\bfseries KNN} & 
            \multicolumn{2}{c|}{\bfseries log-model} & 
            \multicolumn{2}{c|}{\bfseries l2-model} & 
            \multicolumn{2}{c|}{\bfseries COVAR} & 
            \multicolumn{2}{c|}{\bfseries large-model} & 
            \multicolumn{2}{c|}{\bfseries Sublime} &
            \bfseries Base Struct. \\
            & & \bfseries Van. & \bfseries GFlex & \bfseries Van. & \bfseries GFlex & \bfseries Van. & \bfseries GFlex & \bfseries Van. & \bfseries GFlex & \bfseries Van. & \bfseries GFlex & \bfseries Van. & \bfseries GFlex & \bfseries Van. & \bfseries GFlex & \\
            \midrule
            & \gat & $34.23$ & $67.37$ & OOM & \cellcolor{lightgreen}$69.83$ &  OOT & \cellcolor{darkgreen}$69.83$ &  OOT & 68.98 & $50.48$ & $68.56$ & OOT & $66.38$ & OOM & $68.32$ & 70.84\\
            & \sage & $34.23$ & $69.58$ & OOM & $70.28$ &  OOT & $70.28$ &  OOT & \cellcolor{darkgreen}$70.68$ & $51.47$ & \cellcolor{lightgreen}$70.51$ & OOT & $69.32$ & OOM & $70.28$ & 72.57\\
            DBLP & \gcn & $34.12$ & $69.41$ & OOM & \cellcolor{darkgreen}$73.39$ &  OOT & \cellcolor{darkgreen}$73.39$ &  OOT & \cellcolor{lightgreen}73.05 & $51.50$ & $71.75$ & OOT & $68.55$ & OOM & $69.06$ & 74.43\\
            & \gin & $34.01$ & $69.69$ & OOM & $68.19$ &  OOT & $68.19$ &  OOT & \cellcolor{darkgreen}$73.08$ & $52.77$ & \cellcolor{lightgreen}$72.03$ & OOT & $71.18$ & OOM & $71.87$ & 73.92\\
            \midrule
            & \gat & $12.47$ & $60.89$ & OOM & 61.09 & OOT & 60.95 & 18.64 & 61.06 & 58.96 & \cellcolor{darkgreen}$88.06$ & OOT & \cellcolor{lightgreen}$86.22$ & OOM & $64.21$ & 60.75\\
            & \sage & $12.70$ & $78.81$ & OOM & 79.43 & OOT & 79.06 & 19.24 & 78.94 & 56.97 & \cellcolor{darkgreen}$93.30$ & OOT & \cellcolor{lightgreen}$92.79$ & OOM & $78.94$ & 80.33\\
            CS & \gcn & $12.59$ & $63.81$ & OOM & 67.94 & OOT & 69.33 & 19.21 & 66.01 & 58.35 &\cellcolor{darkgreen}91.07 & OOT & \cellcolor{lightgreen}$84.85$ & OOM & $68.92$ & 67.43\\
            & \gin & $13.07$ & $77.62$ & OOM & 78.41 & OOT & 78.55 & 19.24 & 77.61 & 58.26 & \cellcolor{darkgreen}92.07 & OOT & \cellcolor{lightgreen}$86.03$ & OOM & $77.61$ & 55.65\\
            \midrule
            & \gat & $49.49$ & $83.71$ & OOM & \cellcolor{darkgreen}$84.60$ &  OOT & \cellcolor{darkgreen}$84.60$ &  OOT & \cellcolor{lightgreen}84.04 & $72.63$ & $83.97$ & OOT & $81.15$ & OOM & $82.15$ & 84.04\\
            & \sage & $50.43$ & $87.27$ & OOM & \cellcolor{lightgreen}$87.34$ &  OOT & \cellcolor{lightgreen}$87.34$ &  OOT & \cellcolor{darkgreen}87.42 & $73.57$ & $86.68$ & OOT & \cellcolor{lightgreen}$87.34$ & OOM & $83.45$ & 88.88\\
            Pub. & \gcn & $50.45$ & $82.06$ & OOM & \cellcolor{lightgreen}$83.56$ &  OOT &  \cellcolor{lightgreen}$83.56$&  OOT & \cellcolor{darkgreen}83.74 & $73.14$ & $82.39$ & OOT & $78.03$ & OOM & $70.94$ & 85.54\\
            & \gin & $51.82$ & $83.13$ & OOM & \cellcolor{darkgreen}$84.31$ &  OOT & \cellcolor{lightgreen}$84.07$ &  OOT & 82.93 & $73.15$ & $83.51$ & OOT & $82.85$ & OOM & $80.72$  & 86.50\\
            \midrule
            & \gat & $29.18$ & $88.06$ & OOM & \cellcolor{lightgreen}$88.47$ &  OOT & \cellcolor{lightgreen}$88.47$ &  OOT & \cellcolor{darkgreen}88.68 & $58.96$ & $88.06$ & OOT & $86.22$ & OOM & $86.12$ & 88.58\\
            & \sage & $29.57$ & $93.47$ & OOM & $93.47$ &  OOT & $93.47$ &  OOT & \cellcolor{darkgreen}93.78 & $56.97$ & \cellcolor{lightgreen}$93.60$ & OOT & $92.79$ & OOM & $89.58$ & 94.19\\
            Phy. & \gcn & $27.84$ & \cellcolor{lightgreen}$91.27$ & OOM & $91.08$ &  OOT & $91.08$ &  OOT & \cellcolor{darkgreen}91.78 & $58.35$ & $91.07$ & OOT & $84.85$ & OOM & $88.46$ & 91.48\\
            & \gin & $28.38$ & \cellcolor{darkgreen}$92.69$ & OOM & $92.04$ &  OOT & $92.04$ &  OOT & \cellcolor{lightgreen}92.27 & $58.26$ & $92.07$ & OOT & $86.03$ & OOM & $87.20$ & 88.89\\
            \bottomrule
        \end{tabular}}}
        \label{t:gnn_accuracies}
\end{table*}
\vskip -0.15in
\subsection{Computational Efficiency.}  
Existing methods like $k$-NN and $log$-model struggle to learn graph structures even for 20k nodes due to out-of-memory (OOM) or out-of-time (OOT) issues, while $l2$-model and $large$-model struggle beyond 50k nodes. Although $A$-NN and $emp$-Covar. are faster, GraphFLEx outperforms them on sufficiently large graphs (Table~\ref{t:time}). While traditional methods may be efficient for small graphs, GraphFLEx scales significantly better, excelling on large datasets like \textit{Pubmed} and \textit{Syn 5}, where most methods fail. It accelerates structure learning, making $A$-NN 3× faster and $emp$-Covar. 2× faster.

\subsection{Node Classification Accuracy}
\label{sec:node_classification}
\textbf{Experimental Setup.} We now evaluate the prediction performance of GNN models when trained on graph structures learned from three distinct scenarios:  \textbf{1) Original Structure:} GNN models trained on the original graph structure, which we refer to as the Base Structure, \textbf{2) GraphFLEx Structure:} GNN models trained on the graph structure learned from GraphFLEx, and \textbf{3)Vanilla Structure:} GNN models trained on the graph structure learned from other existing methods. \\
For each scenario, a unique graph structure is obtained. We trained GNN models on each of these three structure. For more details on GNN model parameters, see Appendix~\ref{app:model_parameter}.\\
\textbf{GNN Models.}  
Graph neural networks (GNNs) such as $GCN$ \cite{kipf2016semi}, $GraphSage$ \cite{hamilton2017inductive}, $GIN$ \cite{xu2018powerful}, and $GAT$ \cite{velickovic2017graph} rely on accurate message passing, dictated by the graph structure, for effective embedding. We use these models to evaluate the above-mentioned learned structures. Table~\ref{t:gnn_accuracies} reports node classification performance across all methods. Notably, GraphFLEx outperforms vanilla structures by a significant margin across all datasets, achieving accuracies close to those obtained with the original structure. Figure~\ref{fig:sage_acc} in Appendix~\ref{app:model_parameter} illustrates $GraphSage$ classification results, highlighting GraphFLEx’s superior performance. For the $CS$ dataset, GraphFLEx ($large$-model) and GraphFLEx ($empCovar.$-model) even surpass the original structure, demonstrating its ability to preserve key structural properties while denoising edges, leading to improved accuracy.
\subsection{Scalability of GraphFLEx on Large-Scale Graphs.}
To comprehensively evaluate GraphFLEx’s scalability to large-scale graphs, we consider four datasets with a high number of nodes: (a) \textit{Flickr(89k nodes)} \cite{zeng2019graphsaint}, (b) \textit{Reddit (233k nodes)} \cite{zeng2019graphsaint}, (c) \textit{Ogbn-arxiv (169k nodes)} \cite{wang2020microsoft}, and (d) \textit{Ogbn-products (2.4M nodes)} \cite{Bhatia16}. As shown in Table~\ref{tab:large_datasets}, GraphFLEx consistently demonstrates superior scalability across all datasets, outperforming all baseline methods in runtime. In particular, methods such as \textit{log-model}, \textit{l2-model}, and \textit{large-model} fail to run even on \textit{Flickr}, while GraphFLEx successfully scales them on \textit{Flickr}, \textit{Ogbn-arxiv}, and \textit{Reddit}, enabling structure learning where others cannot. For the most computationally demanding dataset, \textit{Ogbn-products}, these methods remain prohibitively expensive even for GraphFLEx. Nonetheless, GraphFLEx efficiently supports scalable structure learning on \textit{Ogbn-products} using the \textit{Covar}, \textit{ANN}, and \textit{KNN} modules. Table~\ref{tab:large_datasets} also reports node classification accuracy, demonstrating that GraphFLEx maintains performance comparable to the original (base) structure across all datasets. These results confirm that GraphFLEx not only scales effectively, but also preserves the quality of learned structures.
\begin{table*}[htbp]
\centering
\footnotesize
\caption{Runtime (sec) and Node Classification Accuracy (\%) across large datasets. Each cell shows: \textbf{Time / Accuracy}. Van = Vanilla, GFlex = GraphFLEx. OOM = Out of Memory, OOT = Out of Time.}
\resizebox{\textwidth}{!}{
\begin{tabular}{l|cc|cc|cc|cc}
\toprule
\textbf{Method} & \multicolumn{2}{c|}{\textbf{ogbn-arxiv (60.13)}} & \multicolumn{2}{c|}{\textbf{ogbn-products (73.72)}} & \multicolumn{2}{c|}{\textbf{Flickr (44.92)}} & \multicolumn{2}{c}{\textbf{Reddit (94.15)}} \\
 & \textbf{Van.} & \textbf{GFlex} & \textbf{Van.} & \textbf{GFlex} & \textbf{Van.} & \textbf{GFlex} & \textbf{Van.} & \textbf{GFlex} \\
\midrule
Covar  & OOM | --     & 3.7k | 60.26 & OOM | --     & 83.1k | 68.23 & 2.3k | 44.65 & 682 | 44.34   & OOM | --     & 6.6k | 94.13 \\
ANN    & 7.8k | 60.14 & 4.8k | 60.22 & OOM | --     & 89.3k | 67.91 & 2.5k | 44.09 & 705 | 44.92   & 12.6k | 94.14 & 6.1k | 94.18 \\
knn    & 8.3k | 60.09 & 6.1k | 60.23 & OOM | --     & 91.8k | 68.47 & 2.7k | 43.95 & 920 | 44.73   & 15.6k | 94.14 & 6.9k | 94.15 \\
l2     & OOT | --     & 9.1k | 58.39 & OOT | --     & OOT | --    & 93.3k | 44.90 & 1.2k | 44.32   & OOT | --      & 5.1 | 93.47 \\
log    & OOT | --     & 45.6k | 58.72& OOT | --     & OOT | --      & OOT | -- & 18.7k | 44.59      & OOT | --      & 60.3k | 94.13 \\
large  & OOT | --     & 5.6k | 60.20 & OOT | --     & OOT | --      & OOT | -- & 2.2k | 44.45      & OOT | --      & 9.3k | 93.71 \\
\bottomrule
\end{tabular}
}
\label{tab:large_datasets}
\vskip -0.20in
\end{table*}

\noindent
\subsection{GraphFLEx for Link Prediction and Graph Classification.}
\captionsetup{font=footnotesize}
To further validate the generalization of our framework, we evaluate GraphFLEx on the link prediction task. The results are presented in Table~\ref{tab:accuracy_gflex}, following the same setting as Table \ref{t:gnn_accuracies}. The structure learned by GraphFLEx demonstrates strong predictive performance, in some cases even outperforming the base structure. This highlights the effectiveness of GraphFLEx in preserving and even enhancing relational information relevant for link prediction.
\begin{table*}[!htbp]
        \caption{Link predication accuracy (\%) across different datasets. The best and the second-best accuracies in each row are highlighted by dark and lighter shades of \textcolor{green}{Green}, respectively.}
    \centering
    \footnotesize{
    \setlength{\tabcolsep}{4.5pt}
    \resizebox{\textwidth}{!}{
        \begin{tabular}{l|c*{7}{c|c}c}
            \toprule
            \textbf{Data} & 
            \multicolumn{2}{c|}{\bfseries ANN} & 
            \multicolumn{2}{c|}{\bfseries KNN} & 
            \multicolumn{2}{c|}{\bfseries log-model} & 
            \multicolumn{2}{c|}{\bfseries l2-model} & 
            \multicolumn{2}{c|}{\bfseries COVAR} & 
            \multicolumn{2}{c|}{\bfseries large-model} &
            \bfseries Base Struct. \\
            & \bfseries Van. & \bfseries GFlex & \bfseries Van. & \bfseries GFlex & \bfseries Van. & \bfseries GFlex & \bfseries Van. & \bfseries GFlex & \bfseries Van. & \bfseries GFlex & \bfseries Van. & \bfseries GFlex \\
            \midrule
            DBLP    & 96.57 & 96.61 & OOM  & 94.23 & OOT  & \cellcolor{darkgreen}97.59 & OOT  & \cellcolor{darkgreen}97.59 & 97.22 & \cellcolor{darkgreen}97.59 & OOT  & 96.24 & 95.13 \\
            Citeseer& 80.12 & 96.32 & 85.17 & 96.24 & 80.48 & 96.24 & 80.48 & \cellcolor{darkgreen}96.48 & 82.05 & 96.24 & 84.50 & \cellcolor{lightgreen}94.38 & 90.78 \\
            Cora    & 84.47 & \cellcolor{lightgreen}95.30  & 79.23 & 95.14 & 90.63 & \cellcolor{darkgreen}95.45 & 90.81 & 95.14 & 86.05 & \cellcolor{lightgreen}95.30 & 90.63 & 94.67 & 89.53 \\
            Pubmed  & 94.24 & 96.91 & OOM  & \cellcolor{darkgreen}97.42 & OOT  & \cellcolor{darkgreen}97.42 & OOT  & 97.37 & 94.89 & 94.64 & OOT  & 94.41 & 94.64\\
            CS      & 94.21 & \cellcolor{lightgreen}95.73 & OOM  & \cellcolor{darkgreen}96.02 & OOT  & 93.17 & OOT  & 93.17 & 93.52 & 92.31 & OOT  & \cellcolor{lightgreen}95.73 & 95.00\\
            Physics & \cellcolor{darkgreen}95.77 & 91.34 & OOM  & \cellcolor{lightgreen}94.63 & OOT  & 90.79 & OOT  & \cellcolor{lightgreen}94.63 & 92.03 & 90.79 & OOT  & 92.97 & 93.96\\
            \bottomrule
        \end{tabular}}}
        \label{tab:accuracy_gflex}
        \vskip -0.15in
\end{table*}
\noindent
While our primary focus is on structure learning for node-level tasks, we briefly discuss the applicability of GraphFLEx to graph classification. In such tasks, especially in domains like molecule or drug discovery, each data point often corresponds to a small individual subgraph. For these cases, applying clustering and coarsening is typically redundant and may introduce unnecessary computational overhead. Nevertheless, GraphFLEx remains flexible—its learning module can be directly used without the clustering or coarsening steps, making it suitable for graph classification as well. This adaptability reinforces GraphFLEx’s utility across a broad range of graph learning tasks.

\subsection{Clustering Quality} 
\label{sec:clust_qual}
\begin{wraptable}{r}{0.5\textwidth}
\vspace{-0.5cm}
  \centering
  \scriptsize
  \setlength{\tabcolsep}{1.2mm}
  \renewcommand{\arraystretch}{1.1}
  \caption{Clustering (NMI, $\mathcal{C}$, $\mathcal{Q}$) and node classification accuracy using GCN, GraphSAGE, GIN, and GAT.}
  \label{tab:clust_results}
  \begin{tabular}{c|c|c|c|c|c|c|c}
    \toprule
    \textbf{Data} & \textbf{NMI} & $\mathcal{C}$ & $\mathcal{Q}$ & \textbf{GCN} & \textbf{SAGE} & \textbf{GIN} & \textbf{GAT} \\
    \midrule
    Bar. M. & 0.716 & 0.057 & 0.741 & 91.2 & 96.2 & 95.1 & 94.9 \\
    Seger.  & 0.678 & 0.102 & 0.694 & 91.0 & 93.9 & 94.2 & 92.3 \\
    Mura.   & 0.843 & 0.046 & 0.706 & 96.9 & 97.4 & 97.5 & 96.4 \\
    Bar. H. & 0.674 & 0.078 & 0.749 & 95.3 & 96.4 & 97.2 & 95.8 \\
    Xin     & 0.741 & 0.045 & 0.544 & 98.6 & 99.3 & 98.9 & 99.8 \\
    MNIST   & 0.677 & 0.082 & 0.712 & 92.9 & 94.5 & 94.9 & 82.6 \\
    \bottomrule
  \end{tabular}
  \vspace{-0.2cm}
\end{wraptable}
We measure three metrics to evaluate the resulting clusters or community assignments: a) Normalized Mutual Information (NMI) ~\cite{DMon}  between the cluster assignments and original labels; b) Conductance ($\mathcal{C}$) \cite{jerrum1988conductance} which measures the fraction of total edge volume that points outside the cluster; and c) Modularity ($\mathcal{Q}$) ~\cite{newman2006modularity} which measures the divergence between the intra-community edges and the expected one.
\noindent
Table ~\ref{tab:clust_results} illustrates these metrics for single-cell RNA and the MNIST dataset (where the whole structure is missing), and Figure~\ref{app:vis_clusters} in Appendix~\ref{app:clustering_quality} shows the PHATE ~\cite{moon2019visualizing} visualization of clusters learned using GraphFLEx's clustering module $\mathcal{M}_{clust}$. We also train the aforementioned GNN models for the node classification task in order to illustrate the efficacy of the learned structures; the accuracy values presented in Table ~\ref{tab:clust_results}, clearly highlight the significance of the learned structures, as reflected by the high accuracy values.
\captionsetup{font=footnotesize}
\begin{figure*}[t]
\begin{center}
  \begin{subfigure}[t]{.15\linewidth}
    \includegraphics[width=\linewidth, trim={2.3cm 1.4cm 1.9cm 1.6cm}, clip]{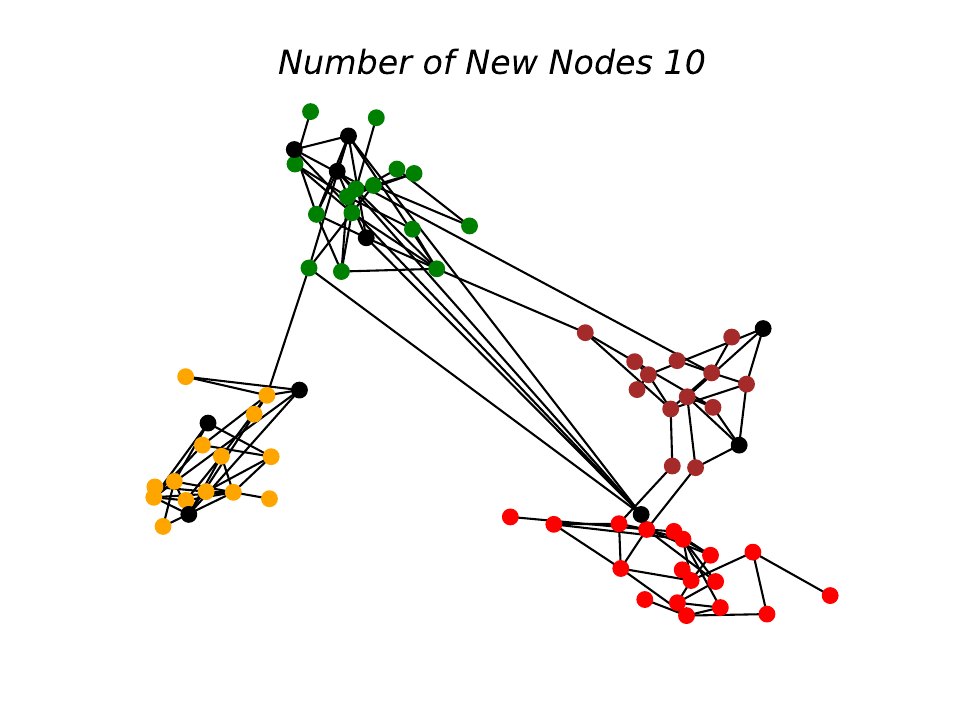}
    \captionsetup{font=scriptsize, justification=centering}
    \caption{10 incoming nodes}
  \end{subfigure}%
  \hfill%
  \begin{subfigure}[t]{.15\linewidth}
    \includegraphics[width=\linewidth, trim={2.3cm 1.4cm 1.9cm 1.6cm}, clip]{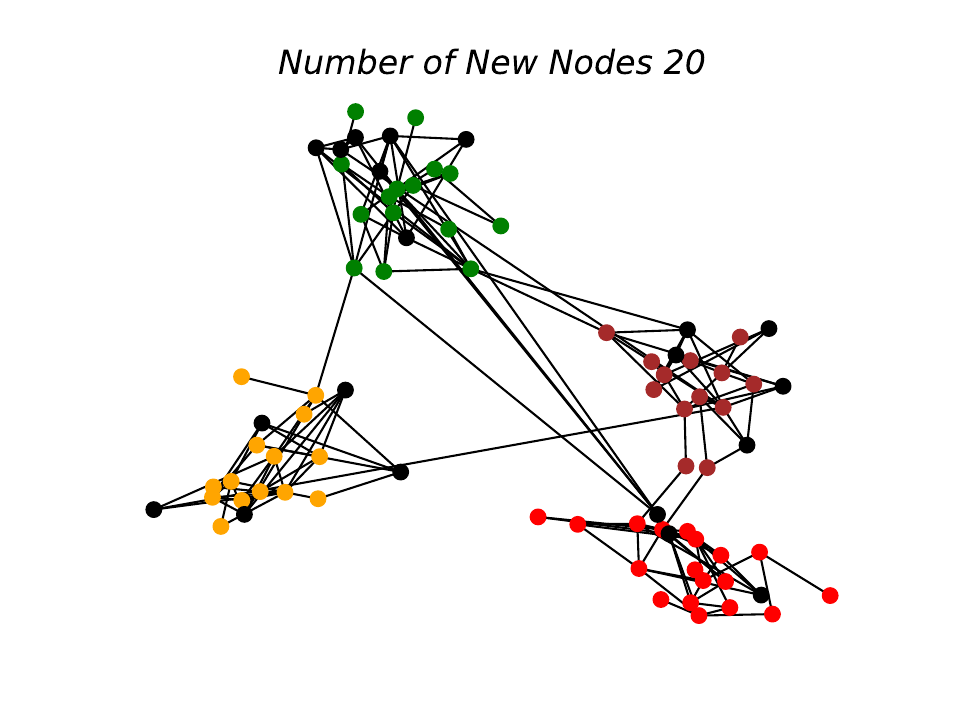}
    \captionsetup{font=scriptsize, justification=centering}
    \caption{20 incoming nodes}
  \end{subfigure}%
  \hfill%
  \begin{subfigure}[t]{.15\linewidth}
    \includegraphics[width=\linewidth, trim={2.3cm 1.4cm 1.9cm 1.6cm}, clip]{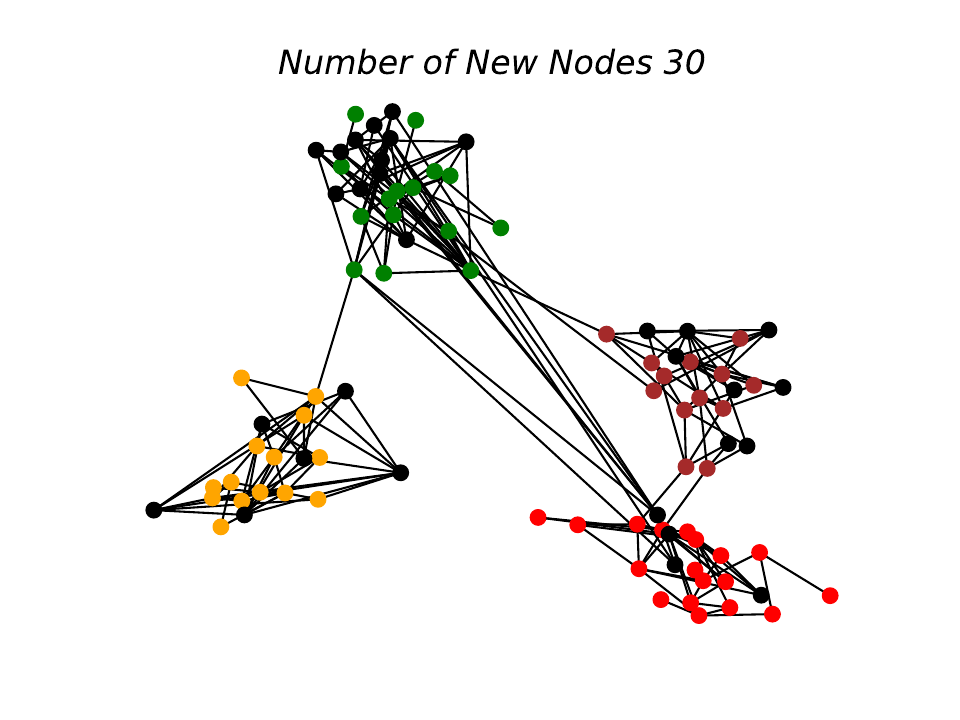}
    \captionsetup{font=scriptsize, justification=centering}
    \caption{30 incoming nodes}
  \end{subfigure}
  \hfill%
  \hfill%
  \begin{subfigure}[t]{.15\linewidth}
    \includegraphics[width=\linewidth, trim={2.3cm 1.4cm 1.9cm 1.6cm}, clip]{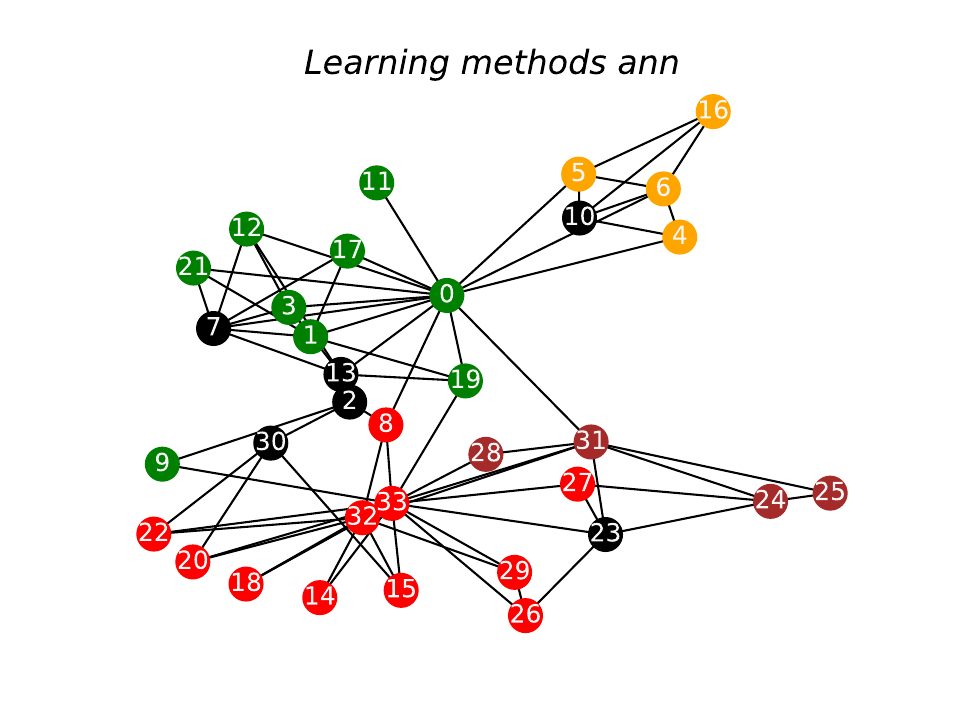}
    \captionsetup{font=scriptsize, justification=centering}
    \caption{ANN as $\mathcal{M}_{gl}$}
  \end{subfigure}%
  \hfill%
  \begin{subfigure}[t]{.15\linewidth}
    \includegraphics[width=\linewidth, trim={2.3cm 1.4cm 1.9cm 1.6cm}, clip]{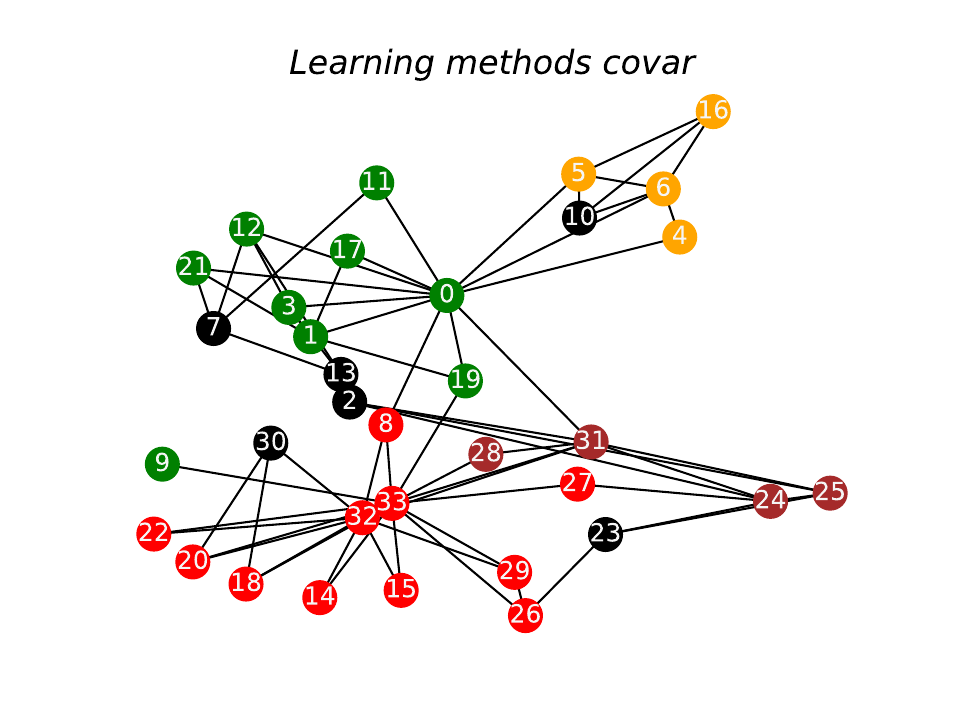}
    \captionsetup{font=scriptsize, justification=centering}
    \caption{Emp. Covr. as $\mathcal{M}_{gl}$}
  \end{subfigure}%
  \hfill%
  \begin{subfigure}[t]{.15\linewidth}
    \includegraphics[width=\linewidth, trim={2.3cm 1.4cm 1.9cm 1.6cm}, clip]{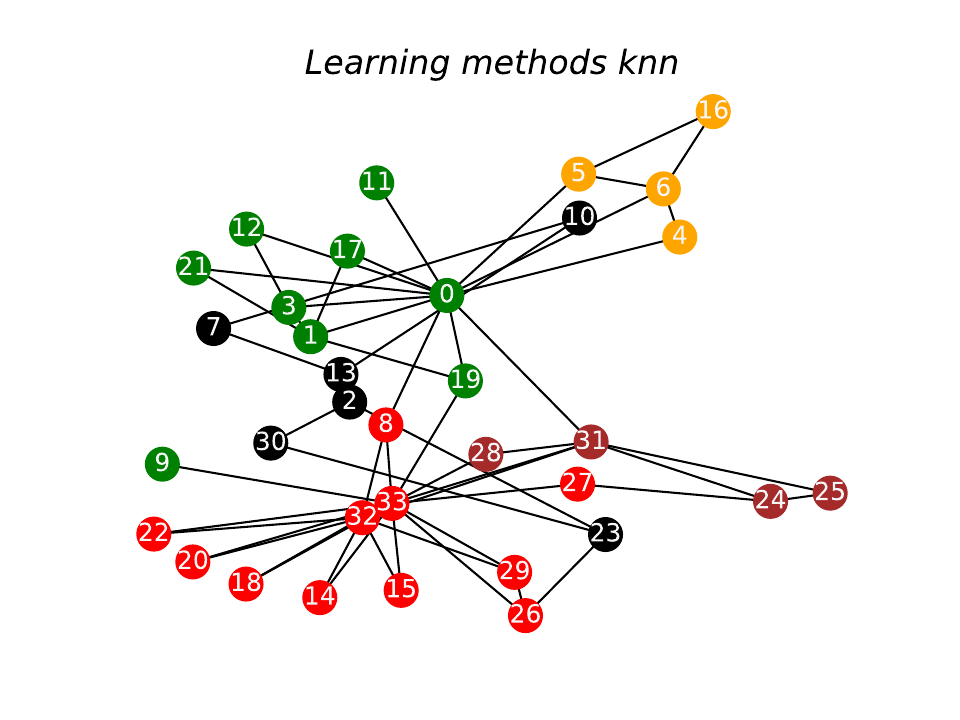}
    \captionsetup{font=scriptsize, justification=centering}
    \caption{kNN as $\mathcal{M}_{gl}$}
  \end{subfigure}
\end{center}
\caption{Figures (a), (b), and (c) illustrate the growing structure learned using GraphFLEx for \textit{HE} synthetic dataset. Figures (d), (e), and (f) illustrate the learned structure on Zachary’s karate dataset when existing methods are employed with GraphFLEx. New nodes are denoted using black color.}
\label{fig:growing_and_karate}
\vskip -0.8cm
\end{figure*}
\subsection{Structure Visualization}
\label{sec:struct_visul}
\begin{wrapfigure}{r}{0.50\textwidth}
\vspace{-0.8cm}
  \centering
  \includegraphics[width=0.23\textwidth]{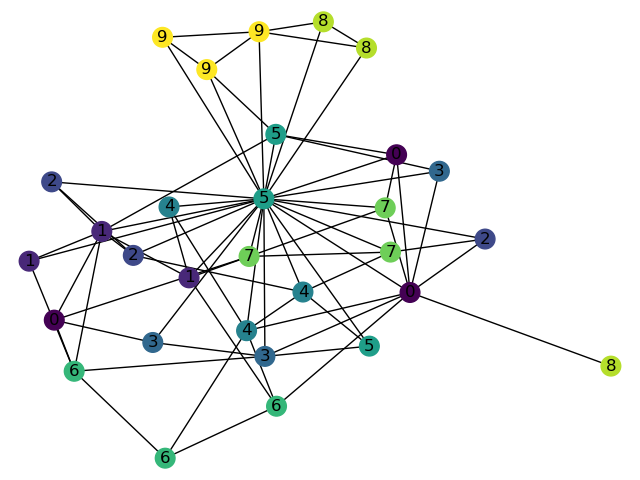}
  \includegraphics[width=0.26\textwidth]{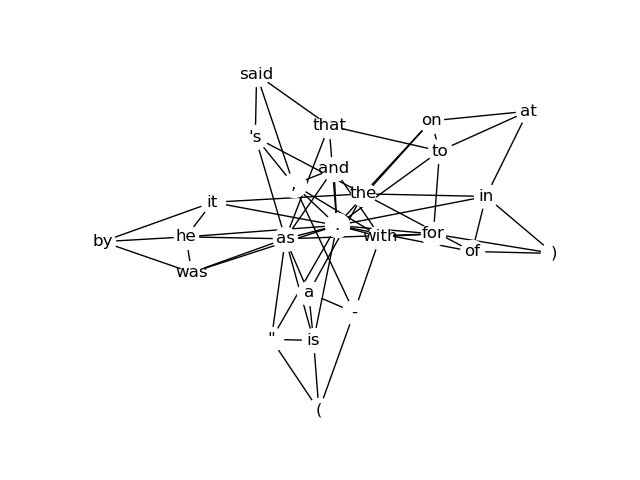}
  \caption{Effectiveness of our framework in learning structure between similar MNIST digits and GloVe embeddings.}
  \label{fig:vis_mnist_words}
\vspace{-0.5cm}
\end{wrapfigure}
We evaluate the structures generated by GraphFLEx through visualizations on four small datasets: (i) MNIST~\cite{lecun2010mnist}, consisting of handwritten digit images, where Figure~\ref{fig:vis_mnist_words}(a) shows that images of the same digit are mostly connected; (ii) Pre-trained GloVe embeddings~\cite{pennington2014glove} of English words, with Figure~\ref{fig:vis_mnist_words}(b) revealing that frequently used words are closely connected; (iii) A synthetic $H.E$ dataset (see Appendix~\ref{app:datasets}), demonstrating GraphFLEx’s ability to handle expanding networks without requiring full relearning. Figure~\ref{fig:growing_and_karate}(a-c) shows the graph structure evolving as 30 new nodes are added over three timestamps; and (iv) Zachary's karate club network~\cite{zachary1977information}, which highlights GraphFLEx’s multi-framework capability. Figure~\ref{fig:growing_and_karate}(d-f) shows three distinct graph structures after altering the learning module. For a comprehensive ablation study, refer to Appendix~\ref{app:ablation}.

%% file: chapters/conclusion.tex
\section{Conclusion}
\label{sec:conclusion}
Large or expanding graphs challenge the best of graph learning approaches. GraphFLEx, introduced in this paper, seamlessly adds new nodes into an existing graph structure. It offers diverse methods for acquiring the graph's structure. GraphFLEx consists of three key modules: Clustering, Coarsening, and Learning which empowers GraphFLEx to serves as a comprehensive framework applicable individually for clustering, coarsening, and learning tasks. Empirically, GraphFLEx outperforms state-of-the-art baselines, achieving up to 3× speedup while preserving structural quality. It achieves accuracies close to training on the original graph, in most instances. The performance across multiple real and synthetic datasets affirms the utility and efficacy of GraphFLEx for graph structure learning.\\
\textbf{Limitations and Future Work.}  GraphFLEx is designed assuming minimal inter-community connectivity, which aligns well with many real-world scenarios. However, its applicability to heterophilic graphs may require further adaptation. Future work will focus on extending the framework to supervised GSL methods and heterophilic graphs, broadening its scalability and versatility.

%% file: chapters/appendix.tex
\clearpage
\begin{center}
    \Large{\textbf{Appendix}}
\end{center}

\section{Degree-Corrected Stochastic Block Model(DC-SBM)}
\label{app:dc_sbm}
The DC-SBM is one of the most commonly used models for networks with communities and postulates that, given node labels $\mathbf{c} = {c_1,...c_n}$, the edge variables $A_{ij}'s$ are generated via the formula $$E[A_{ij}] = \theta_i \theta_j P_{c_{i}} P_{c_{j}}$$, where $\theta_i$ is a "degree parameter" associated with node $i$, reflecting its individual propernsity to form ties, and $P$ is a $K \times K$ symmetric matrix  containing the between/withincommunity edge probabilities and $P_{c_{i}} P_{c_{j}}$ denotes the edge probabilities between community $c_i$ and $c_j$.\\
For DC-SBM model \cite{zhao2012consistency} assumed $P_n$ on $n$ nodes with $k$ classes, each node $v_i$ is given a label/degree pair($c_i, \theta_i$), drawn from a discrete joint distribution $\Pi_{K\times m}$ which is fixed and does not depend on n. This implies that each $\theta_i$ is one of a fixed set of values $0 \leq x_1 \leq....\leq x_m$. To facilitate analysis of asymptotic graph sparsity, we parameterize the edge probability matrix $P$ as $P_n = \rho_n P$ where P is independent of $n$, and $\rho_n = \lambda_n/n$ where $\lambda_n$ is the average degree of the network.\\
\section{Neighbourhood Preservation}
\label{app:neigh_preserv}
\begin{theorem}
\textbf{Neighborhood Preservation.} Let \(\mathcal{N}_k(\mathcal{E}_i)\) denote the neighborhood of incoming nodes \(\mathcal{E}_i\) for the \(i^{\text{th}}\) community. With partition matrix $\mathcal{P}_i$ and $\mathcal{M}_{gl}(S_i,X_\tau^i) = \mathcal{G}_\tau^c(V^c_\tau,A^c_\tau)$ we identify the supernodes connected to incoming nodes $\mathcal{E}_i$ and subsequently select nodes within those supernodes; this subset of nodes is denoted by $\omega_{V_\tau^i}$. Formally, 
\small
\[
\omega_{V_\tau^i} = \bigcup_{v \in \mathcal{E}_i}  
\Bigl\{ \bigcup_{s\in S_i}\{\pi^{-1}(s)|A^c_\tau(v,s) \neq 0\}\Bigl\}
\]
Then, with probability \(\Pi_{\{c \in \phi\}} p(c)\), it holds that $\mathcal{N}_k(\mathcal{E}_i) \subseteq \omega_{V_\tau^i}$ where
\[
p(c) \leq 1 - \frac{2}{\sqrt{2 \pi}} \frac{c}{r}\left[1-e^{-r^2 /\left(2 c^2\right)}\right],
\]
and \(\phi\) is a set containing all pairwise distance values \((c = \|v - u\|)\) between every node \(v \in \mathcal{E}_i\) and the nodes \(u \in \omega_{V_\tau^i}\). Here, $\pi^{-1}(s)$ denotes the set of nodes mapped to supernode s, \(r\) is the bin-width hyperparameter of \(\mathcal{M}_{\text{coar}}\).
\end{theorem}

\textbf{Proof:} The probability that LSH random projection \cite{kataria2024ugc,datar2004locality} preserves the distance between two nodes $v$ and $u$ i.e., $d(u,v)=c$, is given by: 
\[
p(c) = \int_{0}^{r} \frac{1}{c} f_2 \left( \frac{t}{c} \right) \left( 1 - \frac{t}{r} \right) dt,
\]
where \( f_2(x) = \frac{2}{\sqrt{2\pi}} e^{-x^2/2} \) represents the Gaussian kernel when the projection matrix is randomly sampled from $p$-stable($p=2$) distribution \cite{datar2004locality}. \\
The probability \( p(c) \) can be decomposed into two terms:
\[
p(c) = S_1(c) - S_2(c),
\]
$S_1(c)$ and $S_2(c)$ are defined as follows:
\[
S_1(c) = \frac{2}{\sqrt{2\pi}} \int_{0}^{r} e^{-(t/c)^2/2} dt \leq 1,
\]
\[
S_2(c) = \frac{2}{\sqrt{2\pi}} \int_{0}^{r} e^{-(t/c)^2/2} \frac{t}{r} dt.
\]
\[
S_2(c) = \frac{2}{\sqrt{2\pi}} \cdot \frac{c}{r} \int_{0}^{r} e^{-(t/c)^2/2} \frac{t}{c^2} dt
\]
Expanding $S_2(c):$
\[
S_2(c) = \frac{2}{\sqrt{2\pi}} \cdot \frac{c}{r} \int_{0}^{r^2/(2c^2)} e^{-y} dy
\]
\[
S_2(c) = \frac{2}{\sqrt{2\pi}} \cdot \frac{c}{r} \left[ 1 - e^{-r^2/(2c^2)} \right]
\]
Thus, the probability \( p(c) \) can be bounded as:
\[
p(c) \leq 1 - \frac{2}{\sqrt{2\pi}} \frac{c}{r} \left[ 1 - e^{-r^2/(2c^2)} \right].
\]

Now, let \( \phi \) be the set of all pairwise distances $d(u,v)$, where \( v \in \mathcal{E}_i \) and node\( \omega_{V_\tau^i} \). The probability that all nodes in \( \mathcal{N}_k(\mathcal{E}_i) \) are preserved within \( \omega_{V_\tau^i} \), requires that all distances $c\in \phi$ are also preserved. The probability is then given by:
\[
\prod_{c \in \phi} p(c).
\]

\[
\prod_{c \in \phi} p(c) \leq \prod_{c \in \phi} \left( 1 - \frac{2}{\sqrt{2\pi}} \frac{c}{r} \left[ 1 - e^{-r^2/(2c^2)} \right] \right).
\]

\section{Continual Learning and Dynamic Graph Learning}
\label{CL-DGL}
In this subsection, we highlight the key distinctions between Graph Structure Learning (GSL) and related fields to justify our specific selection of related works in Section \ref{sec:related_work}. GSL is often confused with topics such as Continual Learning (CL) and Dynamic Graph Learning (DGL).\\
CL \cite{van2019three, zhang2022cglb, parisi2019continual} addresses the issue of catastrophic forgetting, where a model's performance on previously learned tasks degrades significantly after training on new tasks. In CL, the model has access only to the current task's data and cannot utilize data from prior tasks. Conversely, DGL \cite{kim2022dygrain, wu2023continual, you2022roland} focuses on capturing the evolving structure of graphs and maintaining updated graph representations, with access to all prior information.\\
While both \textit{CL and DGL} aim to \textit{enhance model adaptability} to dynamic data, GSL is primarily concerned with generating \textit{high-quality graph structures} that can be leveraged for downstream tasks such as node classification \cite{kipf2016semi}, link prediction \cite{lu2011link}, and graph classification \cite{vogelstein2012graph}. Moreover, in CL and DGL, different tasks typically involve distinct data distributions, whereas GSL assumes a consistent data distribution throughout.

\section{Related Work}
\label{app:related_work}
Table~\ref{t:graph_learning_methods} presents the formulations and associated time complexities of various unsupervised Graph Structure Learning methods.
\captionsetup{font=footnotesize}
\begin{table}[ht]
    \caption{Unsupervised Graph Structure Learning Methods}
    \centering
    \footnotesize
    \setlength{\tabcolsep}{1mm}
    \begin{tabular}{lcc}
        \toprule
        \textbf{Method} & \textbf{Time Complexity} & \textbf{Formulation} \\
        \midrule
        $GLasso$ & $O(N^3)$ & $\max_{\Theta} \log\det \Theta - \text{tr}(\hat{\Sigma}\Theta) - \rho\|\Theta\|_1$ \\
        \midrule
        $log$-model & $O(N^2)$ & $\min_{W \in \mathcal{W}} \|W \circ Z\|_{1,1} - \alpha \mathbf{1}^T \log(W\mathbf{1}) + \frac{\beta}{2}\|W\|_F^2$ \\
        \midrule
        $l2$-model & $O(N^2)$ & $\min_{W \in \mathcal{W}} \|W \circ Z\|_{1,1} + \alpha\|W\mathbf{1}\|^2 + \alpha\|W\|_F^2 + \mathbf{1}\{\|W\|_{1,1}=n\}$ \\
        \midrule
        $large$-model & $O(N\log(N))$ & $\min_{W \in \tilde{W}} \|W \circ Z\|_{1,1} -\alpha \mathbf{1}^T \log(W\mathbf{1}) + \frac{\beta}{2}\|W\|_F^2$ \\
        \bottomrule
    \end{tabular}
    \label{t:graph_learning_methods}
    \vskip -0.09in
\end{table}

\section{Run Time Analysis}
\label{app:run_time}
In the context of clustering module, $k-NN$ is the fastest algorithm, while Spectral Clustering is the slowest.  
Suppose we aim to learn the structure of a graph with $N$ nodes. The clustering module, however, is only applied to a randomly sampled, smaller, static subgraph with $k$ nodes, where $k \ll N$.  In the worst-case scenario, spectral clustering requires $\mathcal{O}(k^3)$ time, whereas in the best case, $k-NN$ requires $\mathcal{O}(k^2)$ time. For coarsening module, LSH-based coarsening framework \cite{kataria2023linear}, has the best time complexity of $\mathcal{O}(\frac{k_\tau}{c})$ while FGC denotes the worst case with a time-complexity of $\mathcal{O}((\frac{k_\tau}{c})^2 \|S_\tau^i\|)$ where $c$ is the number of communities detected by clustering module $\mathcal{M}_{\text{clust}}$, $\|S_\tau^i\|$ is the number of coarsened node in the relevant community at $\tau$ timestamp and $k_\tau$ denotes number of nodes at $\tau$ timestamp. For learning module,  $A-NN$ is the most efficient algorithm with time complexity as $\mathcal{O}(NlogN)$, while $GLasso$ has the worst computational cost of $\mathcal{O}(N^3)$. So, the effective time complexity of GraphFLEx is upper bounded by $\mathcal{O}(k^3 + (\frac{k_\tau}{c})^2 \|S_\tau^i\| + \alpha^3)$ and lower bounded by $\mathcal{O}(k^2 + \frac{k_\tau}{c} + \alpha log\alpha)$ where $\alpha = \|S_\tau^i\| + \|\mathcal{E}^i_\tau\|$. GraphFLEx's efficiency in term of computational time is evident in Figure~\ref{fig:expo_time} and further quantified in Table~\ref{t:time}.\\
Out of the three modules of GraphFLEx first module($\mathcal{M}_{\text{clust}}$) is trained once, and hence its run time is always bounded; computational time for second module($\mathcal{M}_{\text{coar}}$) can also be controlled because some of the methods either needs training once \cite{jin2021graph} or have linear time complexity \cite{kataria2023linear}. Consequently, both the clustering and coarsening modules contribute linearly to the overall time complexity, denoted as $\mathcal{O}(N)$.
Thus, the effective time complexity of GraphFLEx is given by $\mathcal{O}(N + \mathcal{O}(\mathcal{M}_{gl}(\|S_i, X^i_\tau\|))$. The overall complexity scales either linearly or sub-linearly, depending on $\alpha$ and the $\mathcal{M}_{gl}$ module. For instance, when $\mathcal{M}_{gl}$ is $A$-NN the complexity remains linear, if $\alpha \log(\alpha) \approx N$, whereas for $GLasso$, a linear behavior is observed when $\alpha^3 \approx N$.

\section{Datasets}
\label{app:datasets}
\begin{figure}[ht]
\vskip 0.2in
\begin{center}
  \vspace{1mm} 
  
  \begin{subfigure}[t]{.23\linewidth}
    \includegraphics[width=\linewidth, trim={2.3cm 1.4cm 1.9cm 1.6cm}, clip]{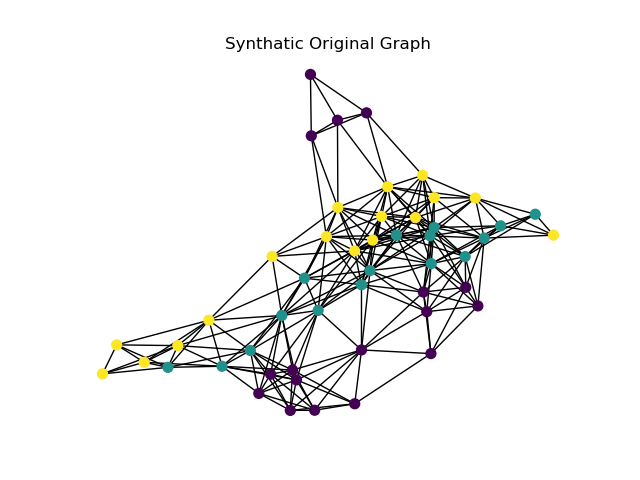}
    \captionsetup{font=scriptsize, justification=centering}
    \caption{PyGSP-Sensor, N = 50, $\alpha$=3}
  \end{subfigure}
  \hfill
  \begin{subfigure}[t]{.23\linewidth}
    \includegraphics[width=\linewidth, trim={2.3cm 1.4cm 1.9cm 1.6cm}, clip]{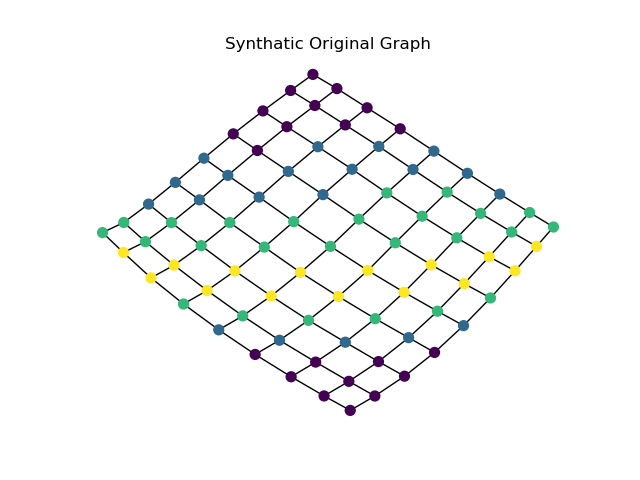}
    \captionsetup{font=scriptsize, justification=centering}
    \caption{PyGSP-Grid, N = 80, $\alpha$=3}
  \end{subfigure}
  \hfill
  \begin{subfigure}[t]{.23\linewidth}
    \includegraphics[width=\linewidth, trim={2.3cm 1.4cm 1.9cm 1.6cm}, clip]{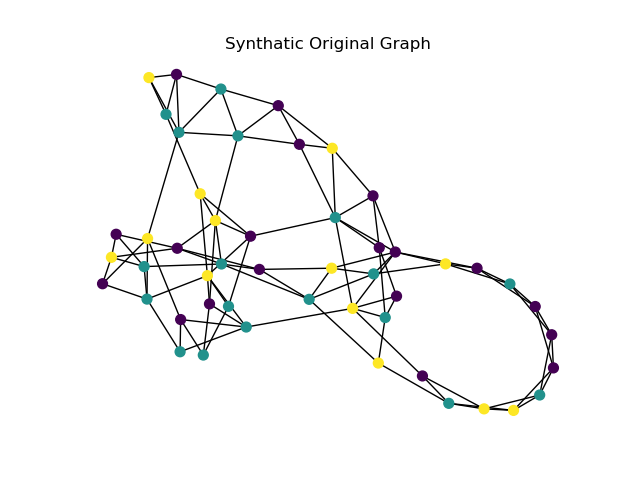}
    \captionsetup{font=scriptsize, justification=centering}
    \caption{SW, N = 50, $\alpha$=3}
  \end{subfigure}
  \hfill
  \begin{subfigure}[t]{.23\linewidth}
    \includegraphics[width=\linewidth, trim={2.3cm 1.4cm 1.9cm 1.6cm}, clip]{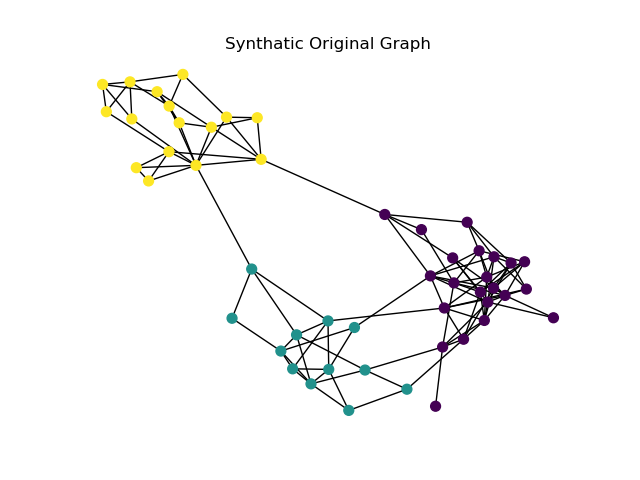}
    \captionsetup{font=scriptsize, justification=centering}
    \caption{HE, N = 50, $\alpha$=3}
  \end{subfigure}
  \hfill
\end{center}
\caption{This figure illustrates different types of synthetic graphs generated using i)PyGSP, ii) Watts–Strogatz's small world(SW), and iii) Heterophily(HE). N denotes the number of nodes, while $\alpha$ denotes the number of classes.}
\label{fig:syn_graphs}
\vskip -0.2in
\end{figure}

Datasets used in our experiments vary in size, with nodes ranging from 1k to 60k. Table~\ref{t:datasets} lists all the datasets we used in our work. We evaluate our proposed framework $GraphFlex$ on real-world datasets $\textit{Cora ,Citeseer, Pubmed}$ \cite{yang2016revisiting}, $\textit{CS, Physics}$ \citep{shchur2018pitfalls}, $\textit{DBLP}$ \citep{fu2020magnn}, all of which include graph structures. 
These datasets allow us to compare the learned structures with the originals. Additionally, we utilize single-cell RNA pancreas datasets ~\cite{yang2022scbert}, including Baron, Muraro, Segerstolpe, and Xin, where the graph structure is missing. The Baron dataset was downloaded from the Gene Expression Omnibus (GEO) (accession no. GSE84133). The Muraro dataset was downloaded from GEO (accession no. GSE85241). The Segerstolpe dataset was accessed from ArrayExpress (accession no. E-MTAB-5061). The Xin dataset was downloaded from GEO (accession no. GSE81608). We simulate the expanding graph scenario by splitting the original dataset across different $\mathcal{T}$ timestamps. 
We assumed 50\% of the nodes were static, with the remaining nodes arriving as incoming nodes at different timestamps.\\
\textbf{Synthetic datasets:} Different data generation techniques validate that our results are generalized to different settings. Please refer to Table~\ref{t:datasets} for more details about the number of nodes, edges, features, and classes, $Syn$ denotes the type of synthetic datasets. Figure~\ref{fig:syn_graphs} shows graphs generated using different methods. We have employed three different ways to generate synthetic datasets which are mentioned below:
\begin{itemize}
    \item \textbf{PyGSP(PyGsp): } We used synthetic graphs created by PyGSP \cite{pygsp} library. PyG-G and PyG-S denotes grid and sensor graphs from PyGSP. 
    \item \textbf{Watts–Strogatz's small world(SW): } \cite{watts1998collective} proposed a generation model that produces graphs with small-world properties, including short average path lengths and high clustering.
    \item \textbf{Heterophily(HE): } We propose a method for creating synthetic datasets to explore graph behavior across a heterophily spectrum by manipulating heterophilic factor $\alpha$, and classes. $\alpha$ is determined by dividing the number of edges connecting nodes from different classes by the total number of edges in the graph.
\end{itemize}
\textbf{Visulization Datasets:} To evaluate, the learned graph structure, we have also included three datasets: (i) MNIST~\cite{lecun2010mnist}, consisting of handwritten digit images; (ii) Pre-trained GloVe embeddings~\cite{pennington2014glove} of English words; and (iii) Zachary's karate club network~\cite{zachary1977information}.\\
\textbf{Large Datasets:} To comprehensively evaluate GraphFLEx’s scalability to large-scale graphs, we consider four datasets with a high number of nodes: (a) \textit{Flickr(89k nodes)} \cite{zeng2019graphsaint}, (b) \textit{Reddit (233k nodes)} \cite{zeng2019graphsaint}, (c) \textit{Ogbn-arxiv (169k nodes)} \cite{wang2020microsoft}, and (d) \textit{Ogbn-products (2.4M nodes)} \cite{Bhatia16}.\\

\noindent
\textit{System Specifications:} All the experiments conducted for this work were performed on an Intel Xeon W-295 CPU with 64GB of RAM desktop using the Python environment.



\begin{table*}[ht]
\centering
\setlength{\tabcolsep}{4pt} 
\resizebox{\textwidth}{!}{
\begin{tabular}{c c c c c c c}
\toprule
Category & Data & Nodes & Edges & Feat. & Class & Type \\
\midrule
\multirow{7}{*}{\begin{tabular}{@{}c@{}}Original\\Structure\\Known\end{tabular}} & Cora & 2,708 & 5,429 & 1,433 & 7 & Citation network \\
& Citeseer & 3,327 & 9,104 & 3,703 & 6 & Citation network \\
& DBLP & 17,716 & 52.8k & 1,639 & 4 & Research paper \\
& CS & 18,333 & 163.7k & 6,805 & 15 & Co-authorship network \\
& PubMed & 19,717 & 44.3k & 500 & 3 & Citation network \\
& Physics & 34,493 & 247.9k & 8,415 & 5 & Co-authorship network\\
\midrule
\multirow{5}{*}{\begin{tabular}{@{}c@{}}Original\\Structure\\Not Known\end{tabular}} & Xin & 1,449 & NA & 33,889 & 4 & Human Pancreas \\
& Baron Mouse & 1,886 & NA & 14,861 & 13 & Mouse Pancreas \\
& Muraro & 2,122 & NA & 18,915 & 9 & Human Pancreas \\
& Segerstolpe & 2,133 & NA & 22,757 & 13 & Human Pancreas \\
& Baron Human & 8,569 & NA & 17,499 & 14 & Human Pancreas \\
\midrule
\multirow{8}{*}{\begin{tabular}{@{}c@{}}Synthetic\end{tabular}} & Syn 1 & 2,000 & 8,800 & 150 & 4 & SW \\
& Syn 2 & 5,000 & 22k & 150 & 4 & SW \\
& Syn 3 & 10,000 & 44k & 150 & 7 & SW \\
& Syn 4 & 50,000 & 220k & 150 & 7 & SW \\
& Syn 5 & 400 & 1,520 & 100 & 4 & PyG-G \\
& Syn 6 & 2,500 & 9,800 & 100 & 4 & PyG-S \\
& Syn 7 & 1,000 & 9,990 & 150 & 4 & HE \\
& Syn 8 & 2,000 & 40k & 150 & 4 & HE \\
\midrule
\multirow{3}{*}{\begin{tabular}{@{}c@{}}Visulization Datasets\end{tabular}} & MNIST & 60,000 & NA & 784 & 10 & Images \\
& Zachary’s karate & 34 & 156 & 34 & 4 & Karate club network \\
& Glove & 2,000 & NA & 50 & NA & GloVe embeddings \\
\midrule
\multirow{3}{*}{\begin{tabular}{@{}c@{}}Large dataset\end{tabular}} & Flickr & 89,250 & 899,756 & 500 & 7 & - \\
& Reddit & 232,965 & 11.60M & 602 & 41 & - \\
& Ogbn-arxiv & 169,343 & 1.16M & 128 & 40 & - \\
& Ogbn-products & 2,449,029 & 61.85M & 100 & 47 & - \\
\bottomrule
\end{tabular}}
\caption{Summary of the datasets.}
\label{t:datasets}
\end{table*}

\section{Algorithm}
\label{app:algo}
\begin{algorithm}[ht]
    \caption{GraphFlex: A Unified Structure Learning framework for expanding and Large Scale Graphs}
    \label{alg:algorithm}
    \textbf{Input}: Graph $G_0(X_0,A_0)$, expanding nodes set $\mathcal{E}_1^T = \mathcal{\{E_\tau(V_\tau,X_\tau)}\}^T_{\tau=1}$ \\
    \textbf{Parameter}: GClust, GCoar, GL $\leftarrow$ Clustering, Coarsening and Learning  Module\\
    \textbf{Output}: Graph $G_T(X_T,A_T)$
    \begin{algorithmic}[1] 
        \STATE Train clustering module \textit{train}($\mathcal{M}_{clust}$, GClust, $G_0$) 
        \FOR{each $E_t(V_t,X_t)$ in $\mathcal{E}_1^T$}
        \STATE $C_t = \textit{infer}(\mathcal{M}_{clust},X_t)$, $C_t \in \mathbb{R}^{N_t}$ denotes the communities of $N_t$ nodes at time $t$.
        \STATE $I_{t} = \textit{unique}(C_t)$.
        \FOR{each $I^i_t$ in $I_t$}
            \STATE $G^i_{t-1}$ = subgraph($G_{t-1}$, $I^i_t$)
            \STATE $\{S^i_{t-1}, P^i_{t-1}\} = \mathcal{M}_{coar}(G^i_{t-1})$, $S^i_{t-1} \in \mathbb{R}^{k \times d}$ are features of $k$ supernodes, $P^i_{t-1} \in \mathbb{R}^{k \times N^i_t}$ is the partition matrix.
            \STATE $Gc^i_{t-1}(S^i_{t-1},A^i_{t-1}) = \mathcal{M}_{gl}(S^i_{t-1}, X^i_{t})$, $Gc^i_{t-1}$ is the learned graph on super-nodes $S^i_{t-1}$ and new node $X^i_t$.
            \STATE $\omega^i_t \leftarrow$ \texttt{[]}
            \FOR{$x \in X^i_t$}
            \STATE $\omega^i_t.append(x)$
            \STATE $n_p=\{n\mid A^i_{t-1}[n] > 0\}$
            \STATE $\omega^i_t.append(n_p)$            
            \ENDFOR
            \STATE $G_{t-1} = \textit{update}(G_{t-1},\mathcal{M}_{gl}(\omega^i_t))$
        \ENDFOR
        \STATE $G_t = G_{t-1}$
        \ENDFOR
        \STATE \textbf{return} $G_T(X_T,A_T)$
    \end{algorithmic}
\end{algorithm}

\section{Other GNN models}
\label{app:model_parameter}
We used four GNN models, namely GCN, GraphSage, GIN, and GAT. Table~\ref{t:gnn_para} contains parameter details we used to train GraphFlex. We have used these parameters across all methods.

\begin{figure}[ht]
\begin{center}
\centerline{\includegraphics[width=1\columnwidth, trim={0cm 1cm 0cm 0cm}, clip]{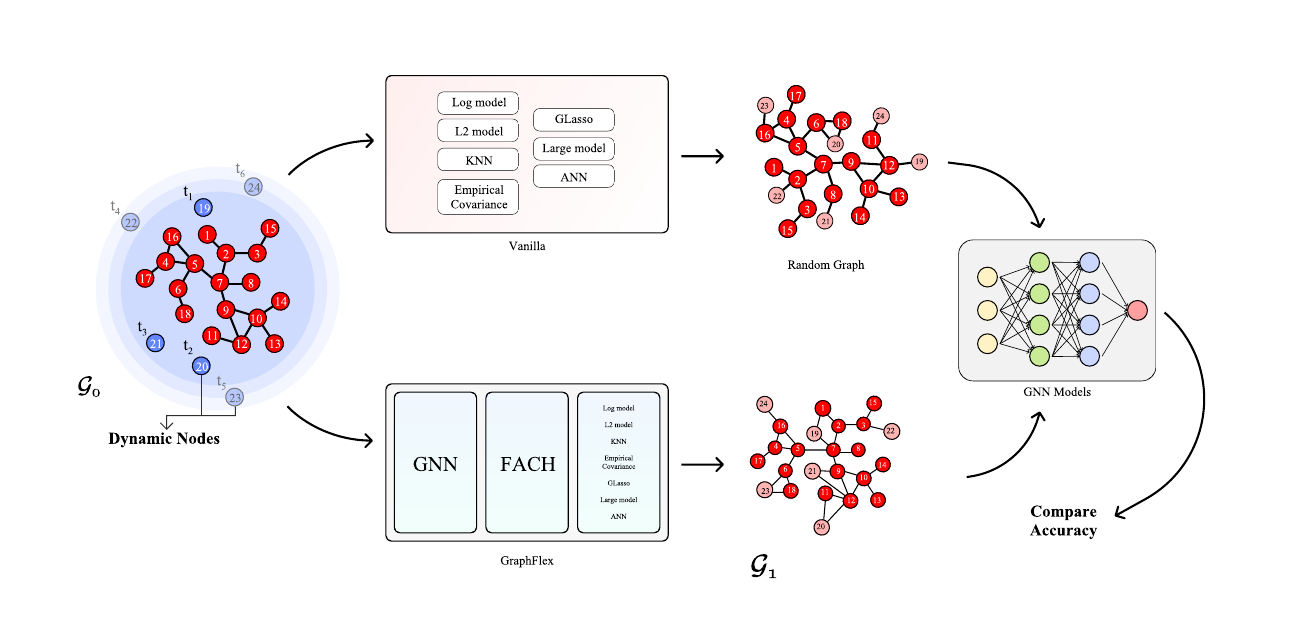}}
\caption{GNN training pipeline.}
\label{fig:gnn_pipeline}
\end{center}
\vskip -0.35in
\end{figure}

\begin{figure}[ht]
\begin{center}
\centerline{\includegraphics[width=1\columnwidth]{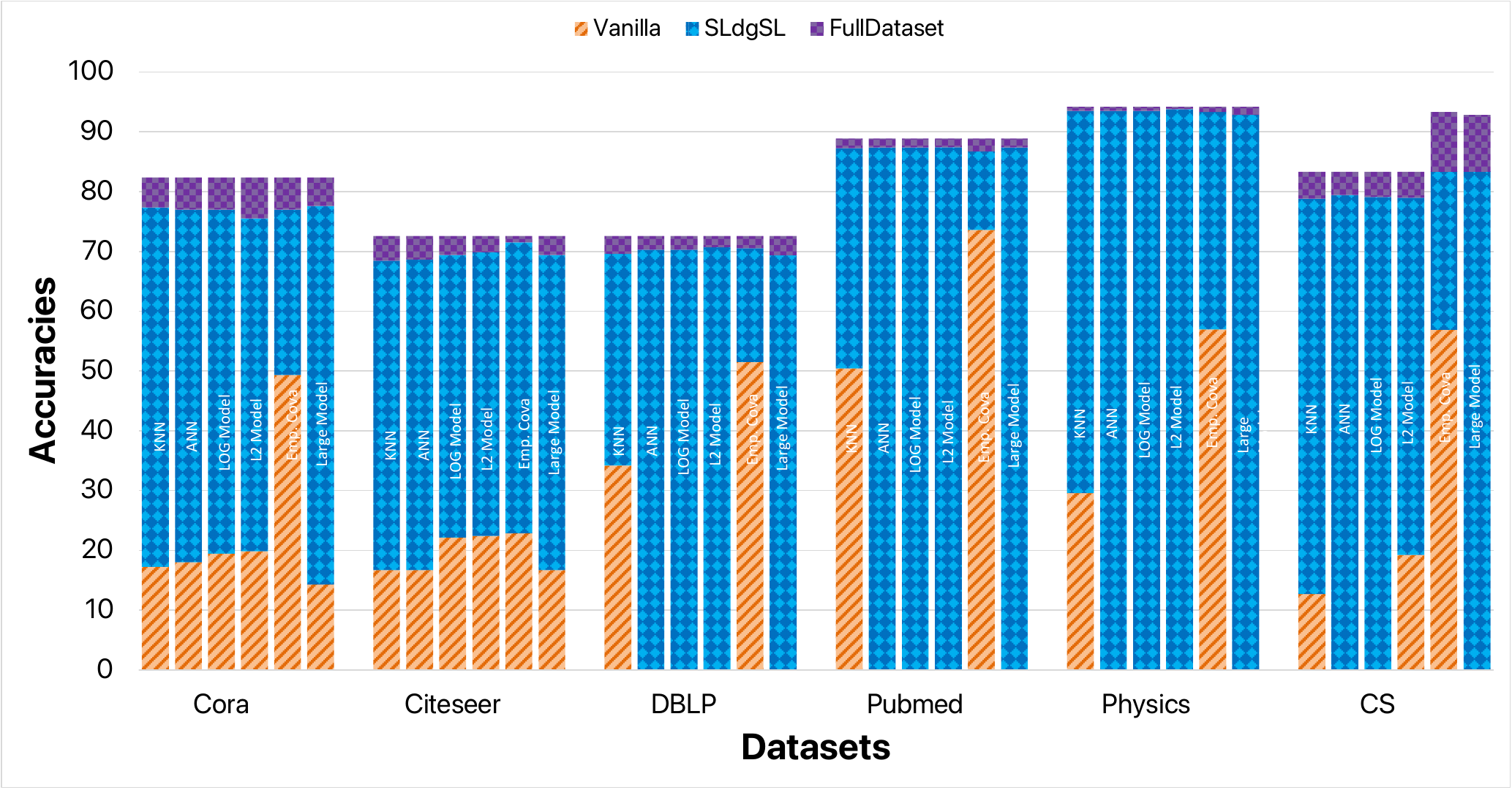}}
\caption {GraphSage accuracies when structure is learned or given with 3 different scenarios(Vanilla, GraphFlex, Original) across different datasets, highlighting performance with 30\% node growth over 25 timestamps.}
\label{fig:sage_acc}
\end{center}
\end{figure}

Figure~\ref{fig:gnn_pipeline} illustrates the pipeline for training our GNN models. Graph structures were learned using both existing methods and GraphFlex, and GNN models were subsequently trained on both structures. Results across all datasets are presented in Table~\ref{t:_syn_gnn_accuracies} and Table~\ref{t:gnn_accuracies}.

\begin{table}[ht]
\centering
\caption{GNN model parameters.}
\begin{tabular}{lccccr}
\toprule
Model & Hidden Layers & L.R & Decay & Epoch\\
\midrule
GCN & $\{{64,64\}}$& 0.003 & 0.0005 & 500\\
GraphSage & $\{{64,64\}}$& 0.003 & 0.0005 & 500\\
GIN & $\{{64,64\}}$& 0.003 & 0.0005 & 500\\
GAT & $\{{64,64\}}$& 0.003 & 0.0005 & 500\\
\bottomrule
\end{tabular}
\label{t:gnn_para}
\end{table}
We randomly split data in 60\%, 20\%, 20\% for training-validation-test.
The results for these models on synthetic datasets are presented in Table~\ref{t:_syn_gnn_accuracies}.

Figure~\ref{fig:gnn_pipeline} illustrates the pipeline for training our GNN models. Graph structures were learned using both existing methods and GraphFlex, and GNN models were subsequently trained on both structures.

\begin{table*}[!ht]
    \centering
    \setlength{\tabcolsep}{1mm}
    \caption{Node classification accuracies on different GNN models using GraphFLEx (GFlex) with existing Vanilla (Van.) methods. The experimental setup involves treating 70\% of the data as static, while the remaining 30\% of nodes are treated as new nodes coming in 25 different timestamps. The best and the second-best accuracies in each row are highlighted by dark and lighter shades of \textcolor{green}{Green}, respectively. GraphFLEx's structure beats all of the vanilla structures for every dataset. OOM and OOT denotes out-of-memory and out-of-time respectively.}
    \resizebox{\textwidth}{!}{
    \begin{tabular}{lrrrrrrrrrrrrrc}
    \toprule
    \textbf{Dataset} & \bfseries Model & \multicolumn{2}{c}{\bfseries ANN} & \multicolumn{2}{c}{\bfseries KNN} & \multicolumn{2}{c}{\bfseries log-model} & \multicolumn{2}{c}{\bfseries l2-model} & \multicolumn{2}{c}{\bfseries COVAR} & \multicolumn{2}{c}{\bfseries large-model} & \bfseries Base Struc. \\
            &       & \bfseries Van. & \bfseries GFlex & \bfseries Van. & \bfseries GFlex & \bfseries Van. & \bfseries GFlex & \bfseries Van. & \bfseries GFlex & \bfseries Van. & \bfseries GFlex & \bfseries Van. & \bfseries GFlex &  \\ 
            \midrule
            & \gat & $18.73$ & $73.84$ & $20.96$ & $73.65$ & $16.14$ & $72.36$ & $18.74$ & $73.10$ & $49.72$ & \cellcolor{darkgreen}$77.55$ & $14.28$ & \cellcolor{lightgreen}$76.43$ & 79.77 \\
            & \sage & $17.25$ & $77.37$ & $18.00$ & $76.99$ & $19.48$ & \cellcolor{lightgreen}$77.40$ & $19.85$ & $75.51$ & $49.35$ & $76.99$ & $14.28$ & \cellcolor{darkgreen}$77.55$ & 82.37\\
            Cora & \gcn & $17.99$ & $78.11$ & $17.81$ & $77.92$ & $18.55$ & $77.74$ & $20.41$ & \cellcolor{lightgreen}$79.22$ & $47.31$ & \cellcolor{darkgreen}$80.52$ & $14.28$ & $79.03$ & 84.60\\
            & \gin & $16.69$ & $76.44$ & $18.74$ & \cellcolor{darkgreen}$80.52$ & $17.44$ & $76.25$ & $19.29$ & $76.62$ & $48.79$ & \cellcolor{lightgreen}$78.85$ & $14.28$ & $76.06$ & 81.63\\
            \midrule
            & \gat & $16.51$ & $61.82$ & $25.00$ & $62.27$ & $19.24$ & \cellcolor{darkgreen}$64.70$ & $18.18$ & \cellcolor{lightgreen}$63.48$ & $20.91$ & $62.73$ & $16.67$ & $62.27$ & 66.42\\
            & \sage & $16.66$ & $68.48$ & $16.67$ & $68.64$ & $22.12$ & $69.39$ & $22.42$ & \cellcolor{lightgreen}$69.85$ & $22.88$ & \cellcolor{darkgreen}$71.52$ & $16.67$ & $69.39$ & 72.57\\
            Citeseer & \gcn & $28.18$ & $60.00$ & $16.67$ & $61.97$ & $20.45$ & \cellcolor{darkgreen}$65.45$ & $19.70$ & $64.24$ & $21.06$ & \cellcolor{lightgreen}$64.70$ & $16.67$ & $63.18$ & 68.03\\
            & \gin & $16.66$ & \cellcolor{lightgreen}$64.39$ & $16.67$ & \cellcolor{darkgreen}$63.94$ & $20.15$ & $59.85$ & $18.64$ & $63.64$ & $22.12$ & $60.30$ & $16.67$ & $61.81$ & 67.38\\      
        \midrule
        & \gat & 29.55 & 92.07 & OOM & 90.86 & OOT & 91.64 & OOT & 91.64 & 35.79 & \cellcolor{lightgreen}92.52 & OOT & \cellcolor{darkgreen}93.74 & 89.49 \\
          & \sage & 26.75 & 87.89 & OOM & \cellcolor{darkgreen}91.05 & OOT & 86.64 & OOT & 86.64 & 32.92 & \cellcolor{lightgreen}90.44 & OOT & 86.01 & 90.03 \\
          Syn 4 & \gcn & 28.85 & \cellcolor{darkgreen}51.97 & OOM & 19.58 & OOT & 18.29 & OOT & 18.92 & 33.80 & 26.60 & OOT & \cellcolor{lightgreen}36.85 & 21.43 \\
          & GIN & 28.50 & \cellcolor{darkgreen}65.61 & OOM & 31.06 & OOT & 26.51  & OOT & 26.56 & 34.03 & 46.40 & OOT & \cellcolor{lightgreen}47.10 & 29.35 \\
        \midrule
        & \gat & 44.00 & 86.80 & 43.60 & 86.60 & 30.00 & 78.75 & 55.40 & \cellcolor{lightgreen}92.80 & 36.20 & \cellcolor{darkgreen}93.60 & 31.80 & \cellcolor{lightgreen}92.80 &  97.20 \\
          & \sage & 41.00 & 93.80 & 41.40 & 93.60 & 33.75 & 88.75 & 57.60 & 94.00 & 35.20 & \cellcolor{lightgreen}94.80 & 28.20 & \cellcolor{darkgreen}95.60 &  97.40 \\
          Syn 6 & \gcn & 43.60 & 88.80 & 42.20 & 87.40 & 26.25 & 81.25 & 55.60 & 92.40 & 31.40 & \cellcolor{darkgreen}94.40 & 25.20 & \cellcolor{lightgreen}94.00 &  99.40 \\
          & GIN & 39.60 & 89.00 & 40.40 & 86.60 & 21.25 & 82.50 & 55.20 & 91.80 & 30.00 & \cellcolor{darkgreen}94.60 & 30.40 & \cellcolor{lightgreen}92.00 &  98.80 \\
        \midrule
         & \gat & 29.55 & \cellcolor{darkgreen}99.75 & 33.75 & 88.75 & 88.25 & \cellcolor{lightgreen}99.25 & 88.25 & \cellcolor{lightgreen}99.25 & 26.00 & 85.50 & 94.00 & 96.00 & 98.50 \\
          & \sage & 26.75 & \cellcolor{darkgreen}100.0 & 32.50 & \cellcolor{darkgreen}100.0 & 88.75 & \cellcolor{lightgreen}99.50 & 88.75 & \cellcolor{lightgreen}99.50 & 26.75 & \cellcolor{darkgreen}100.0 & 92.50 & \cellcolor{darkgreen}100.0 &  100.0 \\
          Syn 8 & \gcn & 28.85 & 98.75 & 31.75 & \cellcolor{lightgreen}99.75 & 88.75 & 99.00 & 88.75 & 99.00 & 28.50 & 99.25 & 95.00 & \cellcolor{darkgreen}100.0 &  100.0 \\
          & GIN & 28.50 & 50.00 & 30.50 & 91.00 & 82.25 & \cellcolor{lightgreen}91.50 & 82.25 & \cellcolor{lightgreen}91.50 & 27.25 & 81.75 & 91.75 & \cellcolor{darkgreen}92.25 & 78.25 \\
    \bottomrule
\end{tabular}}
\label{t:_syn_gnn_accuracies}
\end{table*}


\section{Computational Efficiency} 
Table ~\ref{t_app:time} illustrates the remaining computational time for learning graph structures using GraphFLEx with existing Vanilla methods on Synthetic datasets. While traditional methods may be efficient for small graphs, GraphFLEx scales significantly better, excelling on large datasets like \textit{Pubmed} and \textit{Syn 5}, where most methods fail.
\begin{table*}[!ht]
    \centering
    \setlength{\tabcolsep}{6pt}
    \caption{Computational time for learning graph structures using GraphFLEx (GFlex) with existing methods (Vanilla referred to as Van.). The experimental setup involves treating 50\% of the data as static, while the remaining 50\% of nodes are treated as incoming nodes arriving in 25 different timestamps. The best times are highlighted by color \textcolor{green}{Green}. OOM and OOT denote out-of-memory and out-of-time, respectively.}
    \resizebox{\textwidth}{!}{
        \begin{tabular}{lcc*{5}{cc}}
            \toprule
            \bfseries Data & 
            \multicolumn{2}{c}{\bfseries ANN} & 
            \multicolumn{2}{c}{\bfseries KNN} & 
            \multicolumn{2}{c}{\bfseries log-model} & 
            \multicolumn{2}{c}{\bfseries l2-model} & 
            \multicolumn{2}{c}{\bfseries COVAR} & 
            \multicolumn{2}{c}{\bfseries large-model} \\
            & \bfseries Van. & \bfseries GFlex & \bfseries Van. & \bfseries GFlex & \bfseries Van. & \bfseries GFlex & \bfseries Van. & \bfseries GFlex & \bfseries Van. & \bfseries GFlex & \bfseries Van. & \bfseries GFlex \\
            \midrule
            \midrule
            Syn 1 & 19.4 & \cellcolor{darkgreen}9.8 & \cellcolor{darkgreen}2.5 & 10.5 & 2418 & \cellcolor{darkgreen}56.4 & 37.2 & \cellcolor{darkgreen}8.8 & \cellcolor{darkgreen}3.5 & 8.3 & 205 & \cellcolor{darkgreen}9.4 \\
            Syn 2 & 47.3 & \cellcolor{darkgreen}16.9 & \cellcolor{darkgreen}6.6 & 18.3 & 14000  & \cellcolor{darkgreen}144 & 214 & \cellcolor{darkgreen}22.6 & 20.3 & \cellcolor{darkgreen}18.6 & 1259 & \cellcolor{darkgreen}16.4 \\
            Syn 5 & \cellcolor{darkgreen}5.1  & 11.5 & \cellcolor{darkgreen}0.8 & 7.3 & 57.4 & \cellcolor{darkgreen}28 & \cellcolor{darkgreen}1.1 & 5.8 & \cellcolor{darkgreen}0.2 & 4.8 & \cellcolor{darkgreen}3.2 & 5.3 \\
            Syn 6 & 16.6 & \cellcolor{darkgreen}9.9 & 2.8 & \cellcolor{darkgreen}11.4 & 1766 & \cellcolor{darkgreen}96.3 & 193 & \cellcolor{darkgreen}101 & \cellcolor{darkgreen}5.3 & 8.9 & 324 & \cellcolor{darkgreen}9.6 \\
            Syn 7 & 10.6 & \cellcolor{darkgreen}7.4 & \cellcolor{darkgreen}1.4 & 8.9 & 704 & \cellcolor{darkgreen}85.2 & 10.3 & \cellcolor{darkgreen}7.9 & \cellcolor{darkgreen}0.9 & 6.4 & 36.5 & \cellcolor{darkgreen}8.2 \\
            Syn 8 & 19.6 & \cellcolor{darkgreen}11.2 & \cellcolor{darkgreen}2.5 & 11.7 & 2416 & \cellcolor{darkgreen}457 & 37.2 & \cellcolor{darkgreen}17.0 & \cellcolor{darkgreen}3.4 & 10.9 & 204 & \cellcolor{darkgreen}11.7 \\
            \bottomrule
        \end{tabular}}
        \label{t_app:time}
\end{table*}

\section{Visualization of Growing graphs}
\label{sec:growing_graphs}
\begin{figure*}[!ht]
\begin{center}
  \textbf{PyGsp} \\
  \vspace{1mm} 
  
  \begin{subfigure}[t]{.19\linewidth}
    \includegraphics[width=\linewidth, trim={2.3cm 1.4cm 1.9cm 1.6cm}, clip]{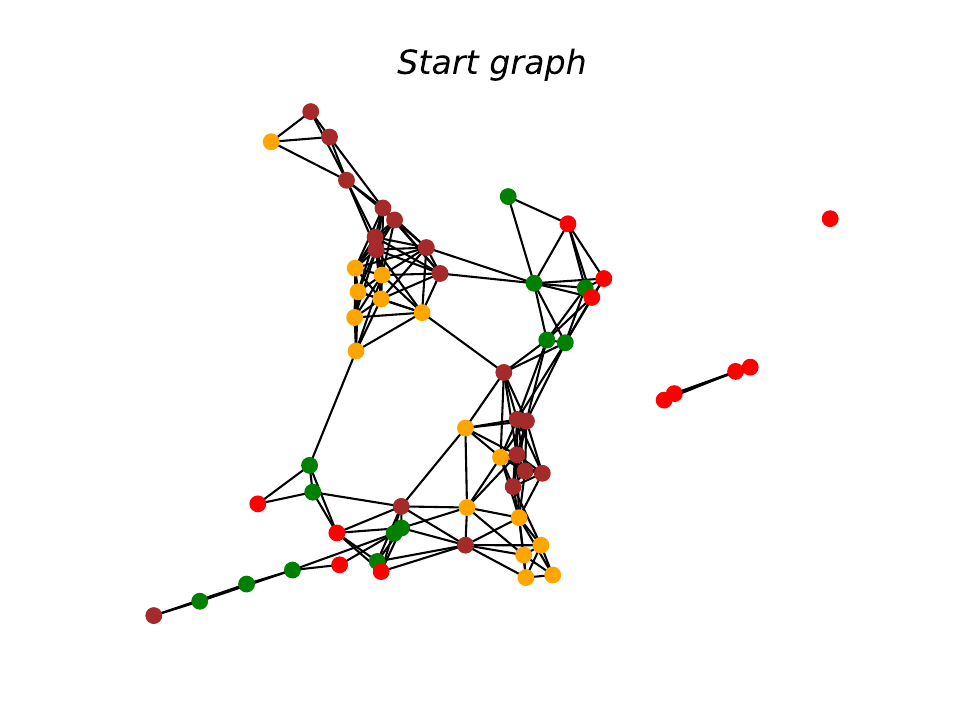}
    \captionsetup{font=scriptsize, justification=centering}
    \caption{Initial graph $G_0$}
  \end{subfigure}
  \hfill
  \begin{subfigure}[t]{.19\linewidth}
    \includegraphics[width=\linewidth, trim={2.3cm 1.4cm 1.9cm 1.6cm}, clip]{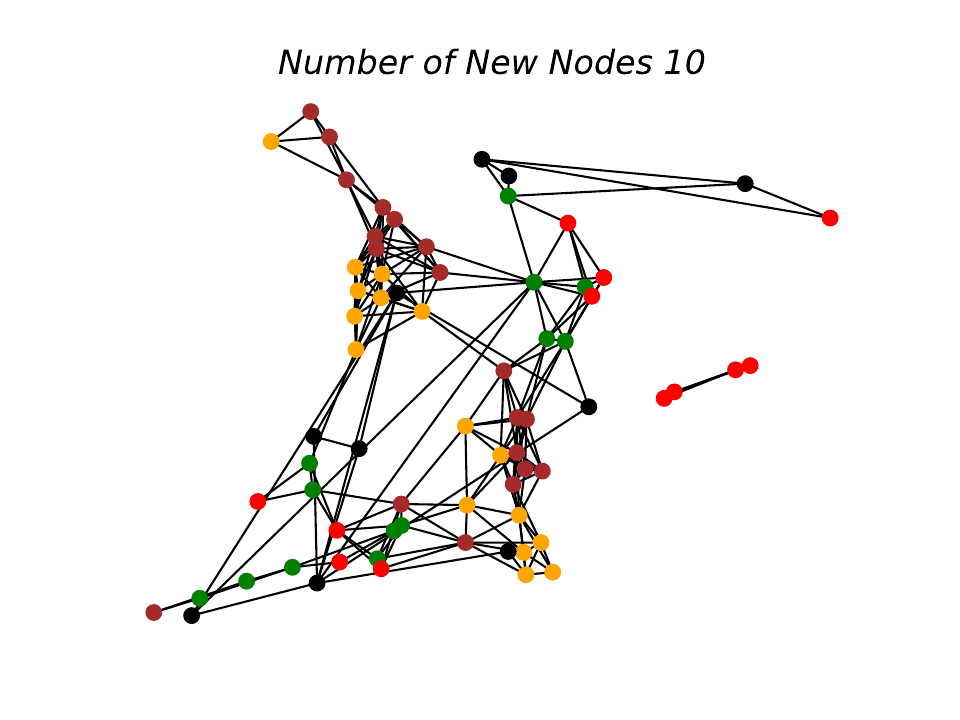}
    \captionsetup{font=scriptsize, justification=centering}
    \caption{$\alpha$= 10, $G_1$}
  \end{subfigure}
  \hfill
  \begin{subfigure}[t]{.19\linewidth}
    \includegraphics[width=\linewidth, trim={2.3cm 1.4cm 1.9cm 1.6cm}, clip]{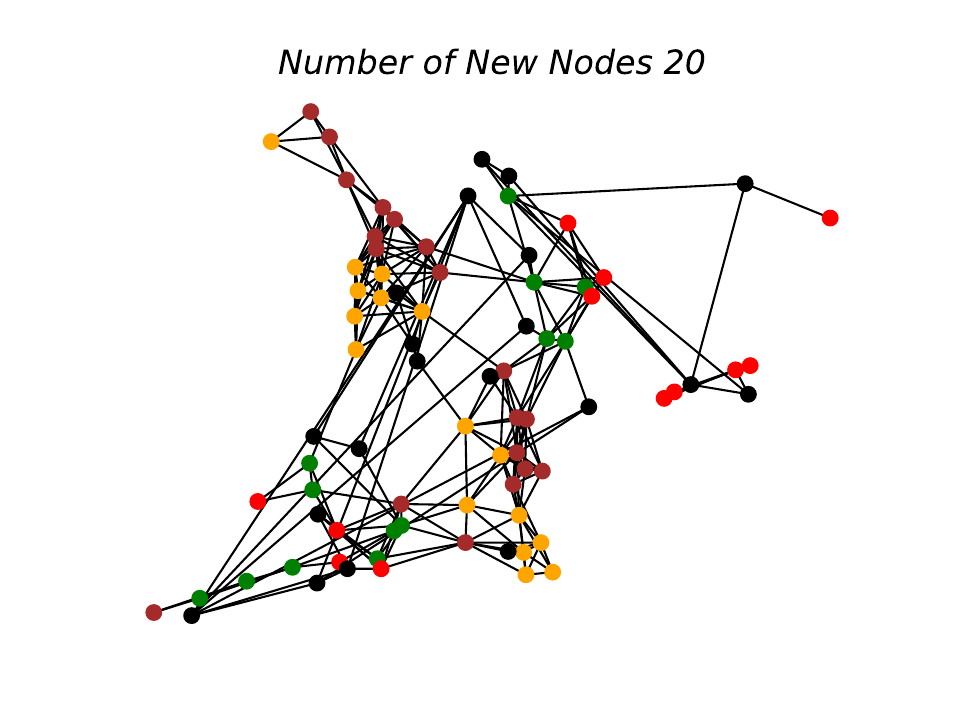}
    \captionsetup{font=scriptsize, justification=centering}
    \caption{$\alpha$= 20, $G_2$}
  \end{subfigure}
  \hfill
  \begin{subfigure}[t]{.19\linewidth}
    \includegraphics[width=\linewidth, trim={2.3cm 1.4cm 1.9cm 1.6cm}, clip]{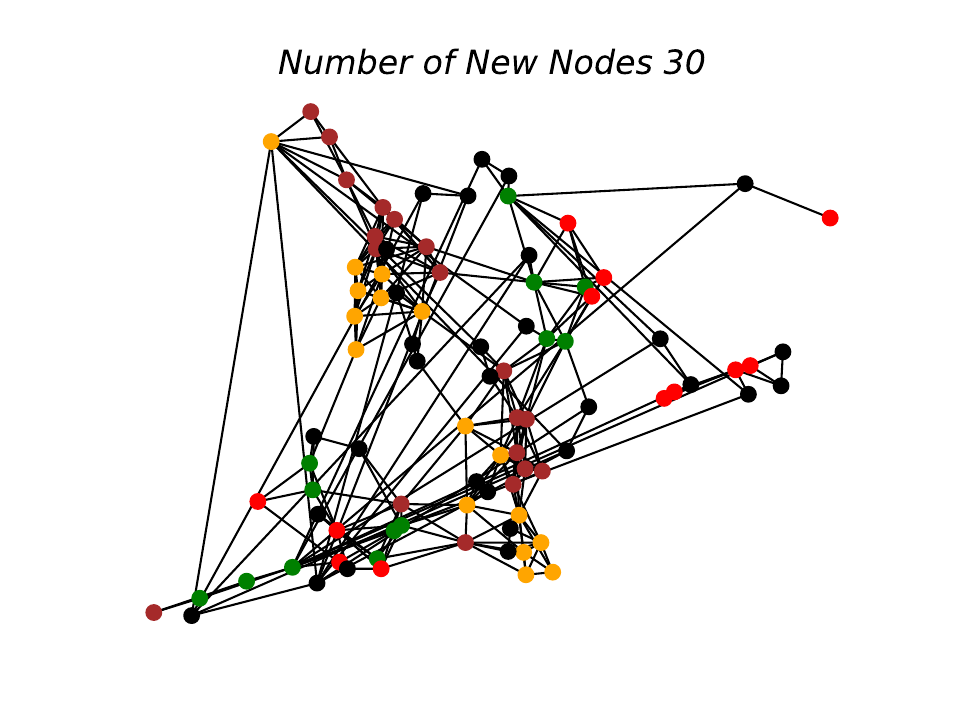}
    \captionsetup{font=scriptsize, justification=centering}
    \caption{$\alpha$ = 30, $G_3$}
  \end{subfigure}
  \hfill
  \begin{subfigure}[t]{.19\linewidth}
    \includegraphics[width=\linewidth, trim={2.3cm 1.4cm 1.9cm 1.6cm}, clip]{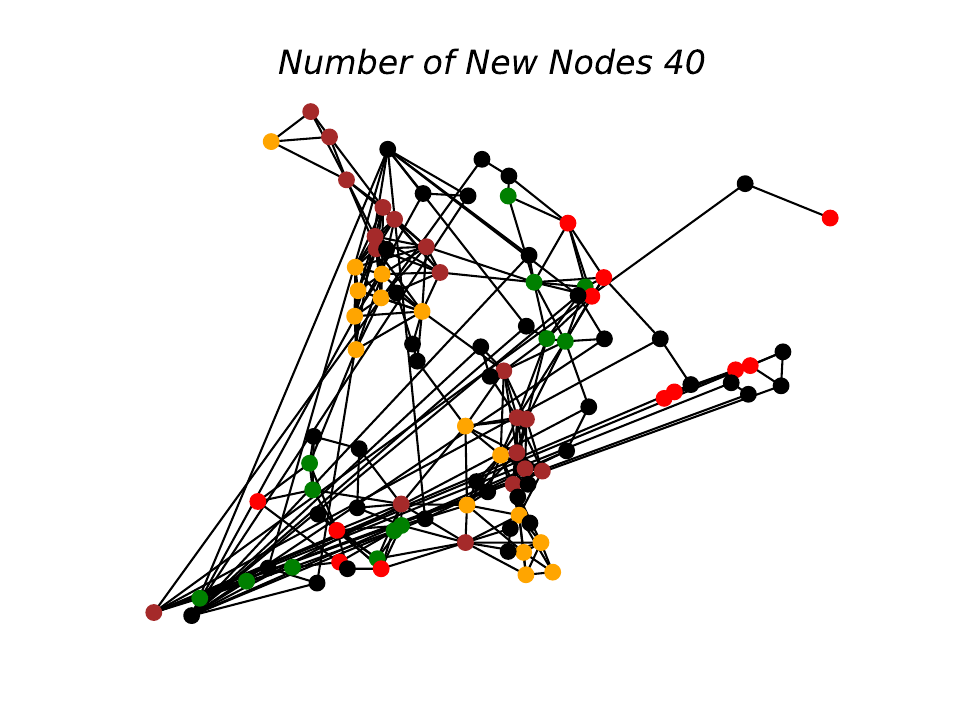}
    \captionsetup{font=scriptsize, justification=centering}
    \caption{$\alpha$ = 40, $G_4$}
  \end{subfigure}
  \hfill
  
\end{center}
\caption{This figure illustrates the growing structure learned using GraphFlex for dynamic nodes. New nodes are denoted using black color, and $\alpha$ denotes number of new nodes. \textit{PyGsp} denotes type synthetic graph.}
\label{fig:incremental}
\vskip -0.2in
\end{figure*}

\begin{figure*}[!ht]
\begin{center}
  \textbf{HE} \\
  \vspace{1mm} 
  
  \begin{subfigure}[t]{.16\linewidth}
    \includegraphics[width=\linewidth, trim={2.3cm 1.4cm 1.9cm 1.6cm}, clip]{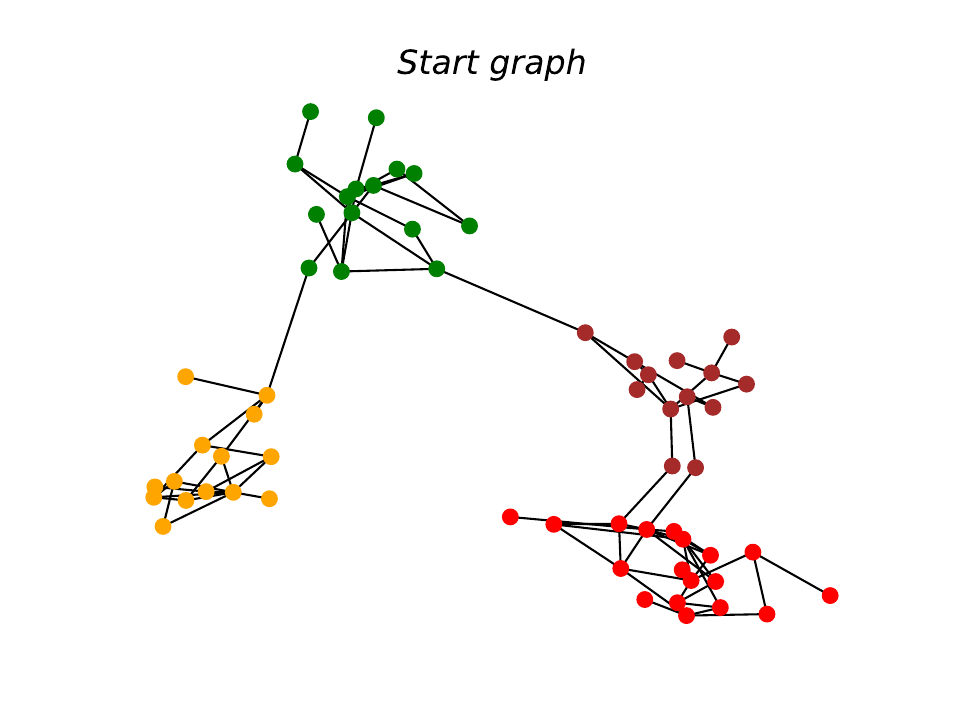}
    \captionsetup{font=scriptsize, justification=centering}
    \caption{Initial graph $G_0$}
  \end{subfigure}
  \hfill
  \begin{subfigure}[t]{.16\linewidth}
    \includegraphics[width=\linewidth, trim={2.3cm 1.4cm 1.9cm 1.6cm}, clip]{images/svg/Mohit_learned_graph_synthatic_hetero__100_sensor_FACH_gnn_ann_0.4__3.pdf}
    \captionsetup{font=scriptsize, justification=centering}
    \caption{$\alpha$ = 10, $G_1$}
  \end{subfigure}
  \hfill
  \begin{subfigure}[t]{.16\linewidth}
    \includegraphics[width=\linewidth, trim={2.3cm 1.4cm 1.9cm 1.6cm}, clip]{images/svg/Mohit_learned_graph_synthatic_hetero__100_sensor_FACH_gnn_ann_0.5__3.pdf}
    \captionsetup{font=scriptsize, justification=centering}
    \caption{$\alpha$ = 20, $G_2$}
  \end{subfigure}
  \hfill
  \begin{subfigure}[t]{.16\linewidth}
    \includegraphics[width=\linewidth, trim={2.3cm 1.4cm 1.9cm 1.6cm}, clip]{images/svg/Mohit_learned_graph_synthatic_hetero__100_sensor_FACH_gnn_ann_0.6__3.pdf}
    \captionsetup{font=scriptsize, justification=centering}
    \caption{$\alpha$= 30, $G_3$}
  \end{subfigure}
  \hfill
  \begin{subfigure}[t]{.16\linewidth}
    \includegraphics[width=\linewidth, trim={2.3cm 1.4cm 1.9cm 1.6cm}, clip]{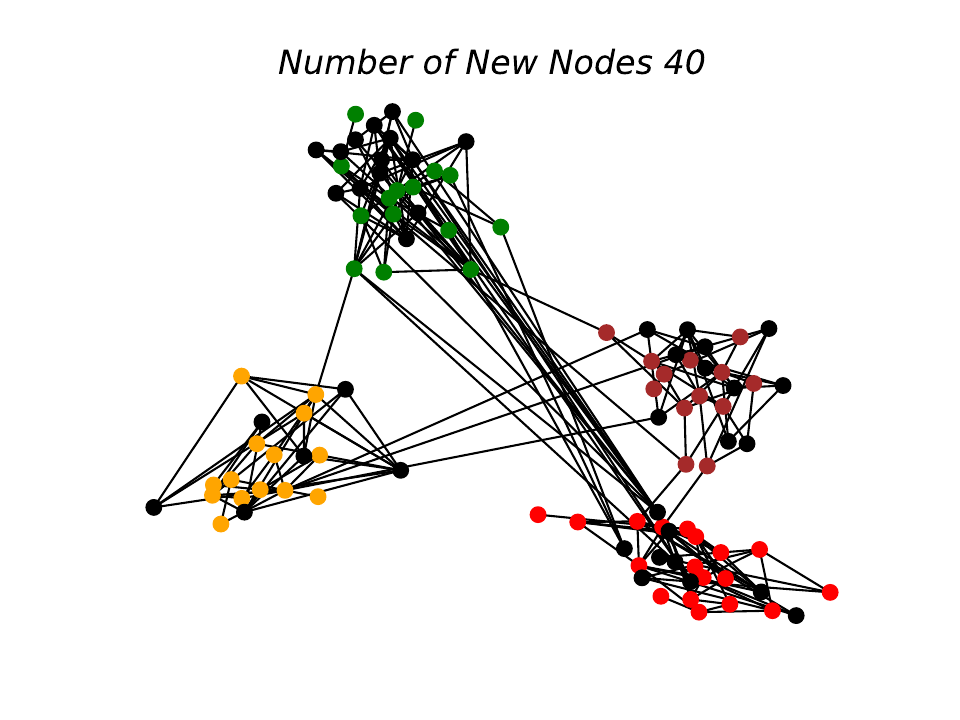}
    \captionsetup{font=scriptsize, justification=centering}
    \caption{$\alpha$ = 40, $G_4$}
  \end{subfigure}
  \hfill
\end{center}
\caption{This figure illustrates the growing structure learned using GraphFlex for dynamic nodes. New nodes are denoted using black color, and $\alpha$ denotes the number of new nodes. \textit{HE} denotes the type of synthetic graph.}
\label{fig:add_incremental}
\vskip -0.2in
\end{figure*}

This section helps us visualize the phases of our growing graphs. 
We have generated a synthetic graph of 60 nodes using PyGSP-Sensor and HE methods mentioned in Appendix~\ref{app:datasets}. We then added 40 new nodes denoted using black color in these existing graphs at four different timestamps. Figure~\ref{fig:incremental} and Figure~\ref{fig:add_incremental} shows the learned graph structure after each timestamp for two different Synthetic graphs.

\begin{figure*}[!ht]
\begin{center}
\centerline{\includegraphics[width=0.6\columnwidth, trim={1cm 0.5cm 0cm 0.5cm}, clip]{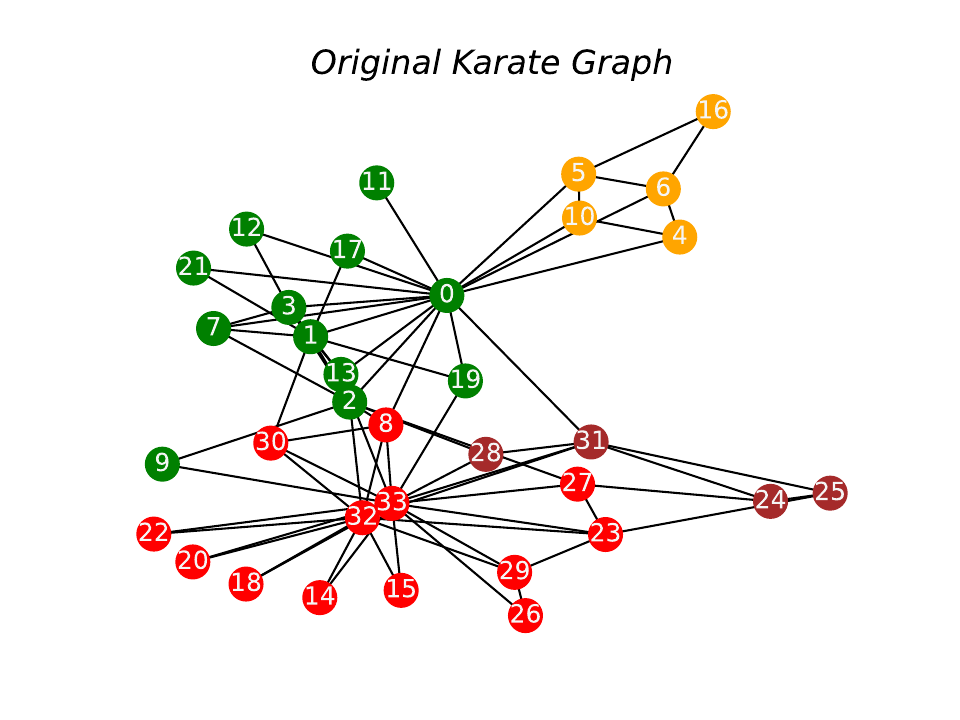}}
\caption{Original Karate Graph}
\label{fig:ori_karate}
\end{center}
\vskip -0.2in
\end{figure*}

\section{Clustering Quality}
\label{app:clustering_quality}
\begin{figure*}[!ht]
\centering
  \begin{subfigure}[t]{.32\textwidth}
    \includegraphics[width=\textwidth, trim={0.9cm 0.8cm 0.9cm 0.6cm}, clip]{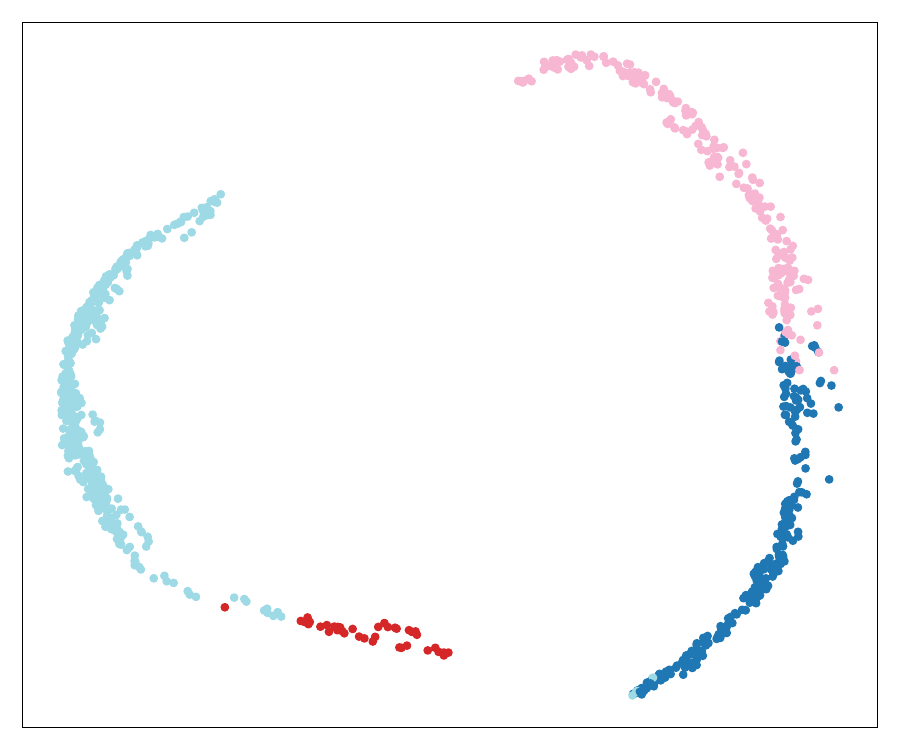}
    \captionsetup{font=scriptsize, justification=centering}
    \caption{Xin}
  \end{subfigure}
  \hfill
  \begin{subfigure}[t]{.32\textwidth}
    \includegraphics[width=\textwidth, trim={0.9cm 0.8cm 0.9cm 0.6cm}, clip]{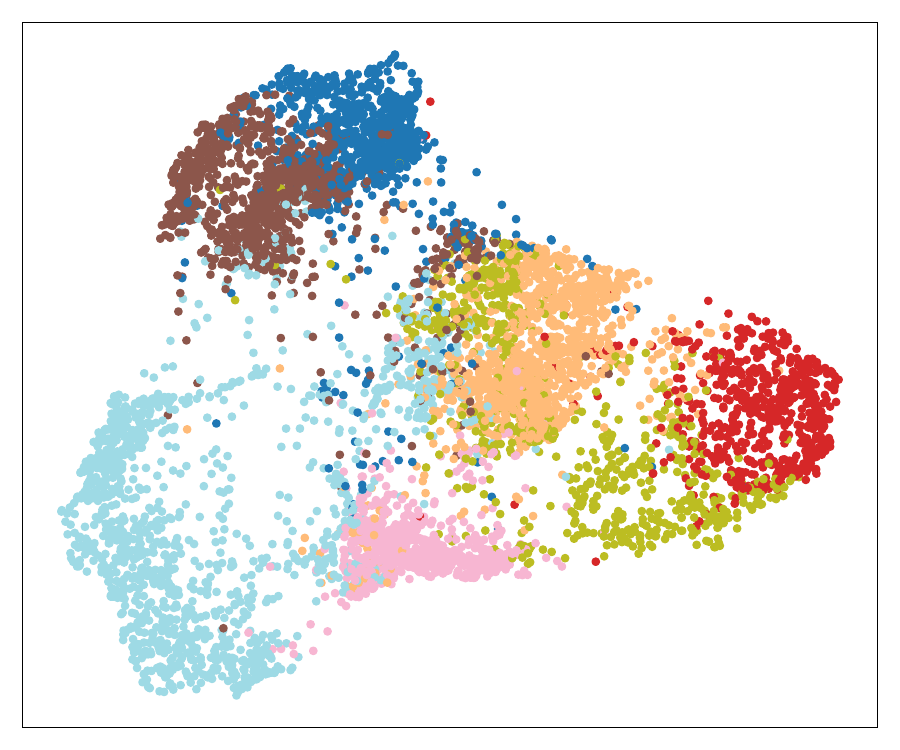}
    \captionsetup{font=scriptsize, justification=centering}
    \caption{MNIST}
  \end{subfigure}
  \hfill
  \begin{subfigure}[t]{.32\textwidth}
    \includegraphics[width=\textwidth, trim={0.9cm 0.8cm 0.9cm 0.6cm}, clip]{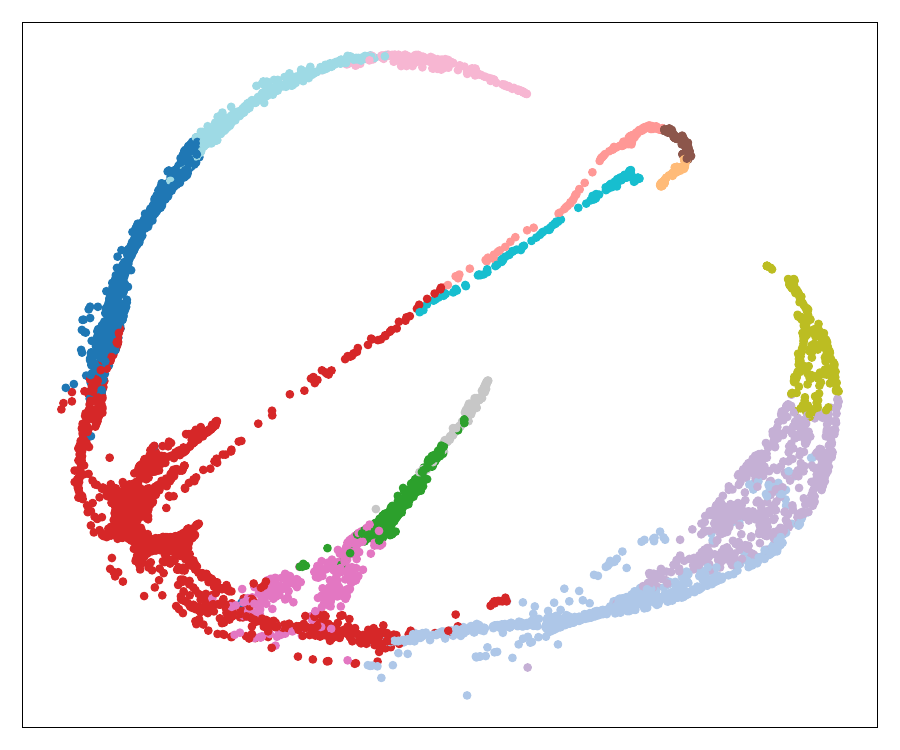}
    \captionsetup{font=scriptsize, justification=centering}
    \caption{Baron Human}
  \end{subfigure}
  \begin{subfigure}[t]{.32\textwidth}
    \includegraphics[width=\textwidth, trim={0.9cm 0.8cm 0.9cm 0.6cm}, clip]{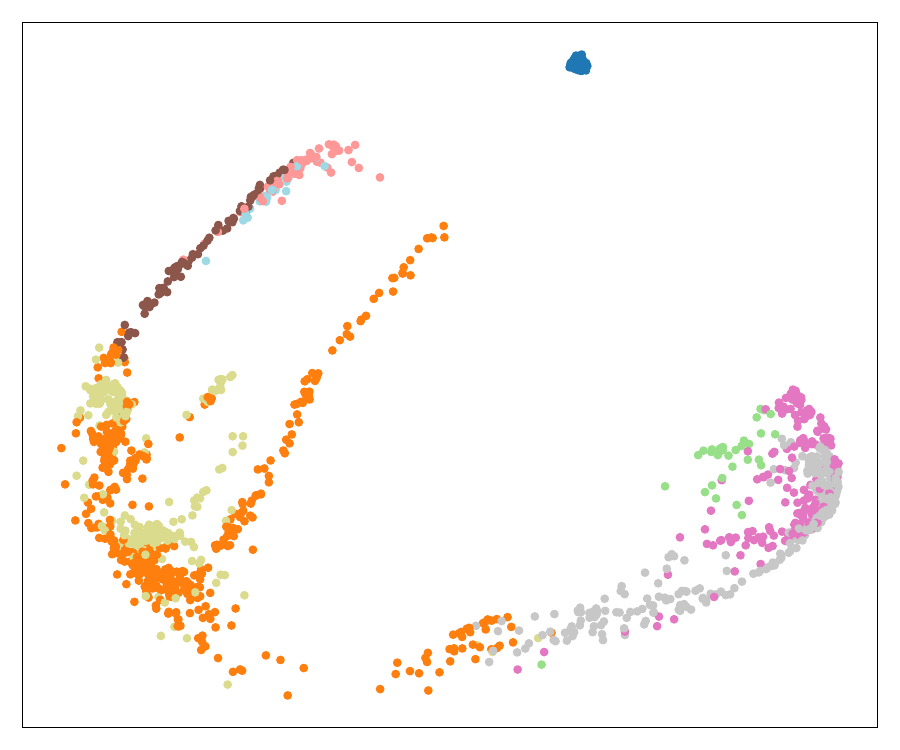}
    \captionsetup{font=scriptsize, justification=centering}
    \caption{Muraro}
  \end{subfigure}
  \hfill
  \begin{subfigure}[t]{.32\textwidth}
    \includegraphics[width=\textwidth, trim={0.9cm 0.8cm 0.9cm 0.6cm}, clip]{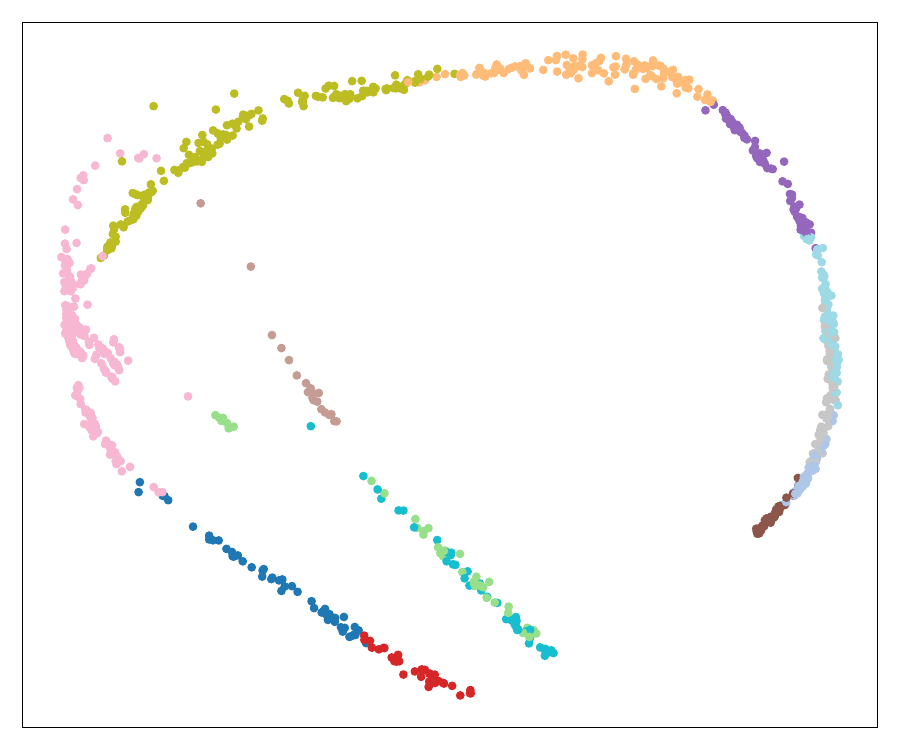}
    \captionsetup{font=scriptsize, justification=centering}
    \caption{Baron Mouse}
  \end{subfigure}
  \hfill
  \begin{subfigure}[t]{.32\textwidth}
    \includegraphics[width=\textwidth, trim={0.9cm 0.8cm 0.9cm 0.6cm}, clip]{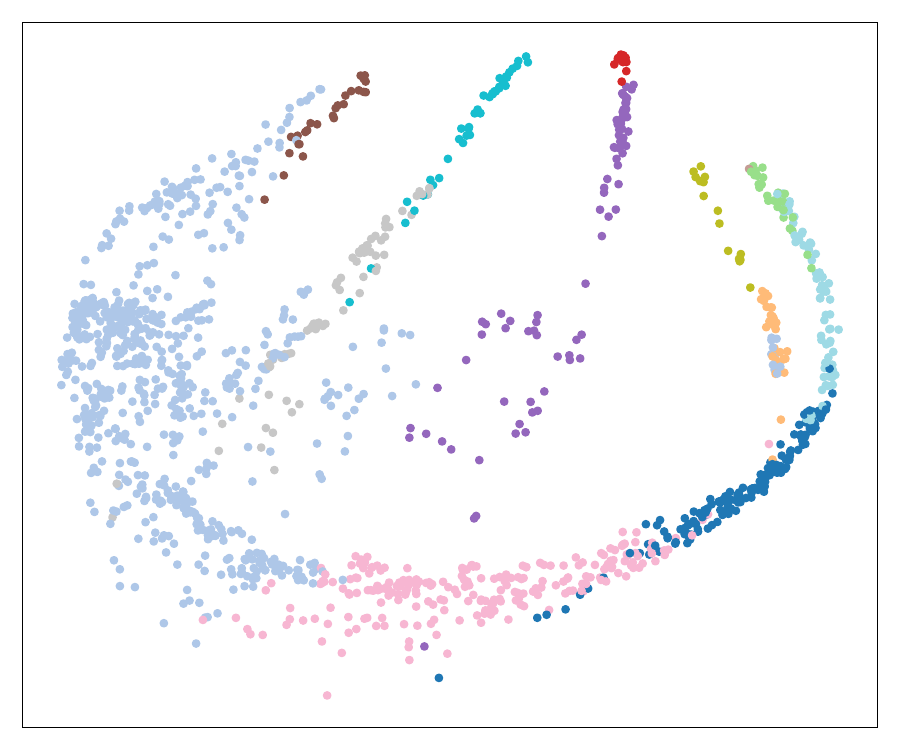}
    \captionsetup{font=scriptsize, justification=centering}
    \caption{Segerstolpe}
  \end{subfigure}
\caption{PHATE visualization of clusters learnt using GraphFlex clustering module for scRNA-seq datasets.}
\label{app:vis_clusters}
\end{figure*}

Figure ~\ref{app:vis_clusters} shows the PHATE ~\cite{moon2019visualizing} visualization of clusters learned using GraphFLEx's clustering module $\mathcal{M}_{clust}$ for 6 single-cell RNA datasets, namely $Xin$, $MNIST$, $Baron-Human$, $Muraro$, $Baron Mouse$, and $Segerstolpe$ datasets.

\begin{figure*}[!ht]
\begin{center}
  \textbf{Vanilla} \\
  \vspace{1mm} 
  
  \begin{subfigure}[t]{.18\linewidth}
    \includegraphics[width=\linewidth, trim={2.3cm 1.4cm 1.9cm 1.6cm}, clip]{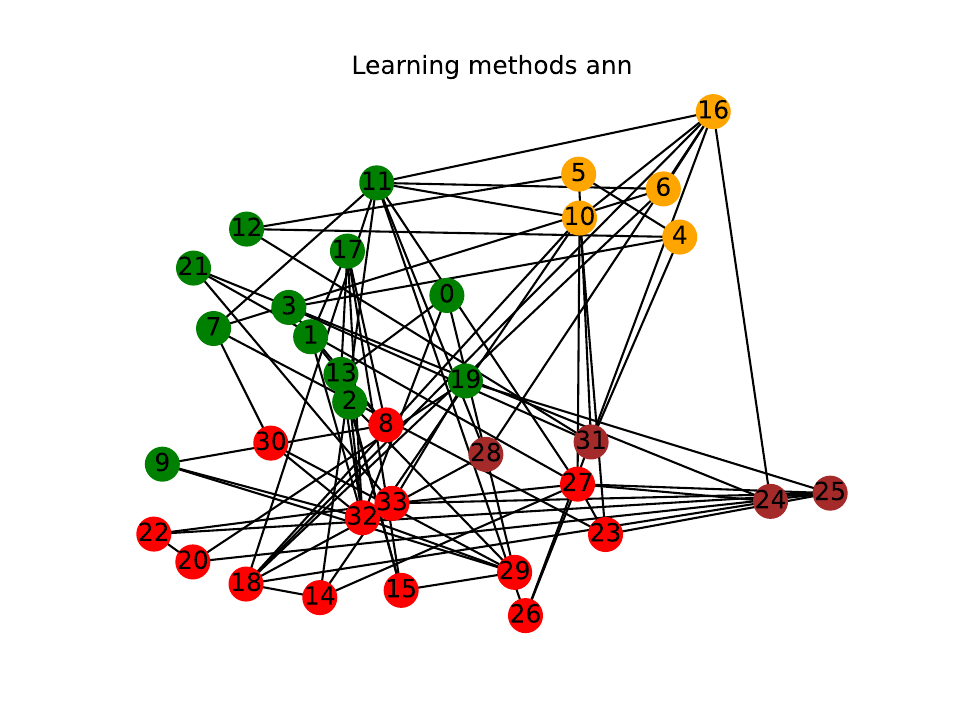}
    \captionsetup{font=scriptsize, justification=centering}
    \caption{ANN}
  \end{subfigure}
  \hfill
  \begin{subfigure}[t]{.18\linewidth}
    \includegraphics[width=\linewidth, trim={2.3cm 1.4cm 1.9cm 1.6cm}, clip]{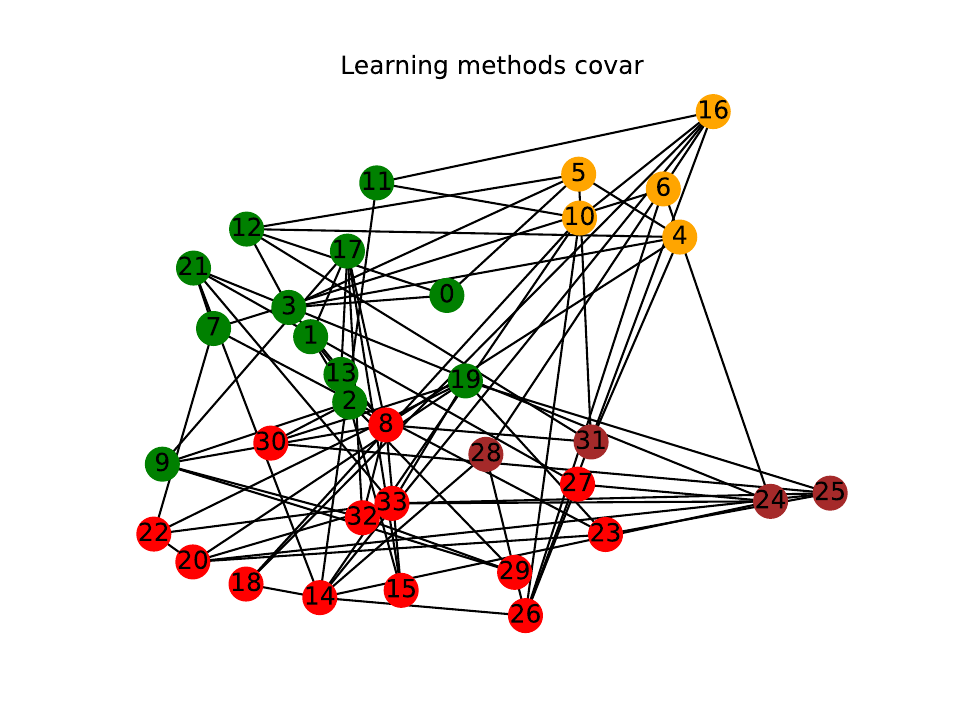}
    \captionsetup{font=scriptsize, justification=centering}
    \caption{Emp. Cov.}
  \end{subfigure}
  \hfill
  \begin{subfigure}[t]{.18\linewidth}
    \includegraphics[width=\linewidth, trim={2.3cm 1.4cm 1.9cm 1.6cm}, clip]{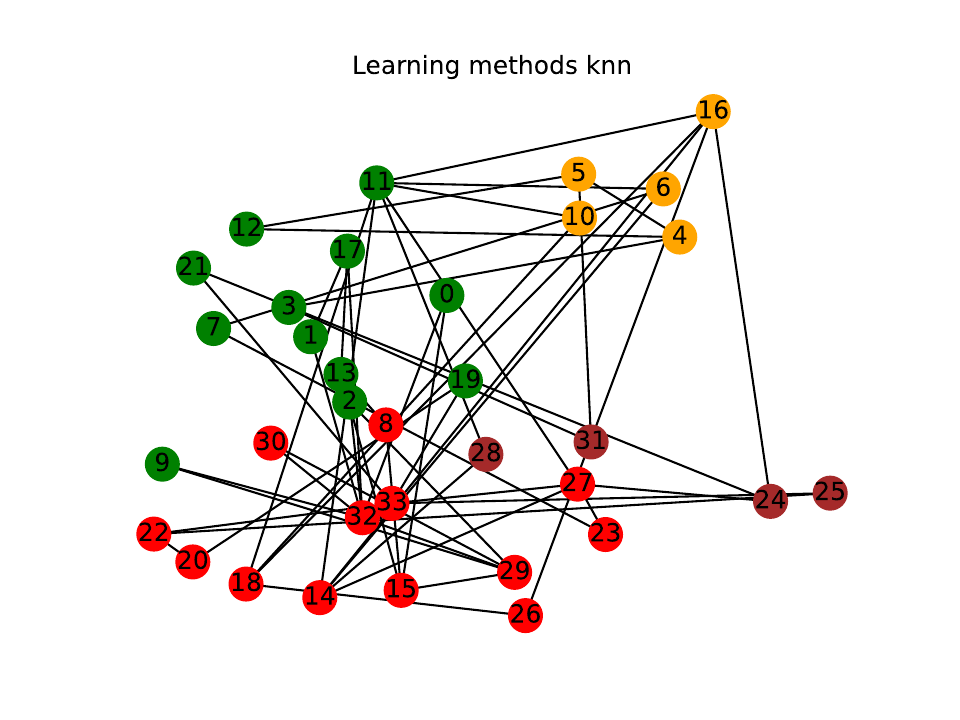}
    \captionsetup{font=scriptsize, justification=centering}
    \caption{KNN}
  \end{subfigure}
  \hfill
  \begin{subfigure}[t]{.18\linewidth}
    \includegraphics[width=\linewidth, trim={2.3cm 1.4cm 1.9cm 1.6cm}, clip]{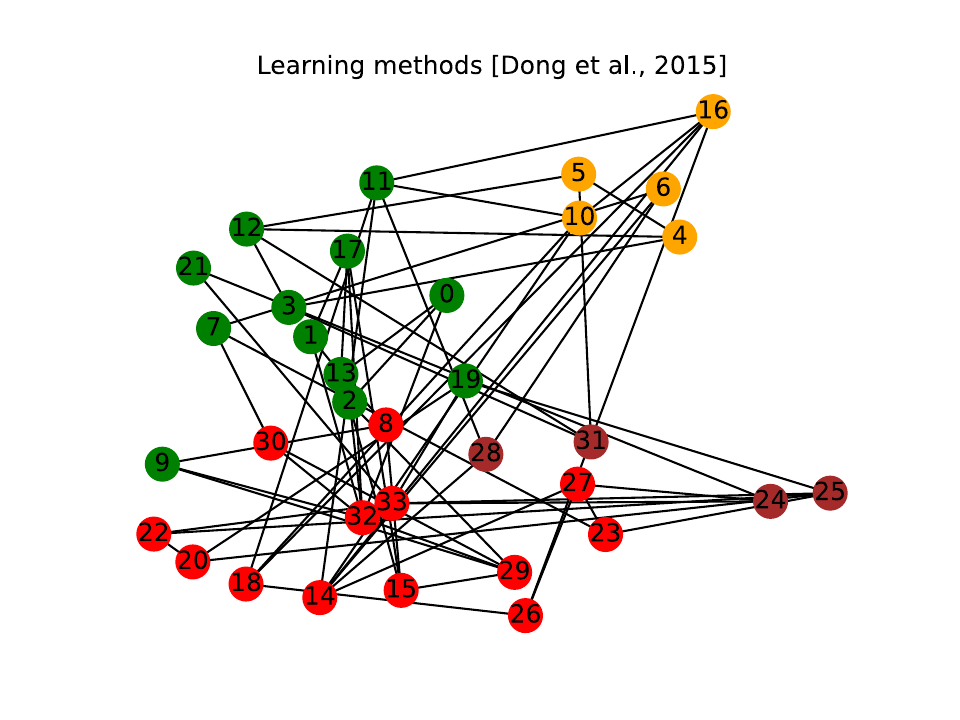}
    \captionsetup{font=scriptsize, justification=centering}
    \caption{L2 model}
  \end{subfigure}
  \hfill
  \begin{subfigure}[t]{.18\linewidth}
    \includegraphics[width=\linewidth, trim={2.3cm 1.4cm 1.9cm 1.6cm}, clip]{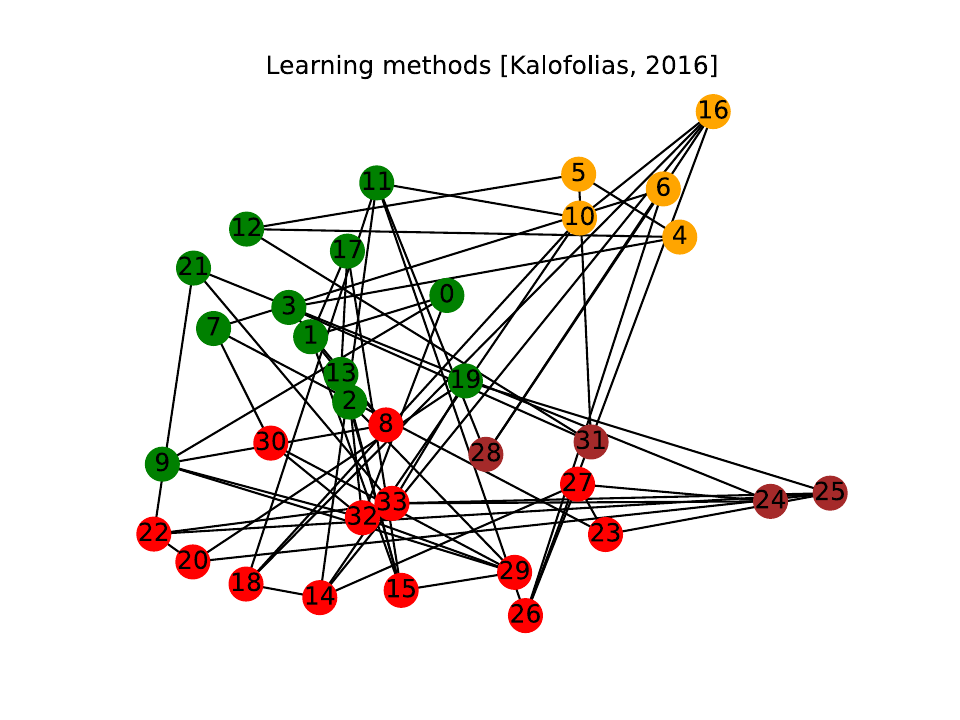}
    \captionsetup{font=scriptsize, justification=centering}
    \caption{Log model}
  \end{subfigure}
  \hfill
  
  \vspace{2mm} 
  
  \textbf{GraphFlex} \\
  \vspace{1mm} 
  
  \begin{subfigure}[t]{.18\linewidth}
    \includegraphics[width=\linewidth, trim={2.3cm 1.4cm 1.9cm 1.6cm}, clip]{images/svg/Mohit_learned_graph_karate_small_world__34_sensor_FACH_gnn_ann_0.3__3.pdf}
    \captionsetup{font=scriptsize, justification=centering}
    \caption{ANN}
  \end{subfigure}
  \hfill
  \begin{subfigure}[t]{.18\linewidth}
    \includegraphics[width=\linewidth, trim={2.3cm 1.4cm 1.9cm 1.6cm}, clip]{images/svg/Mohit_learned_graph_karate_small_world__34_sensor_FACH_gnn_covar_0.3__3.pdf}
    \captionsetup{font=scriptsize, justification=centering}
    \caption{Emp Cov.}
  \end{subfigure}
  \hfill
  \begin{subfigure}[t]{.18\linewidth}
    \includegraphics[width=\linewidth, trim={2.3cm 1.4cm 1.9cm 1.6cm}, clip]{images/svg/Mohit_learned_graph_karate_small_world__34_sensor_FACH_gnn_knn_0.3__3.pdf}
    \captionsetup{font=scriptsize, justification=centering}
    \caption{KNN}
  \end{subfigure}
  \hfill
  \begin{subfigure}[t]{.18\linewidth}
    \includegraphics[width=\linewidth, trim={2.3cm 1.4cm 1.9cm 1.6cm}, clip]{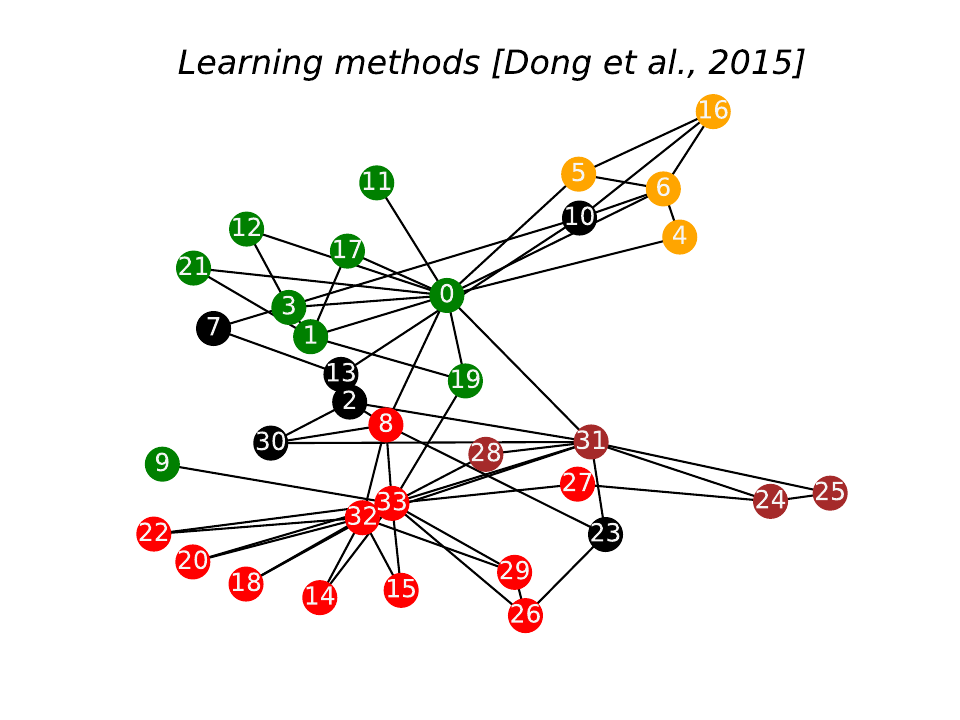}
    \captionsetup{font=scriptsize, justification=centering}
    \caption{L2 model}
  \end{subfigure}
  \hfill
  \begin{subfigure}[t]{.18\linewidth}
    \includegraphics[width=\linewidth, trim={2.3cm 1.4cm 1.9cm 1.6cm}, clip]{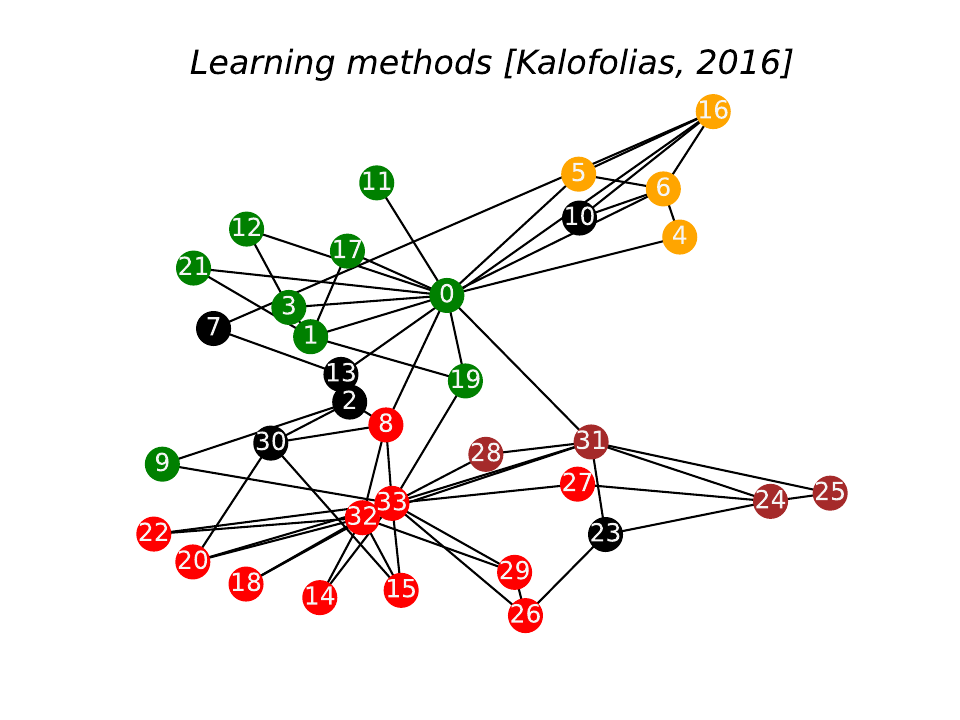}
    \captionsetup{font=scriptsize, justification=centering}
    \caption{Log model}
  \end{subfigure}
  \hfill
  
\end{center}
\caption{This figure compares the structures learned on Zachary's karate dataset when existing methods are employed with GraphFlex and when existing methods are used individually. We consider six nodes, denoted in black, as dynamic nodes.}
\label{fig:karate_dataset}
\vskip -0.2in
\end{figure*}

\section{Ablation Study}
\label{app:ablation}
In this section, we present an ablation study to analyze the role of individual modules within GraphFLEx and their influence on the final graph structure. Specifically, we focus on two aspects: (i) the significance of the clustering module, and (ii) the effect of varying module configurations on the learned graph topology.

\subsection{Clustering Module Evaluation}
To evaluate the effectiveness of the clustering module, we compute standard metrics such as Normalized Mutual Information (NMI), Conductance (C), and Modularity (Q) across various datasets (see Table ~\ref{tab:clust_results} in Section~\ref{sec:clust_qual}). These metrics collectively validate the quality of the discovered clusters, thereby justifying the use of a clustering module as a foundational step in GraphFLEx. Since clustering in GraphFLEx is applied only once on a randomly sampled small set of nodes, selecting the right method can be considered as part of hyperparameter tuning, where these clustering measures can guide the optimal choice based on dataset characteristics.
\subsection{Impact of Module Choices on Learned Graph Structure}
This section involves a comparison of the graph structure learned from GraphFlex with existing methods. Six nodes were randomly selected and considered as new nodes. Figure~\ref{fig:karate_dataset} visually depicts the structures learned using GraphFlex compared to other methods. It is evident from the figure that the structure known with GraphFlex closely resembles the original graph structure. Figure~\ref{fig:ori_karate} shows the original structure of Zachary's karate club network \cite{zachary1977information}. We assumed six random nodes to be dynamic nodes, and the structure learned using GraphFlex compared to existing methods is shown in Figure~\ref{fig:karate_dataset}.

%% file: main.bbl
\begin{thebibliography}{10}

\bibitem{zhou2020graph}
J.~Zhou, G.~Cui, S.~Hu, Z.~Zhang, C.~Yang, Z.~Liu, L.~Wang, C.~Li, and M.~Sun, ``Graph neural networks: A review of methods and applications,'' {\em AI open}, vol.~1, pp.~57--81, 2020.

\bibitem{fout2017protein}
A.~Fout, J.~Byrd, B.~Shariat, and A.~Ben-Hur, ``Protein interface prediction using graph convolutional networks,'' {\em Advances in neural information processing systems}, vol.~30, 2017.

\bibitem{wu2020graph}
Y.~Wu, D.~Lian, Y.~Xu, L.~Wu, and E.~Chen, ``Graph convolutional networks with markov random field reasoning for social spammer detection,'' in {\em Proceedings of the AAAI conference on artificial intelligence}, vol.~34, pp.~1054--1061, 2020.

\bibitem{Malik_Gupta_Kumar_2025}
N.~Malik, R.~Gupta, and S.~Kumar, ``Hyperdefender: A robust framework for hyperbolic gnns,'' {\em Proceedings of the AAAI Conference on Artificial Intelligence}, vol.~39, pp.~19396--19404, Apr. 2025.

\bibitem{gu2019scene}
J.~Gu, H.~Zhao, Z.~Lin, S.~Li, J.~Cai, and M.~Ling, ``Scene graph generation with external knowledge and image reconstruction,'' in {\em Proceedings of the IEEE/CVF conference on computer vision and pattern recognition}, pp.~1969--1978, 2019.

\bibitem{zhu2021survey}
Y.~Zhu, W.~Xu, J.~Zhang, Y.~Du, J.~Zhang, Q.~Liu, C.~Yang, and S.~Wu, ``A survey on graph structure learning: Progress and opportunities,'' {\em arXiv preprint arXiv:2103.03036}, 2021.

\bibitem{allen2012comparing}
J.~D. Allen, Y.~Xie, M.~Chen, L.~Girard, and G.~Xiao, ``Comparing statistical methods for constructing large scale gene networks,'' {\em PloS one}, vol.~7, no.~1, p.~e29348, 2012.

\bibitem{liu2022towards}
Y.~Liu, Y.~Zheng, D.~Zhang, H.~Chen, H.~Peng, and S.~Pan, ``Towards unsupervised deep graph structure learning,'' in {\em Proceedings of the ACM Web Conference 2022}, pp.~1392--1403, 2022.

\bibitem{chen2022graph}
Y.~Chen and L.~Wu, ``Graph neural networks: Graph structure learning,'' {\em Graph Neural Networks: Foundations, Frontiers, and Applications}, pp.~297--321, 2022.

\bibitem{wu2022nodeformer}
Q.~Wu, W.~Zhao, Z.~Li, D.~P. Wipf, and J.~Yan, ``Nodeformer: A scalable graph structure learning transformer for node classification,'' {\em Advances in Neural Information Processing Systems}, vol.~35, pp.~27387--27401, 2022.

\bibitem{jin2020graph}
W.~Jin, Y.~Ma, X.~Liu, X.~Tang, S.~Wang, and J.~Tang, ``Graph structure learning for robust graph neural networks,'' in {\em Proceedings of the 26th ACM SIGKDD international conference on knowledge discovery \& data mining}, pp.~66--74, 2020.

\bibitem{lao2022variational}
D.~Lao, X.~Yang, Q.~Wu, and J.~Yan, ``Variational inference for training graph neural networks in low-data regime through joint structure-label estimation,'' in {\em Proceedings of the 28th ACM SIGKDD conference on knowledge discovery and data mining}, pp.~824--834, 2022.

\bibitem{fatemi2021slaps}
B.~Fatemi, L.~El~Asri, and S.~M. Kazemi, ``Slaps: Self-supervision improves structure learning for graph neural networks,'' {\em Advances in Neural Information Processing Systems}, vol.~34, pp.~22667--22681, 2021.

\bibitem{dong2011efficient}
W.~Dong, C.~Moses, and K.~Li, ``Efficient k-nearest neighbor graph construction for generic similarity measures,'' in {\em Proceedings of the 20th international conference on World wide web}, pp.~577--586, 2011.

\bibitem{muja2014scalable}
M.~Muja and D.~G. Lowe, ``Scalable nearest neighbor algorithms for high dimensional data,'' {\em IEEE transactions on pattern analysis and machine intelligence}, vol.~36, no.~11, pp.~2227--2240, 2014.

\bibitem{kmeans}
J.~MacQueen {\em et~al.}, ``Some methods for classification and analysis of multivariate observations,'' in {\em Proceedings of the fifth Berkeley symposium on mathematical statistics and probability}, vol.~1, pp.~281--297, Oakland, CA, USA, 1967.

\bibitem{wang}
F.~Wang and C.~Zhang, ``Label propagation through linear neighborhoods,'' in {\em Proceedings of the 23rd international conference on Machine learning}, pp.~985--992, 2006.

\bibitem{hsieh2011sparse}
C.-J. Hsieh, I.~Dhillon, P.~Ravikumar, and M.~Sustik, ``Sparse inverse covariance matrix estimation using quadratic approximation,'' {\em Advances in neural information processing systems}, vol.~24, 2011.

\bibitem{friedman2008sparse}
J.~Friedman, T.~Hastie, and R.~Tibshirani, ``Sparse inverse covariance estimation with the graphical lasso,'' {\em Biostatistics}, vol.~9, no.~3, pp.~432--441, 2008.

\bibitem{dong2016learning}
X.~Dong, D.~Thanou, P.~Frossard, and P.~Vandergheynst, ``Learning laplacian matrix in smooth graph signal representations,'' {\em IEEE Transactions on Signal Processing}, vol.~64, no.~23, pp.~6160--6173, 2016.

\bibitem{kalofolias2016learn}
V.~Kalofolias, ``How to learn a graph from smooth signals,'' in {\em Artificial intelligence and statistics}, pp.~920--929, PMLR, 2016.

\bibitem{hashemi2024comprehensive}
M.~Hashemi, S.~Gong, J.~Ni, W.~Fan, B.~A. Prakash, and W.~Jin, ``A comprehensive survey on graph reduction: Sparsification, coarsening, and condensation,'' {\em arXiv preprint arXiv:2402.03358}, 2024.

\bibitem{fortunato2010community}
S.~Fortunato, ``Community detection in graphs,'' {\em Physics reports}, vol.~486, no.~3-5, pp.~75--174, 2010.

\bibitem{DMon}
A.~Tsitsulin, J.~Palowitch, B.~Perozzi, and E.~M{\"u}ller, ``Graph clustering with graph neural networks,'' {\em Journal of Machine Learning Research}, vol.~24, no.~127, pp.~1--21, 2023.

\bibitem{SC}
S.~D. Kamvar, D.~Klein, and C.~D. Manning, ``Spectral learning,'' in {\em IJCAI}, vol.~3, pp.~561--566, 2003.

\bibitem{KmeansClust}
K.~Wagstaff, C.~Cardie, S.~Rogers, S.~Schr{\"o}dl, {\em et~al.}, ``Constrained k-means clustering with background knowledge,'' in {\em Icml}, vol.~1, pp.~577--584, 2001.

\bibitem{bruna2014spectral}
J.~Bruna, W.~Zaremba, A.~Szlam, and Y.~LeCun, ``Spectral networks and deep locally connected networks on graphs. arxiv,'' {\em arXiv preprint arXiv:1312.6203}, 2014.

\bibitem{defferrard2016convolutional}
M.~Defferrard, X.~Bresson, and P.~Vandergheynst, ``Convolutional neural networks on graphs with fast localized spectral filtering,'' {\em Advances in neural information processing systems}, vol.~29, 2016.

\bibitem{loukas2019graph}
A.~Loukas, ``Graph reduction with spectral and cut guarantees.,'' {\em J. Mach. Learn. Res.}, vol.~20, no.~116, pp.~1--42, 2019.

\bibitem{kataria2023linear}
M.~Kataria, A.~Khandelwal, R.~Das, S.~Kumar, and J.~Jayadeva, ``Linear complexity framework for feature-aware graph coarsening via hashing,'' in {\em NeurIPS 2023 Workshop: New Frontiers in Graph Learning}, 2023.

\bibitem{FGC}
M.~Kumar, A.~Sharma, and S.~Kumar, ``A unified framework for optimization-based graph coarsening,'' {\em Journal of Machine Learning Research}, vol.~24, no.~118, pp.~1--50, 2023.

\bibitem{kataria2024ugc}
M.~Kataria, S.~Kumar, and J.~Jayadeva, ``{UGC}: Universal graph coarsening,'' in {\em The Thirty-eighth Annual Conference on Neural Information Processing Systems}, 2024.

\bibitem{Zhu}
X.~Zhu, Z.~Ghahramani, and J.~D. Lafferty, ``Semi-supervised learning using gaussian fields and harmonic functions,'' in {\em Proceedings of the 20th International conference on Machine learning (ICML-03)}, pp.~912--919, 2003.

\bibitem{koller2009probabilistic}
D.~Koller and N.~Friedman, {\em Probabilistic graphical models: principles and techniques}.
\newblock MIT press, 2009.

\bibitem{banerjee2008model}
O.~Banerjee, L.~El~Ghaoui, and A.~d'Aspremont, ``Model selection through sparse maximum likelihood estimation for multivariate gaussian or binary data,'' {\em The Journal of Machine Learning Research}, vol.~9, pp.~485--516, 2008.

\bibitem{dempster1972covariance}
A.~P. Dempster, ``Covariance selection,'' {\em Biometrics}, pp.~157--175, 1972.

\bibitem{belkin2006manifold}
M.~Belkin, P.~Niyogi, and V.~Sindhwani, ``Manifold regularization: A geometric framework for learning from labeled and unlabeled examples.,'' {\em Journal of machine learning research}, vol.~7, no.~11, 2006.

\bibitem{hu2013graph}
C.~Hu, L.~Cheng, J.~Sepulcre, G.~El~Fakhri, Y.~M. Lu, and Q.~Li, ``A graph theoretical regression model for brain connectivity learning of alzheimer's disease,'' in {\em 2013 IEEE 10th International Symposium on Biomedical Imaging}, pp.~616--619, IEEE, 2013.

\bibitem{van2019three}
G.~M. Van~de Ven and A.~S. Tolias, ``Three scenarios for continual learning,'' {\em arXiv preprint arXiv:1904.07734}, 2019.

\bibitem{zhang2022cglb}
X.~Zhang, D.~Song, and D.~Tao, ``Cglb: Benchmark tasks for continual graph learning,'' {\em Advances in Neural Information Processing Systems}, vol.~35, pp.~13006--13021, 2022.

\bibitem{parisi2019continual}
G.~I. Parisi, R.~Kemker, J.~L. Part, C.~Kanan, and S.~Wermter, ``Continual lifelong learning with neural networks: A review,'' {\em Neural networks}, vol.~113, pp.~54--71, 2019.

\bibitem{kim2022dygrain}
S.~Kim, S.~Yun, and J.~Kang, ``Dygrain: An incremental learning framework for dynamic graphs.,'' in {\em IJCAI}, pp.~3157--3163, 2022.

\bibitem{wu2023continual}
T.~Wu, Q.~Liu, Y.~Cao, Y.~Huang, X.-M. Wu, and J.~Ding, ``Continual graph convolutional network for text classification,'' in {\em Proceedings of the AAAI Conference on Artificial Intelligence}, vol.~37, pp.~13754--13762, 2023.

\bibitem{you2022roland}
J.~You, T.~Du, and J.~Leskovec, ``Roland: graph learning framework for dynamic graphs,'' in {\em Proceedings of the 28th ACM SIGKDD conference on knowledge discovery and data mining}, pp.~2358--2366, 2022.

\bibitem{xiang2010temporal}
L.~Xiang, Q.~Yuan, S.~Zhao, L.~Chen, X.~Zhang, Q.~Yang, and J.~Sun, ``Temporal recommendation on graphs via long-and short-term preference fusion,'' in {\em Proceedings of the 16th ACM SIGKDD international conference on Knowledge discovery and data mining}, pp.~723--732, 2010.

\bibitem{wang2020microsoft}
K.~Wang, Z.~Shen, C.~Huang, C.-H. Wu, Y.~Dong, and A.~Kanakia, ``Microsoft academic graph: When experts are not enough,'' {\em Quantitative Science Studies}, vol.~1, no.~1, pp.~396--413, 2020.

\bibitem{bianchi2020spectral}
F.~M. Bianchi, D.~Grattarola, and C.~Alippi, ``Spectral clustering with graph neural networks for graph pooling,'' in {\em International conference on machine learning}, pp.~874--883, PMLR, 2020.

\bibitem{bianchi2022simplifying}
F.~M. Bianchi, ``Simplifying clustering with graph neural networks,'' {\em arXiv preprint arXiv:2207.08779}, 2022.

\bibitem{newman2006modularity}
M.~E. Newman, ``Modularity and community structure in networks,'' {\em Proceedings of the national academy of sciences}, vol.~103, no.~23, pp.~8577--8582, 2006.

\bibitem{zhao2012consistency}
Y.~Zhao, E.~Levina, and J.~Zhu, ``Consistency of community detection in networks under degree-corrected stochastic block models,'' 2012.

\bibitem{kipf2016semi}
T.~N. Kipf and M.~Welling, ``Semi-supervised classification with graph convolutional networks,'' {\em arXiv preprint arXiv:1609.02907}, 2016.

\bibitem{hamilton2017inductive}
W.~Hamilton, Z.~Ying, and J.~Leskovec, ``Inductive representation learning on large graphs,'' {\em Advances in neural information processing systems}, vol.~30, 2017.

\bibitem{xu2018powerful}
K.~Xu, W.~Hu, J.~Leskovec, and S.~Jegelka, ``How powerful are graph neural networks?,'' {\em arXiv preprint arXiv:1810.00826}, 2018.

\bibitem{velickovic2017graph}
P.~Velickovic, G.~Cucurull, A.~Casanova, A.~Romero, P.~Lio, Y.~Bengio, {\em et~al.}, ``Graph attention networks,'' {\em stat}, vol.~1050, no.~20, pp.~10--48550, 2017.

\bibitem{zeng2019graphsaint}
H.~Zeng, H.~Zhou, A.~Srivastava, R.~Kannan, and V.~Prasanna, ``Graphsaint: Graph sampling based inductive learning method,'' {\em arXiv preprint arXiv:1907.04931}, 2019.

\bibitem{Bhatia16}
K.~Bhatia, K.~Dahiya, H.~Jain, P.~Kar, A.~Mittal, Y.~Prabhu, and M.~Varma, ``The extreme classification repository: Multi-label datasets and code,'' 2016.

\bibitem{jerrum1988conductance}
M.~Jerrum and A.~Sinclair, ``Conductance and the rapid mixing property for markov chains: the approximation of permanent resolved,'' in {\em Proceedings of the twentieth annual ACM symposium on Theory of computing}, pp.~235--244, 1988.

\bibitem{moon2019visualizing}
K.~R. Moon, D.~Van~Dijk, Z.~Wang, S.~Gigante, D.~B. Burkhardt, W.~S. Chen, K.~Yim, A.~v.~d. Elzen, M.~J. Hirn, R.~R. Coifman, {\em et~al.}, ``Visualizing structure and transitions in high-dimensional biological data,'' {\em Nature biotechnology}, vol.~37, no.~12, pp.~1482--1492, 2019.

\bibitem{lecun2010mnist}
Y.~LeCun, C.~Cortes, and C.~Burges, ``Mnist handwritten digit database,'' {\em ATT Labs [Online]. Available: http://yann.lecun.com/exdb/mnist}, vol.~2, 2010.

\bibitem{pennington2014glove}
J.~Pennington, R.~Socher, and C.~D. Manning, ``Glove: Global vectors for word representation,'' in {\em Proceedings of the 2014 conference on empirical methods in natural language processing (EMNLP)}, pp.~1532--1543, 2014.

\bibitem{zachary1977information}
W.~W. Zachary, ``An information flow model for conflict and fission in small groups,'' {\em Journal of anthropological research}, vol.~33, no.~4, pp.~452--473, 1977.

\bibitem{datar2004locality}
M.~Datar, N.~Immorlica, P.~Indyk, and V.~S. Mirrokni, ``Locality-sensitive hashing scheme based on p-stable distributions,'' in {\em Proceedings of the twentieth annual symposium on Computational geometry}, pp.~253--262, 2004.

\bibitem{lu2011link}
L.~L{\"u} and T.~Zhou, ``Link prediction in complex networks: A survey,'' {\em Physica A: statistical mechanics and its applications}, vol.~390, no.~6, pp.~1150--1170, 2011.

\bibitem{vogelstein2012graph}
J.~T. Vogelstein, W.~G. Roncal, R.~J. Vogelstein, and C.~E. Priebe, ``Graph classification using signal-subgraphs: Applications in statistical connectomics,'' {\em IEEE transactions on pattern analysis and machine intelligence}, vol.~35, no.~7, pp.~1539--1551, 2012.

\bibitem{jin2021graph}
W.~Jin, L.~Zhao, S.~Zhang, Y.~Liu, J.~Tang, and N.~Shah, ``Graph condensation for graph neural networks,'' {\em arXiv preprint arXiv:2110.07580}, 2021.

\bibitem{yang2016revisiting}
Z.~Yang, W.~Cohen, and R.~Salakhudinov, ``Revisiting semi-supervised learning with graph embeddings,'' in {\em International conference on machine learning}, pp.~40--48, PMLR, 2016.

\bibitem{shchur2018pitfalls}
O.~Shchur, M.~Mumme, A.~Bojchevski, and S.~G{\"u}nnemann, ``Pitfalls of graph neural network evaluation,'' {\em arXiv preprint arXiv:1811.05868}, 2018.

\bibitem{fu2020magnn}
X.~Fu, J.~Zhang, Z.~Meng, and I.~King, ``Magnn: Metapath aggregated graph neural network for heterogeneous graph embedding,'' in {\em Proceedings of The Web Conference 2020}, pp.~2331--2341, 2020.

\bibitem{yang2022scbert}
F.~Yang, W.~Wang, F.~Wang, Y.~Fang, D.~Tang, J.~Huang, H.~Lu, and J.~Yao, ``scbert as a large-scale pretrained deep language model for cell type annotation of single-cell rna-seq data,'' {\em Nature Machine Intelligence}, vol.~4, no.~10, pp.~852--866, 2022.

\bibitem{pygsp}
M.~Defferrard, L.~Martin, R.~Pena, and N.~Perraudin, ``Pygsp: Graph signal processing in python.''

\bibitem{watts1998collective}
D.~J. Watts and S.~H. Strogatz, ``Collective dynamics of ‘small-world’networks,'' {\em nature}, vol.~393, no.~6684, pp.~440--442, 1998.

\end{thebibliography}
